\documentclass[10pt,twocolumn,letterpaper]{article}

\usepackage{ijcb}
\usepackage{times}
\usepackage{epsfig}
\usepackage{graphicx}
\usepackage{amsmath}
\usepackage{amssymb}
\usepackage[norule,symbol,perpage]{footmisc}

\usepackage{makecell}
\usepackage{booktabs}
\usepackage{subfig}
\usepackage{multirow}

\usepackage[pagebackref=true,breaklinks=true,letterpaper=true,colorlinks,bookmarks=false]{hyperref}

\ijcbfinalcopy 

\ifijcbfinal\fi

\makeatother

\begin{document}

\title{MiDeCon: Unsupervised and Accurate Fingerprint and Minutia Quality Assessment based on Minutia Detection Confidence}

\newcommand\Mark[1]{\textsuperscript#1}

\author{Philipp Terh\"{o}rst\Mark{1}\Mark{2}, Andr\'e Boller\Mark{2}, Naser Damer\Mark{1}\Mark{2}, Florian Kirchbuchner\Mark{1}\Mark{2}, Arjan Kuijper\Mark{1}\Mark{2}\\
\Mark{1}Fraunhofer Institute for Computer Graphics Research IGD, Darmstadt, Germany\\
\Mark{2}Technical University of Darmstadt, Darmstadt, Germany\\
Email:{\{philipp.terhoerst, naser.damer, florian.kirchbuchner, arjan.kuijper\}@igd.fraunhofer.de}
}

\maketitle
\thispagestyle{empty}


\begin{abstract}
\vspace{-2mm}
An essential factor to achieve high accuracies in fingerprint recognition systems is the quality of its samples. 
Previous works mainly proposed supervised solutions based on image properties that neglects the minutiae extraction process, despite that most fingerprint recognition techniques are based on detected minutiae.
Consequently, a fingerprint image might be assigned a high quality even if the utilized minutia extractor produces unreliable information. 
In this work, we propose a novel concept of assessing minutia and fingerprint quality based on minutia detection confidence (MiDeCon).
MiDeCon can be applied to an arbitrary deep learning based minutia extractor and does not require quality labels for learning.
We propose using the detection reliability of the extracted minutia as its quality indicator.
By combining the highest minutia qualities, MiDeCon also accurately determines the quality of a full fingerprint.
Experiments are conducted on the publicly available databases of the FVC 2006 and compared against several baselines, such as NIST's widely-used fingerprint image quality software NFIQ1 and NFIQ2.
The results demonstrate a significantly stronger quality assessment performance of the proposed MiDeCon-qualities as related works on both, minutia- and fingerprint-level.
The implementation is publicly available.
\end{abstract}
\vspace{-7mm}

\section{Introduction}
\vspace{-0mm}
Fingerprints are one of the most used biometric modalities 
with a wide range of applications such as unlocking consumer smartphones or identity recognition in forensic investigations \cite{DBLP:journals/prl/JainNR16, DBLP:conf/fusion/DamerT0K18}.
The most determinant factor to achieve high recognition accuracy in fingerprint recognition systems is the fingerprint sample quality \cite{DBLP:conf/biosig/GalballyHB18}.
Therefore, many large scale biometric deployments require measuring quality scores, such as the Unique Identification Authority of India (UIDAI) \cite{UIDAI} and the European Union Visa Information System (VIS) \cite{EUVIS}.

In general, the quality of a fingerprint sample is defined as its utility for recognition \cite{DBLP:reference/bio/TabassiG15a}.
The fingerprint quality is the key property for a high recognition performance by ensuring reliable enrolment and matching.
Especially in forensic scenarios, quality assessment is of great significance since the assessment of the strength of evidence is a central activity in forensic casework \cite{DBLP:conf/btas/ZeinstraMVS18}.
However, forensic use-cases mainly have to deal with partial fingerprints and thus, the quality assessment of single minutiae are of great interest \cite{DBLP:conf/cvpr/EzeobiejesiB18, DBLP:journals/access/SankaranVS14}.

Current works on fingerprint quality assessment, such as \cite{DBLP:conf/cvpr/EzeobiejesiB18, DBLP:journals/iet-bmt/TertychnyiOA18}, focused on supervised classifiers to predict quality given the fingerprint image.
However, these approaches often neglect the quality of the extracted minutiae for the fact that current fingerprint recognition systems are mainly based on this minutiae information \cite{DBLP:journals/access/Valdes-RamirezM19}.
Previous works on the quality assessment of single minutiae are based on hand-crafted features and define the quality of a minutia through properties of the image quality \cite{DBLP:conf/cw/MakniC20, 4401958, NBIS}.
Therefore, these solutions do not consider how well the utilized minutia extractor can extract true minutiae.
Consequently, even if the image around a minutia is of high quality, the extraction of reliable minutia information might fail and thus, the quality of this minutia is wrongly estimated as high.

In this work, we propose a novel concept of measuring minutia and fingerprint quality based on minutia detection confidence (MiDeCon).
The proposed concept can be applied to an arbitrary neural network based minutia extractor and does not rely on quality labels for training.
To estimate the quality of a minutia, MiDeCon computes the extractor's confidence that the detected minutia is real.
This confidence is determined over the agreement of stochastic variants of the extraction network created by random node eliminations.
To estimate the quality of a full fingerprint, MiDeCon combines the highest minutia qualities resulting in a continuous quality value of the fingerprint.

The experiments were conducted on the four publicly available databases of the FVC 2006 \cite{FVC2006} and make utilizes the widely-used NFIQ1 \cite{NFIQ1} and NFIQ2 \cite{NFIQ2} developed by the National Institute of Standards and Technology (NIST) as baselines. 
The results demonstrate significantly stronger fingerprint and minutia quality assessment performances of the proposed unsupervised approach in comparison to established approaches.
This holds on all investigated sensor types (electric field, optical, and thermal sensors).
MiDeCon is publicly available under the following link\footnote{\url{https://github.com/pterhoer/FingerprintImageQuality}}.
The main novelty of the proposed MiDeCon approach is that it:
\begin{itemize}
\itemsep0mm
\item \textbf{Does not require quality labels for training -} 
Previous works often rely on error-prone labelling mechanisms without a clear definition of quality. Our approach avoids the use of inaccurate quality labels by using the minutia detection confidence as a quality estimate. Moreover, the training state can be completely avoided if pre-trained minutiae extraction neural network trained with dropout \cite{DBLP:journals/jmlr/SrivastavaHKSS14} is available.
\item \textbf{Considers difficulties in the minutiae extraction -}
Previous works estimates the quality of a fingerprint based on the properties of the image neglecting the minutiae extraction process. However, the extraction process might face difficulties that are not considered in the image properties and thus, produce unreliable minutia information. Our solution defines quality through the prediction confidence of the extractor and thus, considers this problem.
\item \textbf{Produces continuous quality values -}
While previous works often categorize the quality outputs in discrete categories (e.g. \{good, bad, ugly\} \cite{DBLP:conf/cvpr/EzeobiejesiB18}; \{1,2,3,4,5\} \cite{NFIQ1}), our approach produces continuous quality values that allow more fine-grained and flexible enrolment and matching processes.
\item \textbf{Includes quality assessment of single minutiae -}
Unlike previous works, our solution assesses the quality of full fingerprints as well as the quality of single minutiae. This is specifically useful in forensic scenarios where forensic examiners aim to find reliable minutiae suitable for identification. 
\end{itemize}

\section{Related Works}

\subsection{Minutia Quality Assessment}

Despite that many of the best fingerprint recognition algorithms are based on minutiae information \cite{DBLP:journals/access/Valdes-RamirezM19}, only a few works proposed solutions for estimating the quality of single minutiae.
In 2005, Kim et al. \cite{DBLP:conf/avbpa/Kim05} proposed a minutia quality score by analysing the neighbouring ridge structure of the minutia in a thinned fingerprint image.
In \cite{4401958}, 
Uz et al. defined the quality of a minutia based on its frequency it occurs in the different impressions.
Maki et al. proposed the Minutia Confidence Index \cite{DBLP:conf/cw/MakniC20}.
This index estimates the quality of a minutia based on the location, type, and orientation of other minutiae.
NIST roll out Mindtct, a minutiae detection algorithm that additionally states the qualities of detected minutiae by combing low contrast, low flow, and high curve maps of the fingerprint. 

To summarize, previous works define the quality of a minutia through the image quality around the minutia.
However, this does not consider the extraction process which might result in error-prone outputs, e.g. due to model bias.
Therefore, we propose defining the quality of a minutia as the detection confidence of the minutia extractor.

\subsection{Fingerprint Quality Assessment}

In contrast to minutia quality assessment, more research was done on the quality assessment on full fingerprints.
Traditional approaches are based on the use of local \cite{DBLP:conf/avbpa/ShenKK01, DBLP:conf/icip/TabassiW05, DBLP:conf/icb/0001LPLT11, DBLP:conf/iwcf/YoonLJ14, DBLP:conf/caip/SharmaD19} and global hand-crafted \cite{DBLP:conf/icip/LimJY02, DBLP:conf/avbpa/ChenDJ05, DBLP:conf/icb/PhromsuthirakA13, DBLP:conf/eusipco/El-AbedNCR13, DBLP:conf/ciarp/VasconcelosP18, DBLP:journals/tifs/RichterGMTH19} features, or approaches that combine both kinds of features \cite{DBLP:journals/tifs/Alonso-FernandezFOGFKB07, DBLP:conf/btas/Yoon0LJ13, DBLP:conf/btas/SankaranVS13, DBLP:journals/pami/TeixeiraL17}.
Besides the use of hand-crafted features, deep feature-learning solutions \cite{DBLP:journals/iet-bmt/TertychnyiOA18, DBLP:conf/cvpr/OlsenTMB13, DBLP:conf/cvpr/EzeobiejesiB18} were proposed recently.
These approaches aim to predict quality by supervised training of a neural network classifier.

The engagement of NIST in this aspect is also worth mentioning.
They investigated, along with their partners, 155 quality features from the literature and selected the best 14 to train classification models to estimate the quality of a fingerprint.
This resulted in of NFIQ1 and NFIQ2 \cite{DBLP:conf/biosig/BausingerT11, NBIS}.

In concept, current solutions are training supervised classifiers to estimate the quality of a fingerprint image. 
Since current fingerprint recognition systems are still based on minutiae information \cite{DBLP:journals/access/Valdes-RamirezM19} we argue that the quality of a fingerprint must reflect the qualities of its extracted minutiae.
Therefore, we propose to estimate the quality of a fingerprint by combining minutia qualities  defined over the detection confidences of the minutia extractor.

\begin{figure}
\centering
\includegraphics[width=0.5\textwidth]{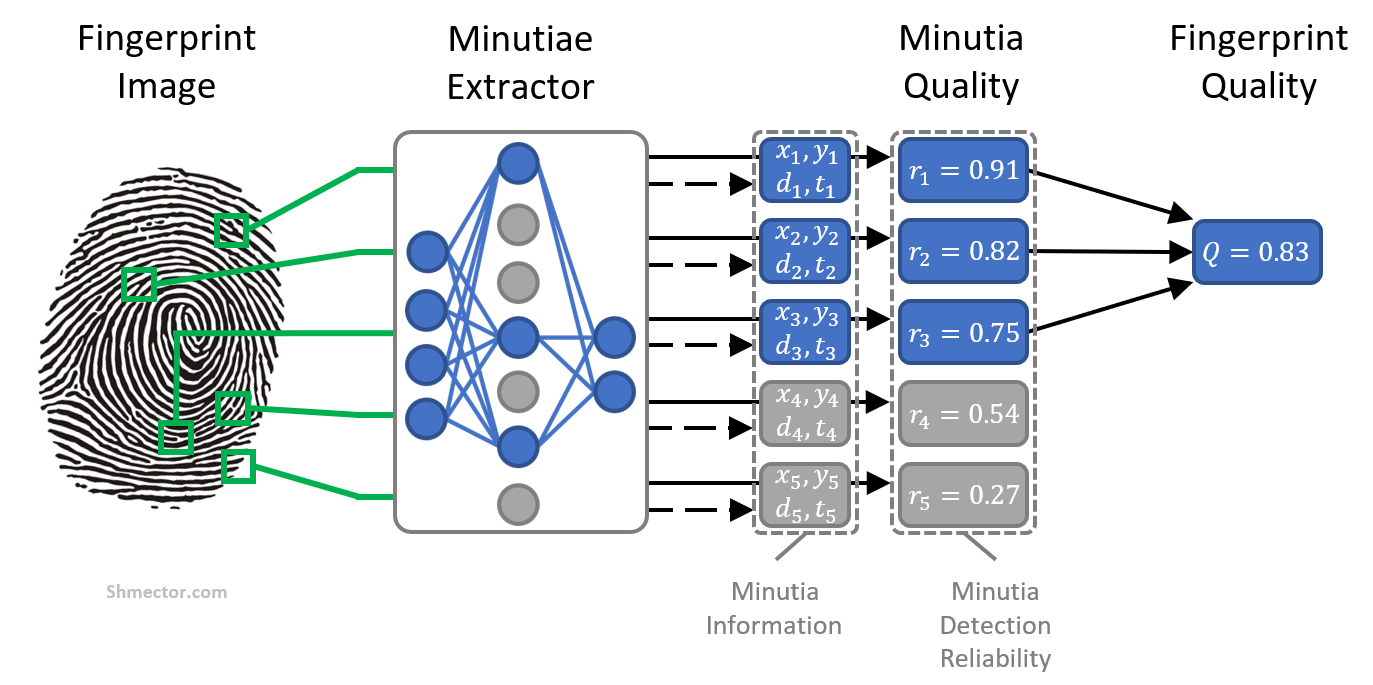}
\caption{Conceptual idea of MiDeCon: the system's minutiae extraction network is modified to additionally provide the minutia detection reliabilities based on the agreement of stochastic network variants. We propose using the detection reliability $r^{(i)}$ of a minutia $m^{(i)}$ as its quality indicator.
To determine the quality of the whole fingerprint, the $n$ highest detection reliabilities are combined. 
By integrating the minutia extractor in the quality assessment process, the proposed concept additionally takes into account difficulties that appear during extraction. \vspace{-3mm}
}
\label{fig:Concept}
\end{figure}

\vspace{-1mm}
\section{Methodology}
\label{sec:Methodology}








We define the quality of a single minutia by its detection reliability of the minutia extractor that is based on the agreement between stochastic variants of the extractor network.
Combing the detection reliabilities of the highest-quality minutiae, MiDeCon assesses the quality of a given fingerprint.
The conceptual idea of the proposed approach is shown in Figure \ref{fig:Concept}.
In the following, we provide a detailed description of how MiDeCon determines the quality of single minutiae and fingerprints given the minutiae extractor MinutiaeNet \cite{DBLP:conf/icb/Nguyen0018}.
MinutiaeNet consists of two networks, CoarseNet and FineNet.
While CoarseNet determines and extracts patches of potential minutiae in a given Fingerprint, FineNet estimates if the candidate patch consists of a minutia and extracts the minutia information from the candidate patches, such as location, orientation, and minutia type.

\vspace{-2mm}
\subsection{Assessing Minutia Quality}

Our minutia quality assessment approach is based on the output of FineNet \cite{DBLP:conf/icb/Nguyen0018} that decides if the candidate patch of the fingerprint consists of a minutia or not.
We calculate a novel minutia detection confidence based of this network and use this as the quality of the detected minutia.

More formally, given a fingerprint capture $f$ and a neural network minutiae extractor \textit{ME}, such as FineNet \cite{DBLP:conf/icb/Nguyen0018}, we propose to use the detection reliability $r^{(i)}$ of the i$^{th}$ minutia as its quality indicator $q^{(i)}$.
The detection reliability follows the idea of  Terh\"{o}rst et al. \cite{DBLP:conf/btas/TerhorstHKZDKK19} by using stochastic forward passes in a dropout-reduced neural network minutiae extractor \textit{ME} trained with dropout \cite{DBLP:journals/jmlr/SrivastavaHKSS14}.
Instead of propagating the input patch one time through the network \textit{ME}, the input image is propagated $m=100$ times through different variations of the network approximating a Gaussian process \cite{DBLP:conf/icml/GalG16}.
The network variants of \textit{ME} are created by randomly eliminating nodes using similar dropout patterns as during the model's training.
For each of the $m$ stochastic forward passes through such a model variant, a stochastic softmax prediction $x_j$ is given describing the probability that the given patch contains a minutia assuming a Boltzmann distribution.
This results in a set of stochastic prediction $x^{(i)}=\{x^{(i)}_j\}_{j=1,\dots,m}$ that is used to estimate the detection reliability of a given minutia $m^{(i)}$.
The detection reliability
\begin{align*}
rel(x^{(i)}) =   \underbrace{\mathbb{E}\left(x^{(i)}\right) }_{\text{MOC}} +  \underbrace{\mathbb{E}\left( \left( x^{(i)} - \mathbb{E}\left( x^{(i)} \right) \right)^2 \right)}_{\text{MOD}}, \label{eq:MinuQuality}
\end{align*}
for a minutia $m^{(i)}$ consists of two parts, the measure of centrality (MOC) and the measure of dispersion (MOD).
The MOC computes the mean of the stochastic outputs aiming at utilizing the probability interpretation of softmax assuming a Boltzmann distribution.
The MOD quantifies the agreement of the different network variants by calculating the mean pairwise absolute distance of all scores in $x^{(i)}$.
A higher agreement and a higher probability indicate a higher detection reliability and vice versa.
MOC and MOD are equally weighted since both measures result in values of similar magnitudes.
We propose using the minutia detection confidence $r^{(i)} = rel(x^{(i)})$ of the minutia $m^{(i)}$ as its quality indicator $q^{(i)}$.

The advantage of using the minutia detection confidence as the quality measure of a single minutia is that, unlike previous works, it takes into account the minutia extraction process, rather than solely depend on the properties of the image patch around the detection.
Neglecting this process can result in over-estimations of the quality when the extractor is not able to reliably detect a minutia of a high-quality image or in under-estimations of the quality in case that the extractor is still able to capture reliable minutia information from a lower-quality image.
Our proposed solution defines the quality of a single minutia through the prediction confidence of the extractor and thus, considers this issue.

\subsection{Assessing Fingerprint Quality}

Given a fingerprint $f$ with extracted minutiae information $\{m^{(i)}\}$ and its corresponding detection reliabilities $r^{(i)}$, we assume that the quality of a fingerprint is based on the highest detection reliabilities of its minutiae. 
Therefore, we define the quality $Q(f)$ of a fingerprint $f$ as 
\begin{align}
Q(f) = \mathbb{E}\left( \left\lbrace \bar{r}^{(l)}(f): l \in \mathbb{N}_+ \, \wedge \, l < n  \right\rbrace  \right),
\end{align}
where $\mathbb{E}$ is mean operator over the first $n$ entries of the sorted list $\bar{r}(f)$.
More precisely, $\bar{r}(f)$ is the sorted list of the minutiae reliabilities $\{r^{(i)}\}$ of fingerprint $f$ such that $\forall \quad l \in [1, \dots N-1]: r^{(l)} \geq r^{(l+1)}$ is satisfied.
Consequently, we define the quality $Q$ of a fingerprint capture $f$ as the mean of its $n$ highest minutia (detection) reliabilities.
We choose $n=20$ since this reflects the average number of minutiae in a partial fingerprint and allows the proposed methodology to be used in forensic scenarios \cite{DBLP:conf/accv/TerhorstD0K18}.
However, this number can be arbitrarily adapted to the target application.

The advantage of this quality assessment approach is its direct use of the minutia qualities defined over their detection reliabilities.
Most of the current fingerprint recognition systems are based on minutiae information  \cite{DBLP:journals/access/Valdes-RamirezM19} and thus, measuring the quality of the direct input of these systems is more likely to succeed in estimating suitable quality values than approaches that only rely on properties of the image quality.

\section{Experimental Setup}


\vspace{-1mm}
\subsection{Databases and Model Training}
Th experiments are done on the publicly available benchmarks of the FVC 2006 \cite{FVC2006}.
This allows us to evaluate quality assessment over multiple sensor types.
FVC2006 consists of 7200 fingerprint captures from 600 different fingers over four databases (DB).
The images were collected by four different organizations in two different countries.
Each database was captured with a different sensor type: DB1 (electric field sensor, 250 dpi), DB2 (optical sensor, 596 dpi), DB3 (thermal sweeping sensor, 500 dpi), and DB4 (synthetic, 500 dpi).
DB4 consists of synthetic data that was generated with SFinFe \cite{DBLP:reference/bio/Cappelli15a}.
Each database consists of subject-disjoint test and training sets, resulting in 6720 fingerprint images for testing and 480 for training.

For the experiments, we apply MiDeCon on the FineNet network\footnote{\textit{https://github.com/luannd/MinutiaeNet}} \cite{DBLP:conf/icb/Nguyen0018}.
This network was originally not trained with dropout.
Therefore, we retrained the last fully connected layer with dropout on the training set following the training procedure described in \cite{DBLP:conf/icb/Nguyen0018}.
Since only one layer was trained, we adapted the initial learning rate to $lr=10^{-4}$.

\vspace{-1mm}
\subsection{Evaluation Metrics}
\label{sec:EvaluationMetrics}
To evaluate the fingerprint quality assessment performance, we follow the methodology by Grother et al. \cite{DBLP:journals/pami/GrotherT07} using error versus reject curves.
These curves show a verification error-rate over the fraction of unconsidered fingerprint images.
Based on the predicted quality values, these unconsidered images are those with the lowest predicted quality and the error rate is calculated on the remaining images.
Error versus reject curves indicates good quality estimation when the verification error decreases consistently for higher ratios of unconsidered images \cite{DBLP:conf/cvpr/TerhorstKDKK20}.

However, for quality assessment of single minutiae this assessment approach does not fit since the number of detected minutia strongly varies between different algorithms.
Therefore, we determine the quality assessment performance of single minutiae by considering only a fixed number (20-40) of the highest quality minutiae for recognition. 
Assuming a lower verification error is achieved if the considered minutiae are of higher quality, this allows us to evaluate the quality assessment performance of single minutiae.

The verification error rates within the error versus reject curves are reported in terms of false non-match rate (FNMR) at a fixed false match rate (FMR) as specified in the international standard \cite{ISO_Metrik}.

\vspace{-1mm}
\subsection{Workflow and Baselines}
Since many of the recognition systems are still based on minutiae information \cite{DBLP:journals/access/Valdes-RamirezM19}, we utilize two frequently used fingerprint recognition systems, Minutia Cylinder-Codes (MCC) from \cite{DBLP:journals/pami/CappelliFM10} and Bozorth3 \cite{NBIS} from NIST Biometric Image Software (NBIS).
The minutiae-extraction for the matcher is based on MinutiaeNet \cite{DBLP:conf/icb/Nguyen0018} and NIST's Mindtct \cite{NBIS}.
To evaluate the quality assessment performance, we compare the proposed solution against the widely-used fingerprint quality assessment solutions, NFIQ1 \cite{NFIQ1} and NFIQ2 \cite{NFIQ2} from NIST.
Both approaches have become a standard in estimating fingerprint quality \cite{DBLP:conf/icb/GalballyHFBT19}.

\vspace{-2mm}
\section{Results}

\vspace{-1mm}
\subsection{Minutia Quality Assessment}

In Table \ref{tab:MinutiaQualityFMR10-2}, the minutia quality assessment performance is compared in terms of FNMR at FMRs of $10^{-2}$ and $10^{-3}$ using only a fixed number the highest quality minutiae.
Moreover, the quality assessment approaches are compared on the four databases containing fingerprint images captured with different sensors.
Generally, considering more minutiae for the recognition process decreases the recognition error.
However, strong differences in the performance are observed for the different quality assessment methods.
Both utilized fingerprint recognition methods (Bozorth3 and MCC) focus on local minutiae neighbourhoods.
Since Mindtct assigns minutiae qualities over a quality map, neighbouring minutiae have similar quality estimates.
This property favours Mindtct when using Bozorth3 or MCC, because these focus on local neighbourhoods, and explains the weaker performance of CoarseNet.
Moreover, it shows that the natural choice of using the model's softmax output as the minutia quality does not work.

On synthetically generated fingerprint data, the quality estimates based on the Mindtct's hand-crafted features achieve the best results.
On data captured from real sensors, our proposed method based on the detection confidence of single minutiae constantly achieves the lowest recognition errors.
This holds for all investigated sensor types, i.e. electric field, optical, and thermal sensors (DB1, DB2, DB3) and demonstrates a strong minutiae quality assessment performance of the proposed approach.

\begin{table*}[]
\renewcommand{\arraystretch}{1.1}
\setlength{\tabcolsep}{4pt}
\centering
\caption{Evaluating minutia quality assessment - only a certain number of the highest quality minutiae are used for recognition. The recognition performance is reported at two FMRs on the Bozorth3 and the MCC matcher. Each DB was captured with a different sensor. Our proposed methodology based on minutia detection confidence shows lower recognition errors than related works \cite{DBLP:conf/icb/Nguyen0018, NBIS} in all cases, except on the less-relevant synthetic data (DB4). This demonstrates the strong quality estimation performance for single minutiae of the proposed method.}
\label{tab:MinutiaQualityFMR10-2}
\begin{tabular}{llrrrrrrrrrr}
\Xhline{2\arrayrulewidth} 
\multicolumn{2}{l}{FNMR@$10^{-2}$FMR} & \multicolumn{5}{c}{Bozorth3}                    & \multicolumn{5}{c}{MCC}                         \\
\cmidrule(lr){3-7} \cmidrule(lr){8-12}
                           &                       & Best 20 & Best 25 & Best 30 & Best 35 & Best 40 & Best 20 & Best 25 & Best 30 & Best 35 & Best 40 \\
                           \hline
\multirow{3}{*}{\rotatebox{90}{\makecell{DB1\\(electric)}}}       & Coarsenet             & 0.705   & 0.713   & 0.724   & 0.751   & 0.751   & 0.712   & 0.710   & 0.709   & 0.708   & 0.708   \\
                           & Mindtct               & 0.634   & 0.569   & 0.513   & 0.527   & 0.524   & 0.686   & 0.631   & 0.592   & 0.574   & 0.558   \\
                           & Ours      & \textbf{0.579}  & \textbf{0.542}   & \textbf{0.506}   & \textbf{0.487}   & \textbf{0.498}   & \textbf{0.670}   & \textbf{0.608}   & \textbf{0.580}   & \textbf{0.568}   & \textbf{0.553}   \\
                           \hline
\multirow{3}{*}{\rotatebox{90}{\makecell{DB2\\(optical)}}}        & Coarsenet             & 0.512   & 0.402   & 0.314   & 0.295   & 0.269   & 0.513   & 0.394   & 0.324   & 0.285   & 0.250   \\
                           & Mindtct               & 0.145   & 0.071   & 0.048   & 0.038   & 0.030   & 0.175   & 0.097   & 0.060   & 0.042   & 0.031   \\
                           & Ours      & \textbf{0.074}   & \textbf{0.043}   & \textbf{0.036}   & \textbf{0.029}   & \textbf{0.029}   & \textbf{0.145}   & \textbf{0.063}   & \textbf{0.041}   & \textbf{0.029}   & \textbf{0.023}   \\
                           \hline
\multirow{3}{*}{\rotatebox{90}{\makecell{DB3\\(thermal)}}}        & Coarsenet             & 0.635   & 0.607   & 0.568   & 0.553   & 0.524   & 0.656   & 0.603   & 0.578   & 0.570   & 0.563   \\
                           & Mindtct               & 0.217   & 0.157   & 0.121   & 0.107   & 0.099   & 0.234   & 0.170   & 0.133   & 0.117   & 0.110   \\
                           & Ours      & \textbf{0.136}   & \textbf{0.104}   & \textbf{0.089}   & \textbf{0.085}   & \textbf{0.085}   & \textbf{0.168}   & \textbf{0.114}   & \textbf{0.096}   & \textbf{0.085}   & \textbf{0.078}   \\
                           \hline
\multirow{3}{*}{\rotatebox{90}{\makecell{DB4\\(synthetic)}}}         & Coarsenet             & 0.873   & 0.845   & 0.839   & 0.836   & 0.840   & 0.855   & 0.821   & 0.798   & 0.796   & 0.797   \\
                           & Mindtct               & \textbf{0.568}   & \textbf{0.485}   & \textbf{0.440} & \textbf{0.394}   & \textbf{0.385}   & \textbf{0.593}   & \textbf{0.477}   & \textbf{0.395}   & \textbf{0.344}   & \textbf{0.313}   \\
                           & Ours      & 0.670   & 0.548   & 0.473   & 0.432   & 0.399   & 0.692   & 0.577   & 0.477   & 0.411   & 0.361  \vspace{1mm}\\ 
                           \Xhline{2\arrayrulewidth} 
                           \\
\end{tabular}
\vspace{-3mm}
\begin{tabular}{llrrrrrrrrrr} 
\Xhline{2\arrayrulewidth} 
\multicolumn{2}{l}{FNMR@$10^{-3}$FMR} & \multicolumn{5}{c}{Bozorth3}                   & \multicolumn{5}{c}{MCC}                         \\
\cmidrule(lr){3-7} \cmidrule(lr){8-12}
                           &                       & Best 20 & Best 25 & Best 30 & Best 35 & Best 40 & Best 20 & Best 25 & Best 30 & Best 35 & Best 40 \\
                           \hline
\multirow{3}{*}{\rotatebox{90}{\makecell{DB1\\(electric)}}}        & Coarsenet             & 0.818   & 0.829   & 0.843   & 0.856   & 0.855   & 0.835   & 0.840   & 0.837   & 0.838   & 0.838   \\
                           & Mindtct               & 0.722   & 0.674   & 0.654   & 0.666   & 0.671   & 0.769   & 0.720   & 0.690   & 0.671   & 0.653   \\
                           & Ours      & \textbf{0.699}   &\textbf{0.655}   & \textbf{0.626}   & \textbf{0.631}   & \textbf{0.650}   & \textbf{0.757}   & \textbf{0.708}   & \textbf{0.677}   & \textbf{0.663}   & \textbf{0.652}   \\
                           \hline
\multirow{3}{*}{\rotatebox{90}{\makecell{DB2\\(optical)}}}       & Coarsenet             & 0.645   & 0.506   & 0.440   & 0.386   & 0.371   & 0.658   & 0.506   & 0.412   & 0.370   & 0.335   \\
                           & Mindtct               & 0.223   & 0.124   & 0.082   & 0.062   & 0.052   & \textbf{0.237}   & 0.137   & 0.085   & 0.061   & 0.047   \\
                           & Ours      & \textbf{0.148}   & \textbf{0.080}   & \textbf{0.055}   & \textbf{0.046}   & \textbf{0.044}   & 0.240   & \textbf{0.098}   & \textbf{0.057}   & \textbf{0.041}   & \textbf{0.034}   \\
                           \hline
\multirow{3}{*}{\rotatebox{90}{\makecell{DB3\\(thermal)}}}       & Coarsenet             & 0.778   & 0.754   & 0.747   & 0.743   & 0.697   & 0.776   & 0.735   & 0.722   & 0.729   & 0.721   \\
                           & Mindtct               & 0.310   & 0.230   & 0.192   & 0.163   & 0.160   & 0.310   & 0.230   & 0.176   & 0.160   & 0.150   \\
                           & Ours      & \textbf{0.207}  & \textbf{0.158}   & \textbf{0.134}   & \textbf{0.124}   & \textbf{0.125}   & \textbf{0.238}   & \textbf{0.161}   & \textbf{0.130}   & \textbf{0.119}   & \textbf{0.112}  \\
                           \hline
\multirow{3}{*}{\rotatebox{90}{\makecell{DB4\\(synthetic)}}}        & Coarsenet             & 0.942   & 0.927   & 0.914   & 0.923   & 0.916   & 0.919   & 0.892   & 0.879   & 0.883   & 0.883   \\
                           & Mindtct               & \textbf{0.694}   & \textbf{0.605}   & \textbf{0.552} & \textbf{0.539}  & \textbf{0.515}   & \textbf{0.689}   & \textbf{0.575}   & \textbf{0.492}   & \textbf{0.442}   & \textbf{0.412}   \\
                           & Ours      & 0.772   & 0.691   & 0.612   & 0.567   & 0.541   & 0.791   & 0.678   & 0.584   & 0.519   & 0.465   \vspace{1mm}\\
                           \Xhline{2\arrayrulewidth} 
\end{tabular}
\end{table*}

\vspace{-1mm}
\subsection{Fingerprint Quality Assessment}

\begin{figure*}[h]
\captionsetup[subfloat]{farskip=5pt,captionskip=1pt}
\centering

\subfloat[DB1 (electric field sensor)\label{fig:FPQ_DB1_1_Bo}]{%
       \includegraphics[width=0.20\textwidth]{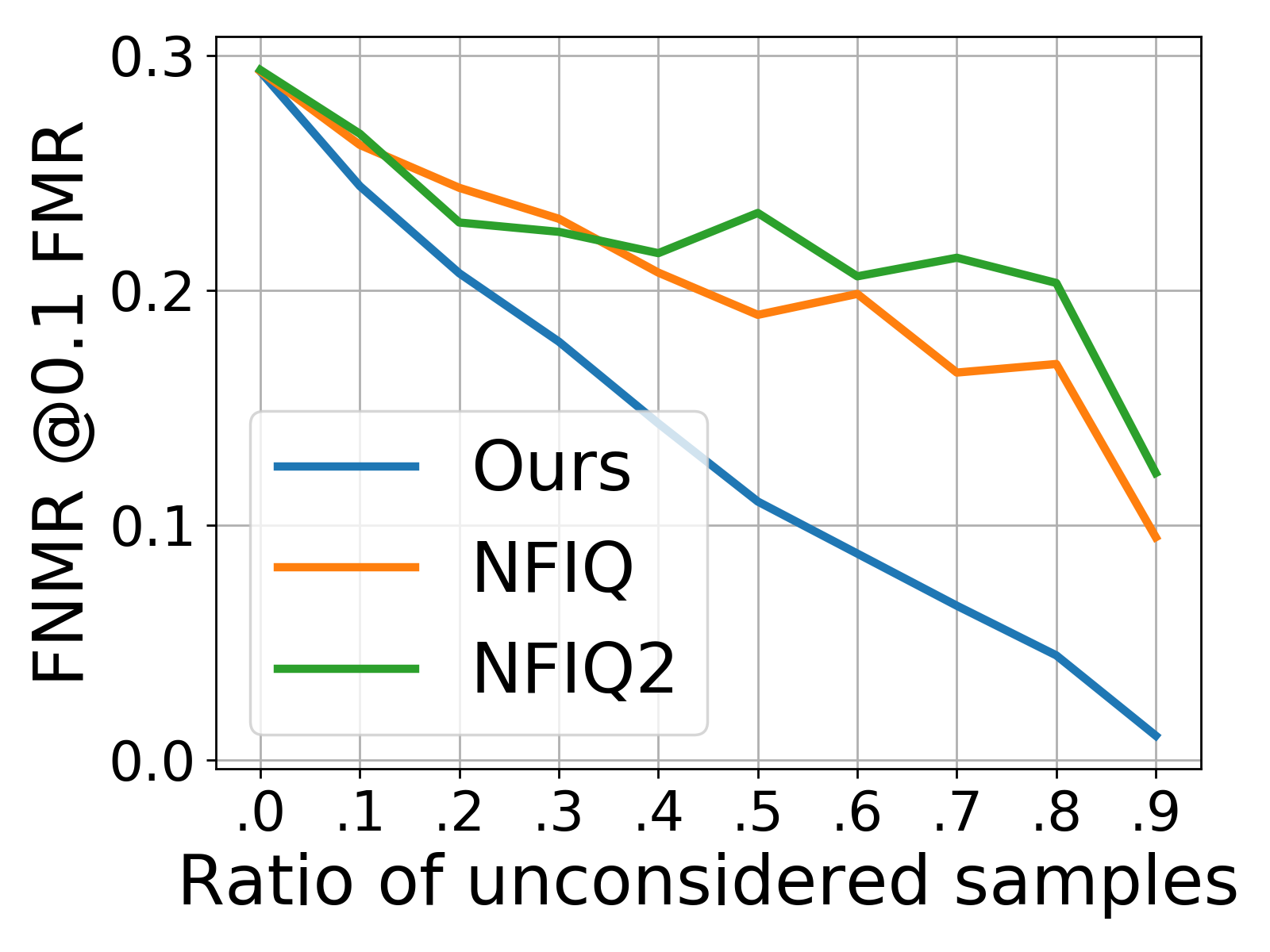}} 
\subfloat[DB2 (optical sensor) \label{fig:FPQ_DB2_1_Bo}]{%
       \includegraphics[width=0.20\textwidth]{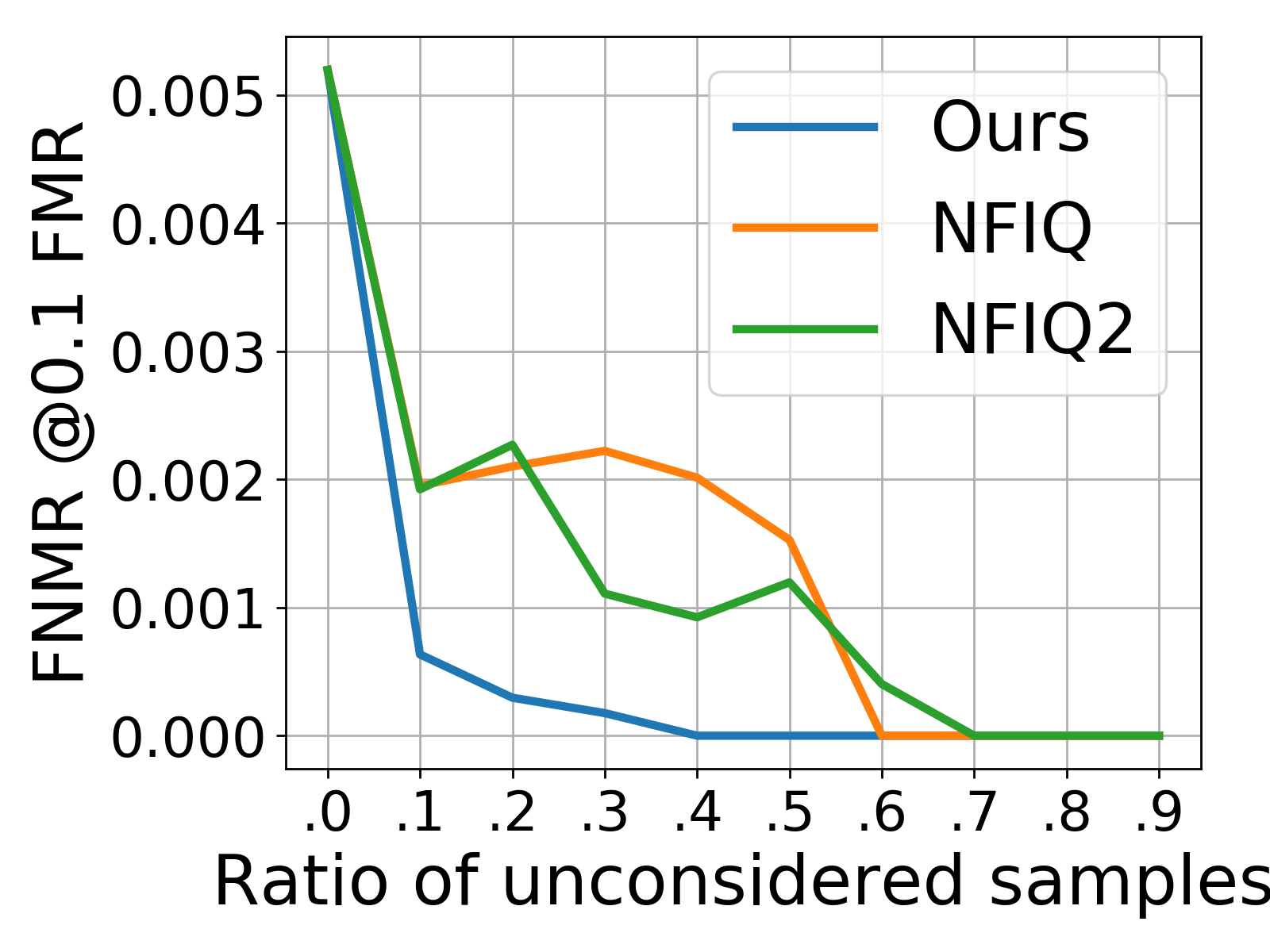}}   
\subfloat[DB3 (thermal sensor) \label{fig:FPQ_DB3_1_Bo}]{%
       \includegraphics[width=0.20\textwidth]{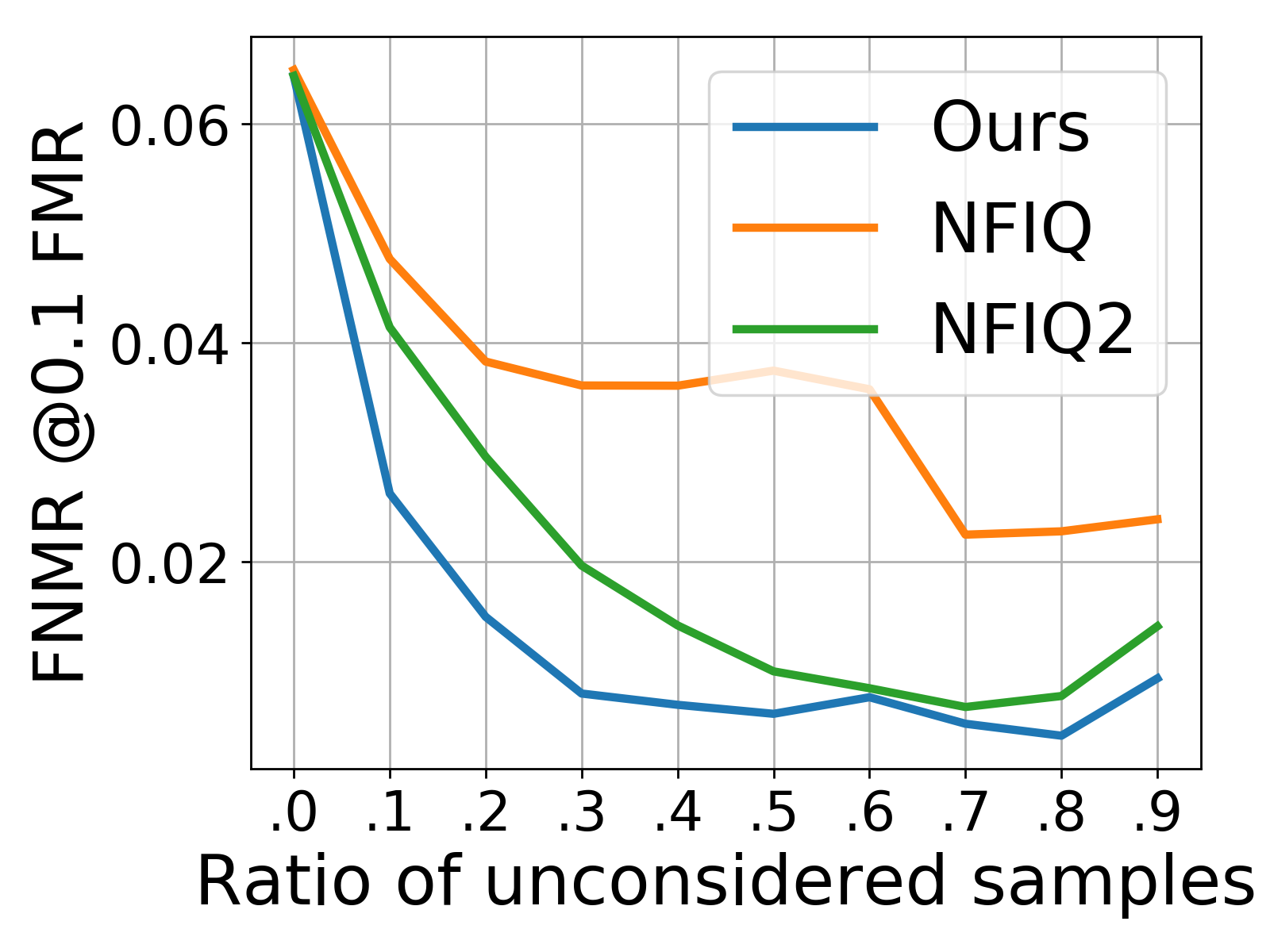}} 
\subfloat[DB4 (synthetic data) \label{fig:FPQ_DB4_1_Bo}]{%
       \includegraphics[width=0.20\textwidth]{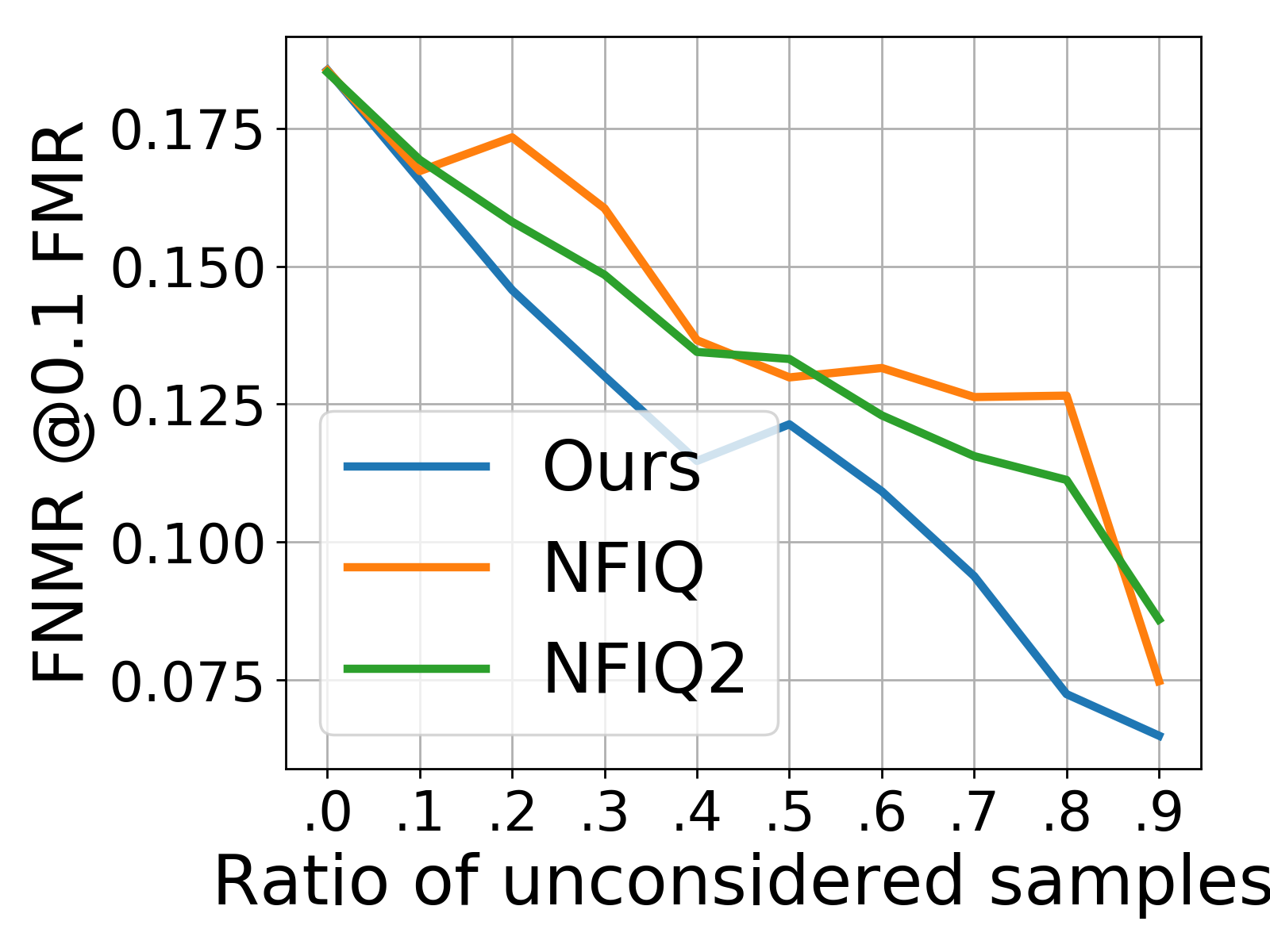}}
       
\subfloat[DB1 (electric field sensor)\label{fig:FPQ_DB1_01_Bo}]{%
       \includegraphics[width=0.20\textwidth]{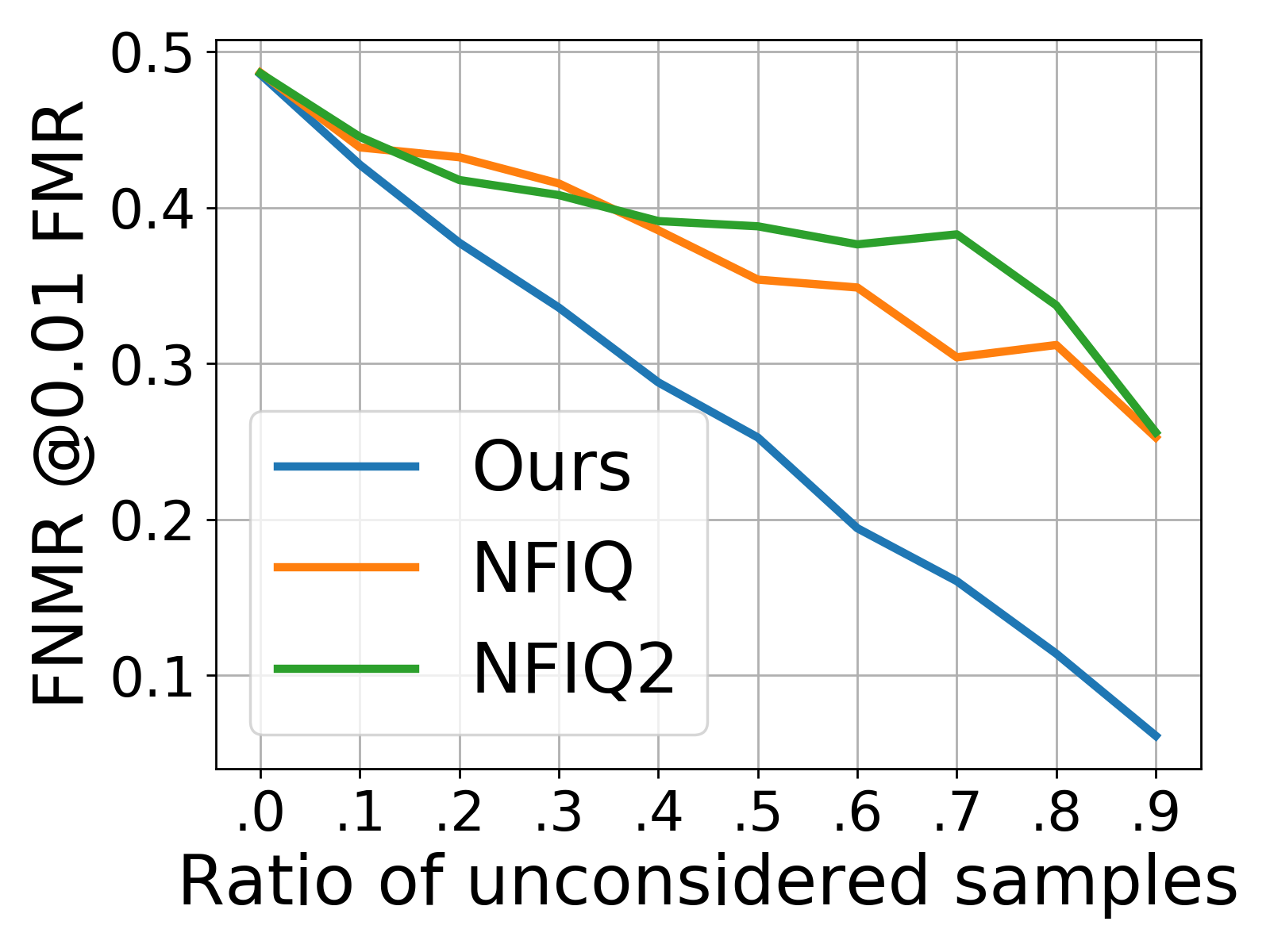}} 
\subfloat[DB2 (optical sensor) \label{fig:FPQ_DB2_01_Bo}]{%
       \includegraphics[width=0.20\textwidth]{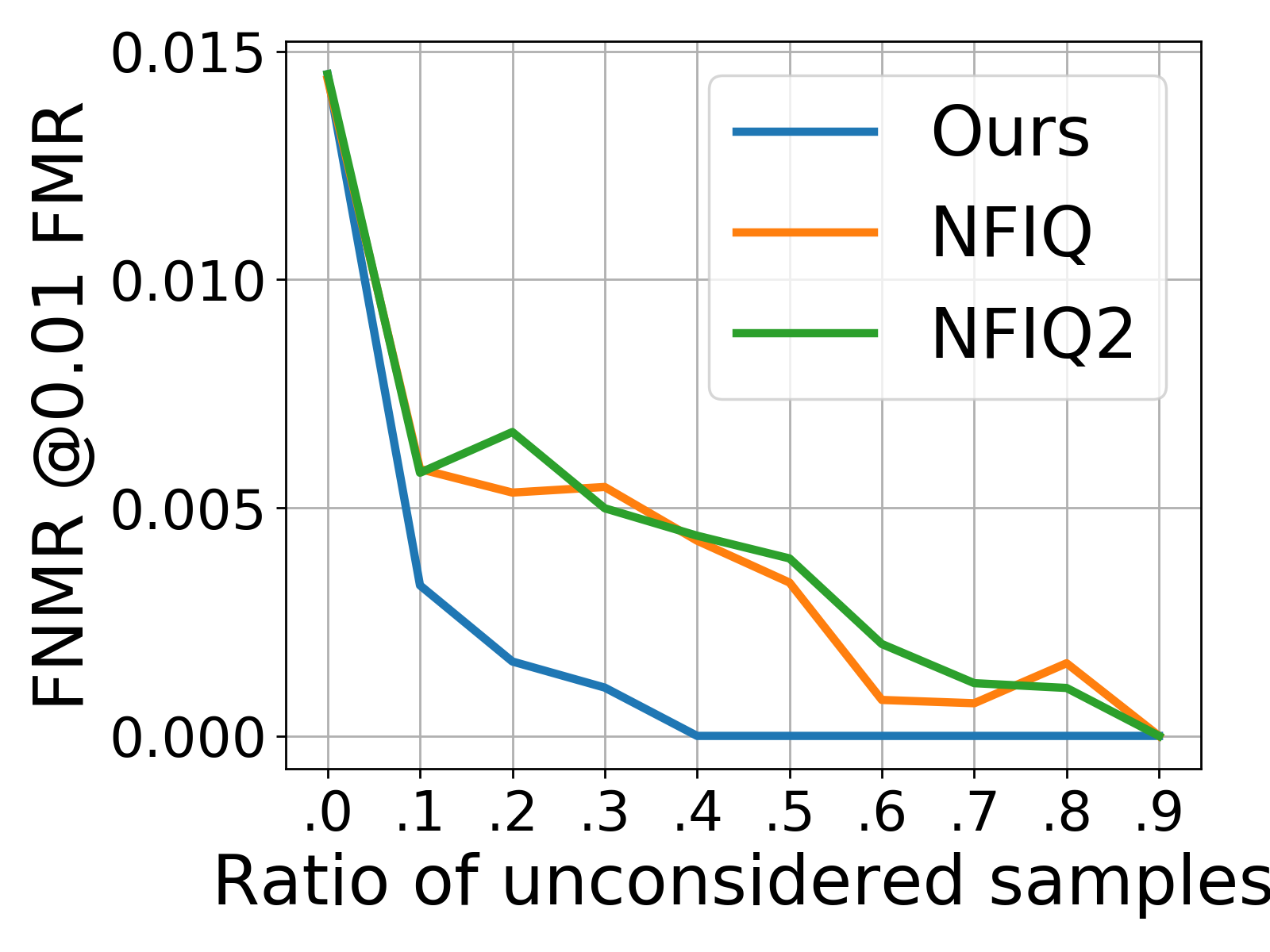}}   
\subfloat[DB3 (thermal sensor) \label{fig:FPQ_DB3_01_Bo}]{%
       \includegraphics[width=0.20\textwidth]{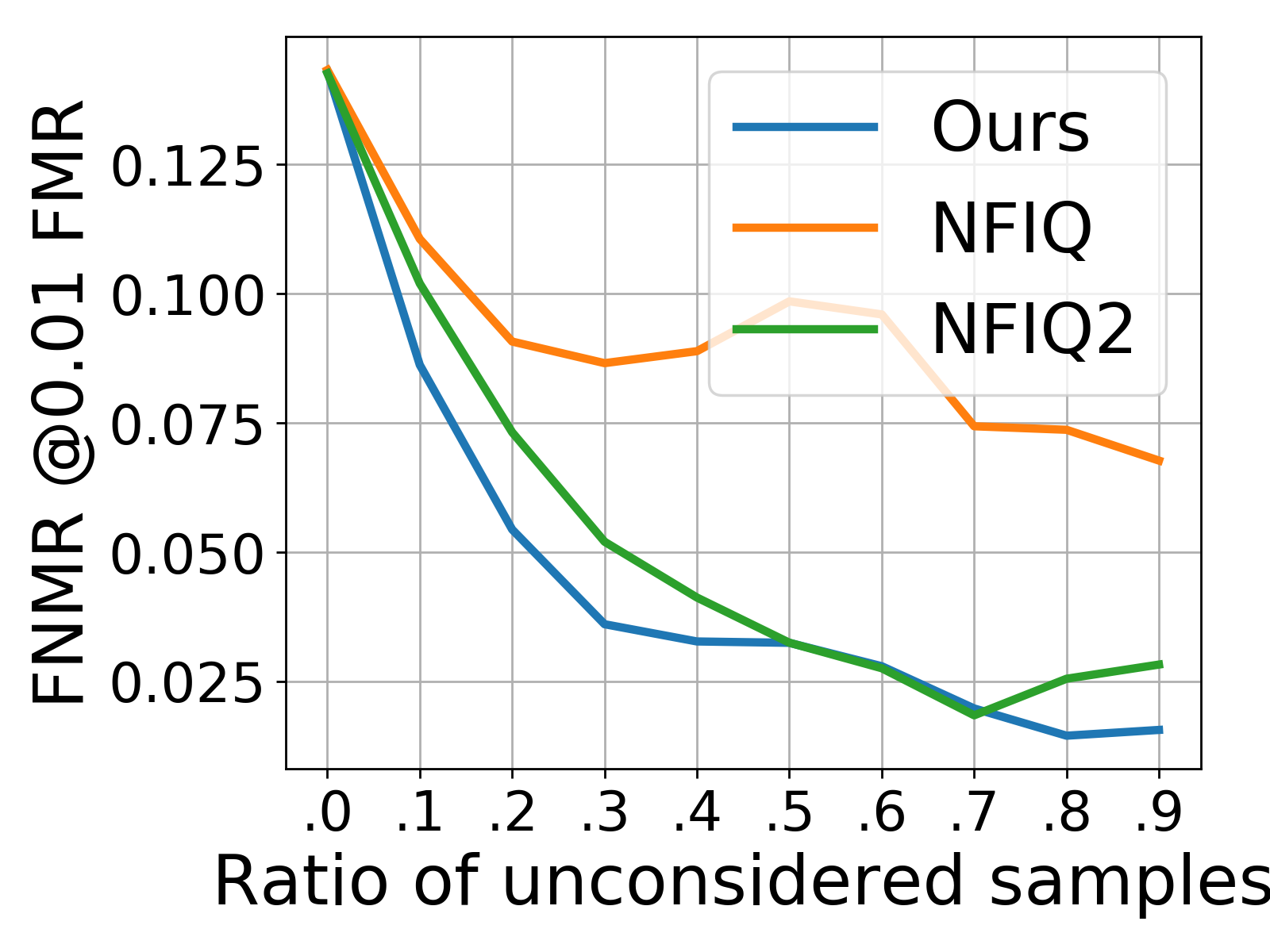}} 
\subfloat[DB4 (synthetic data) \label{fig:FPQ_DB4_01_Bo}]{%
       \includegraphics[width=0.20\textwidth]{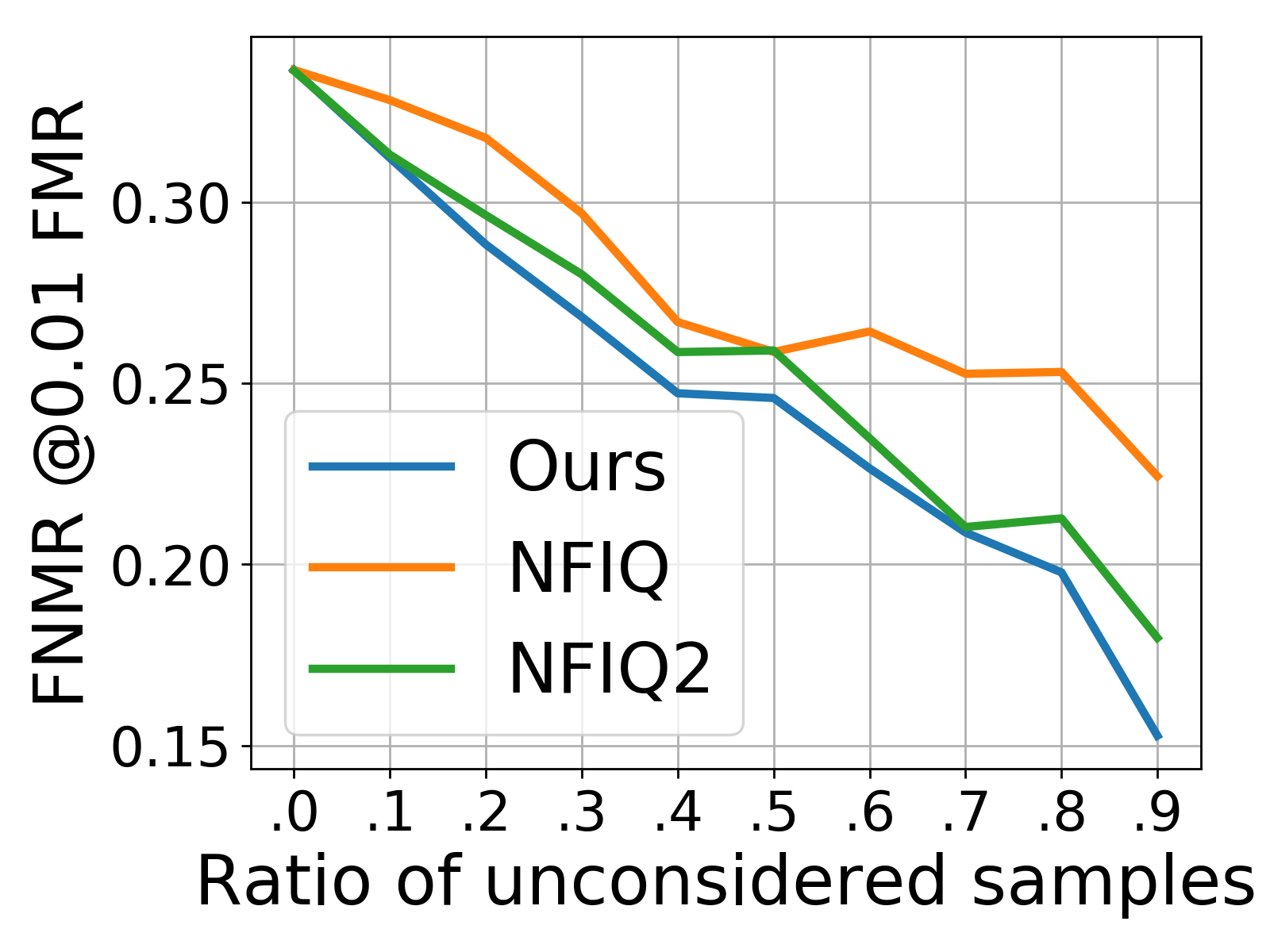}}

\subfloat[DB1 (electric field sensor)\label{fig:FPQ_DB1_001_Bo}]{%
       \includegraphics[width=0.20\textwidth]{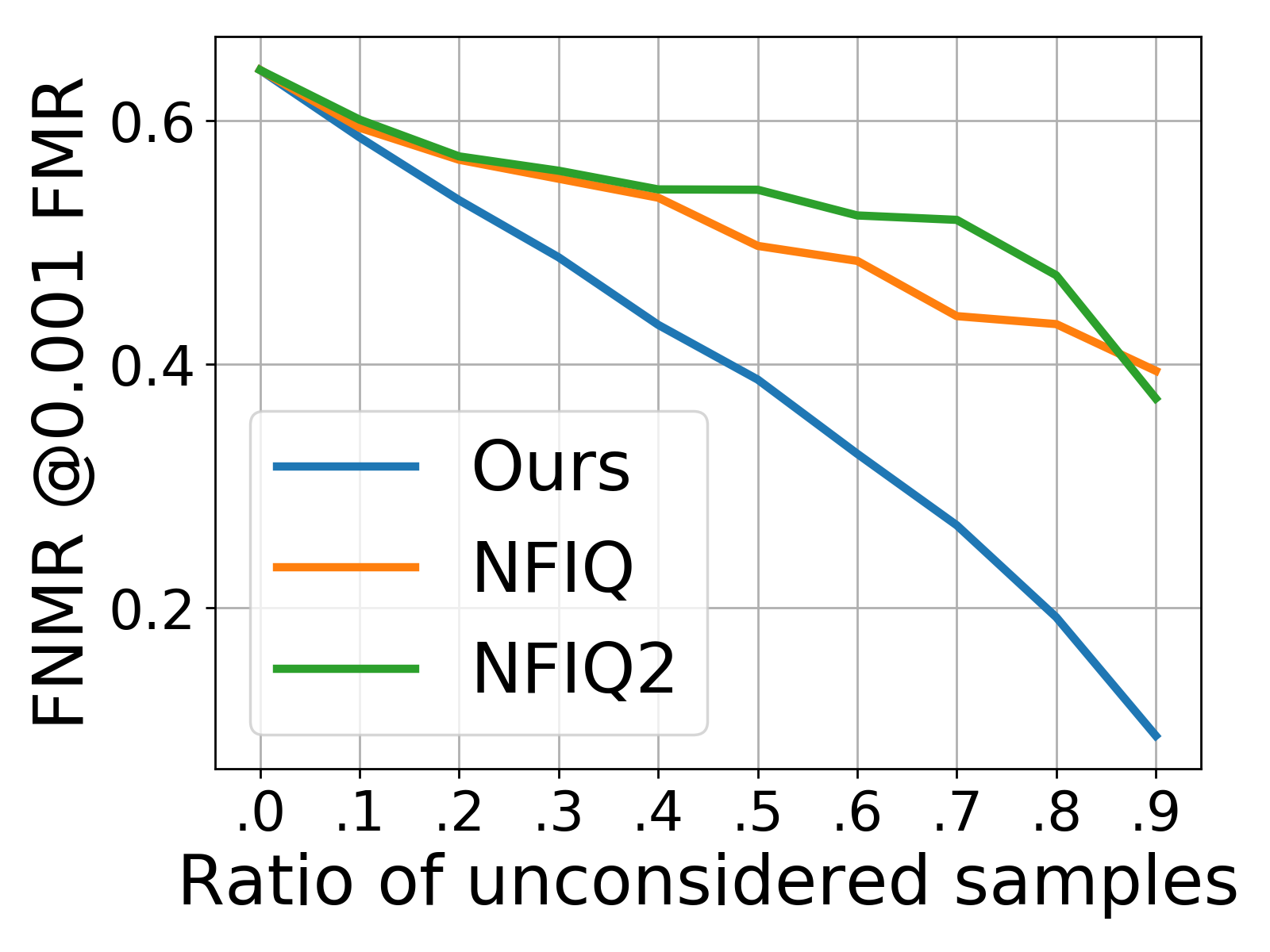}} 
\subfloat[DB2 (optical sensor) \label{fig:FPQ_DB2_001_Bo}]{%
       \includegraphics[width=0.20\textwidth]{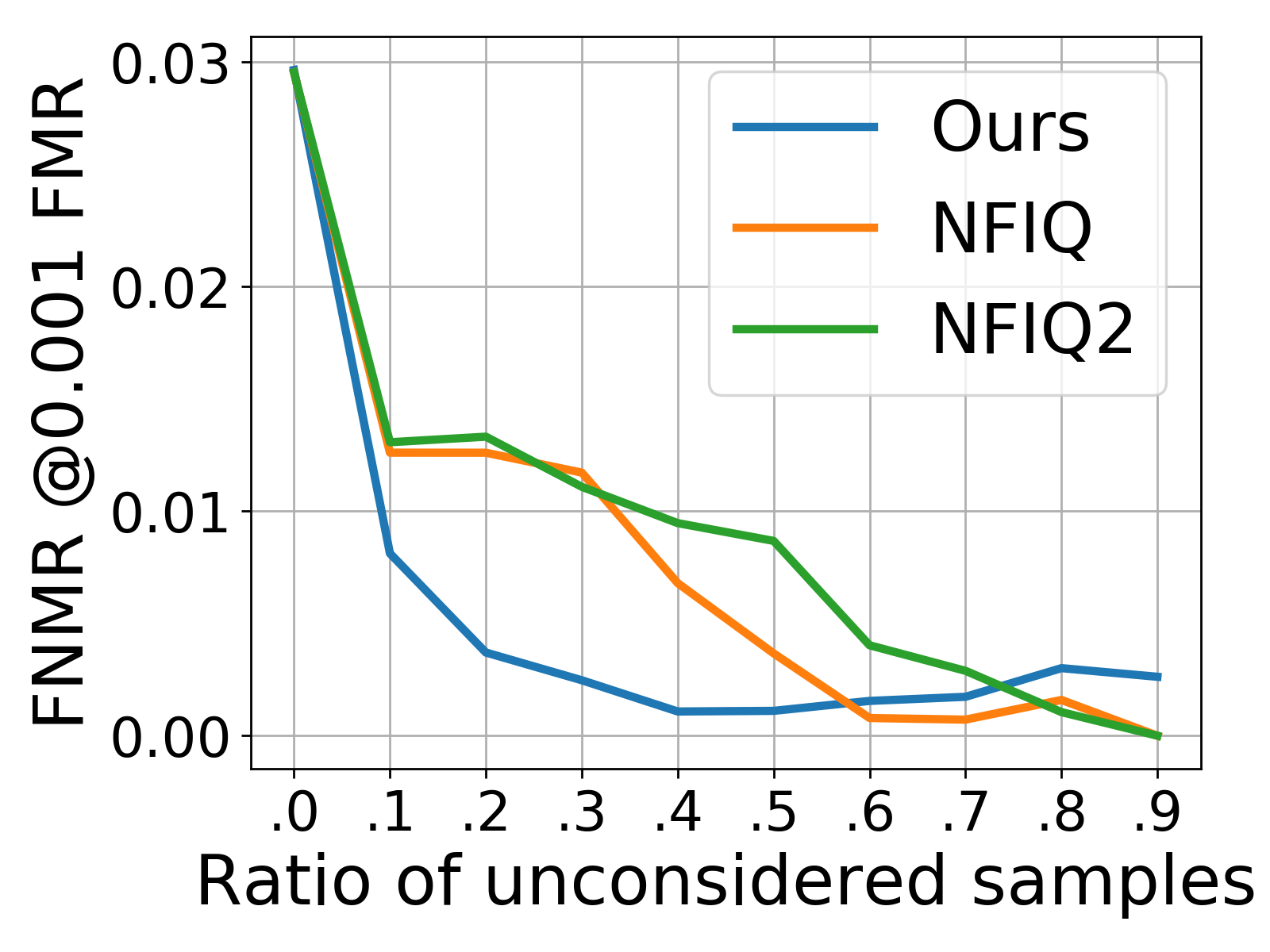}}   
\subfloat[DB3 (thermal sensor) \label{fig:FPQ_DB3_001_Bo}]{%
       \includegraphics[width=0.20\textwidth]{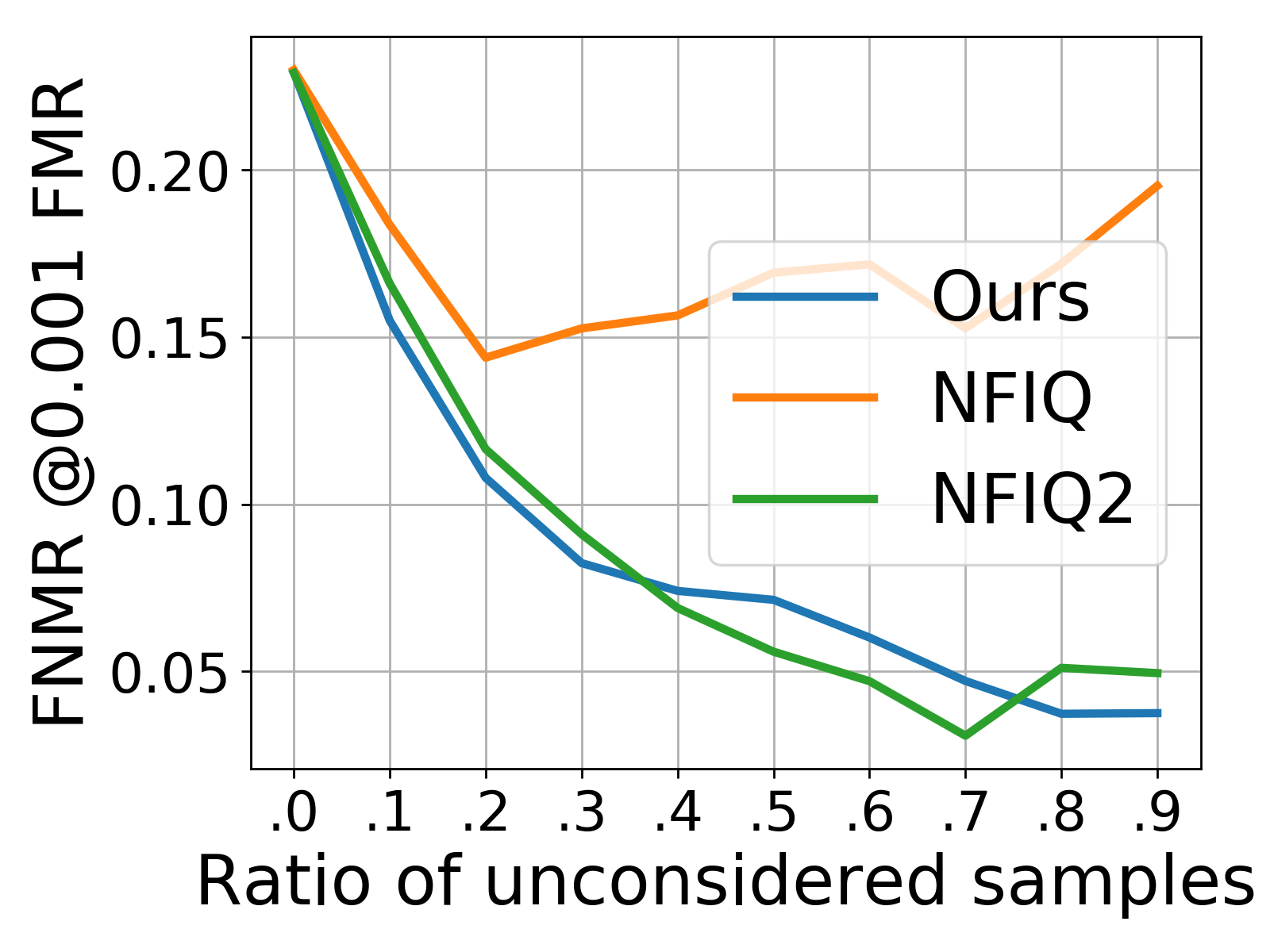}} 
\subfloat[DB4 (synthetic data) \label{fig:FPQ_DB4_001_Bo}]{%
       \includegraphics[width=0.20\textwidth]{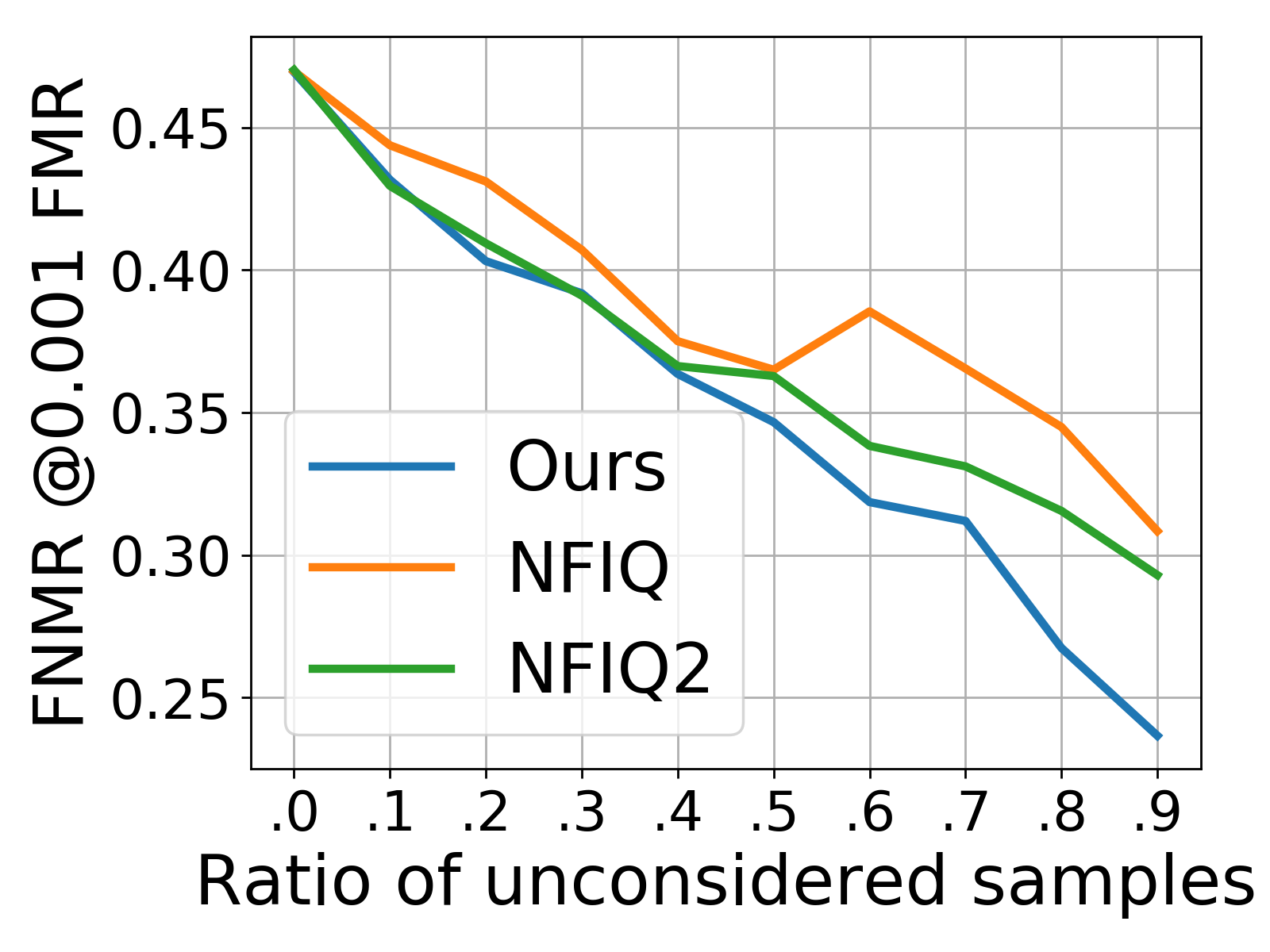}}

\caption{Fingerprint quality assessment on the Bozorth3 \cite{NBIS} matcher.
Each row represents the recognition error at a different FMR ($10^{-1}$, $10^{-2}$, and $10^{-3}$). Especially on the real-world sensor data, the proposed approach outperforms the widely-used NFIQ and NFIQ2 baselines. This holds true for all investigated sensor-types.}
\label{fig:FingerprintQuality_Bozorth3} \vspace{-3mm}
\end{figure*}

\begin{figure*}[h]
\captionsetup[subfloat]{farskip=5pt,captionskip=1pt}
\centering

\subfloat[DB1 (electric field sensor)\label{fig:FPQ_DB1_1_MCC}]{%
       \includegraphics[width=0.20\textwidth]{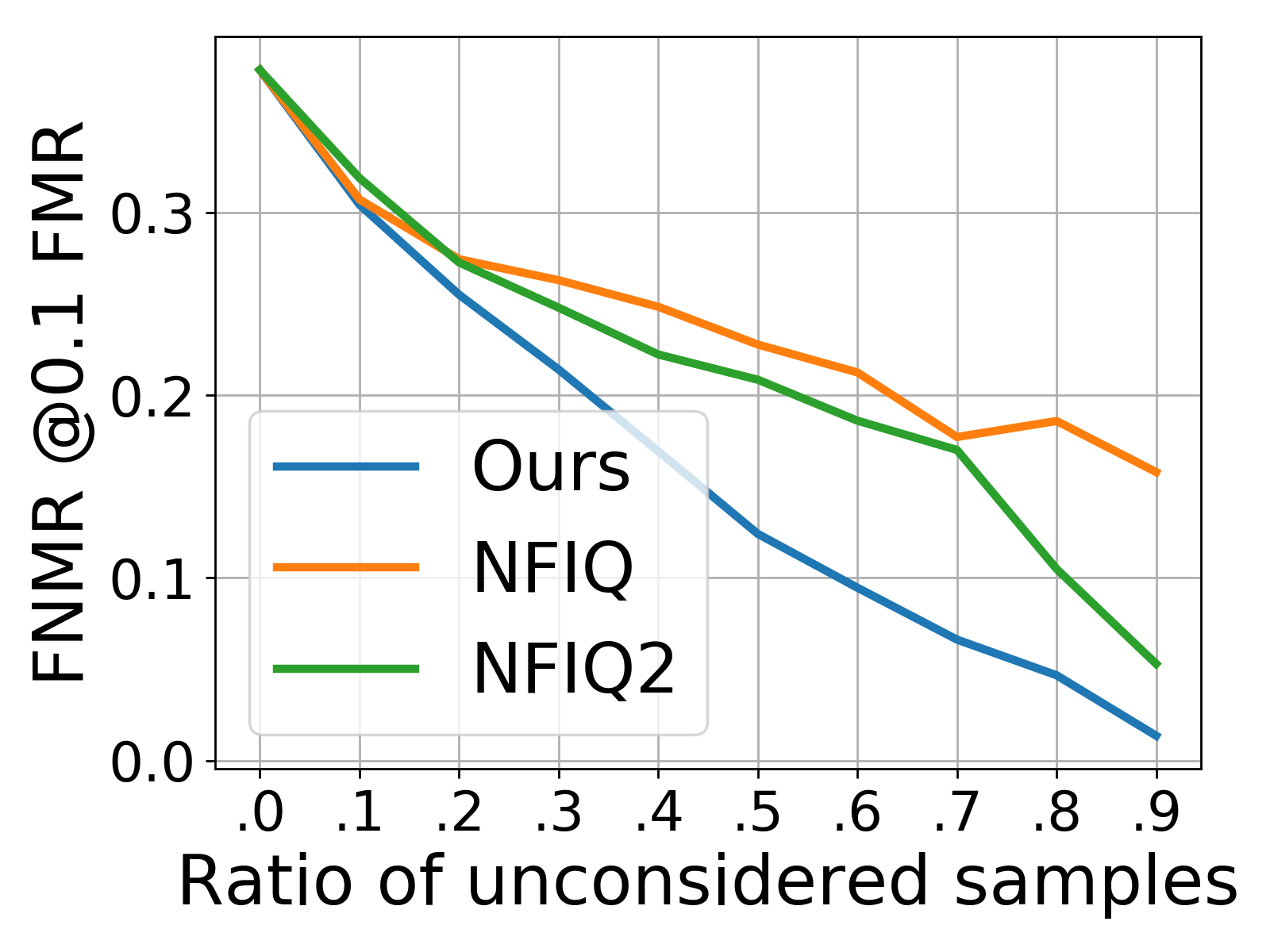}} 
\subfloat[DB2 (optical sensor) \label{fig:FPQ_DB2_1_MCC}]{%
       \includegraphics[width=0.20\textwidth]{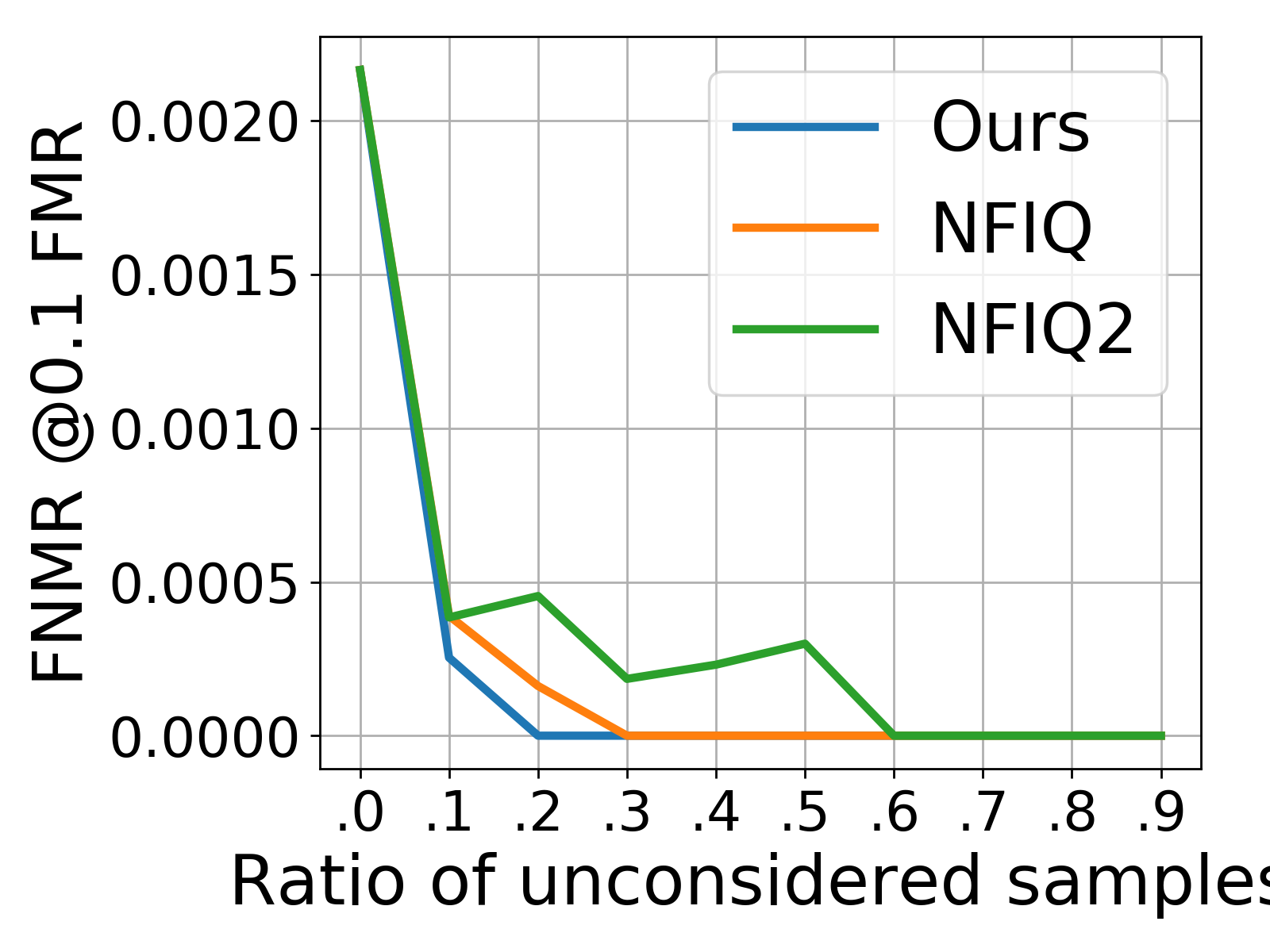}}   
\subfloat[DB3 (thermal sensor) \label{fig:FPQ_DB3_1_MCC}]{%
       \includegraphics[width=0.20\textwidth]{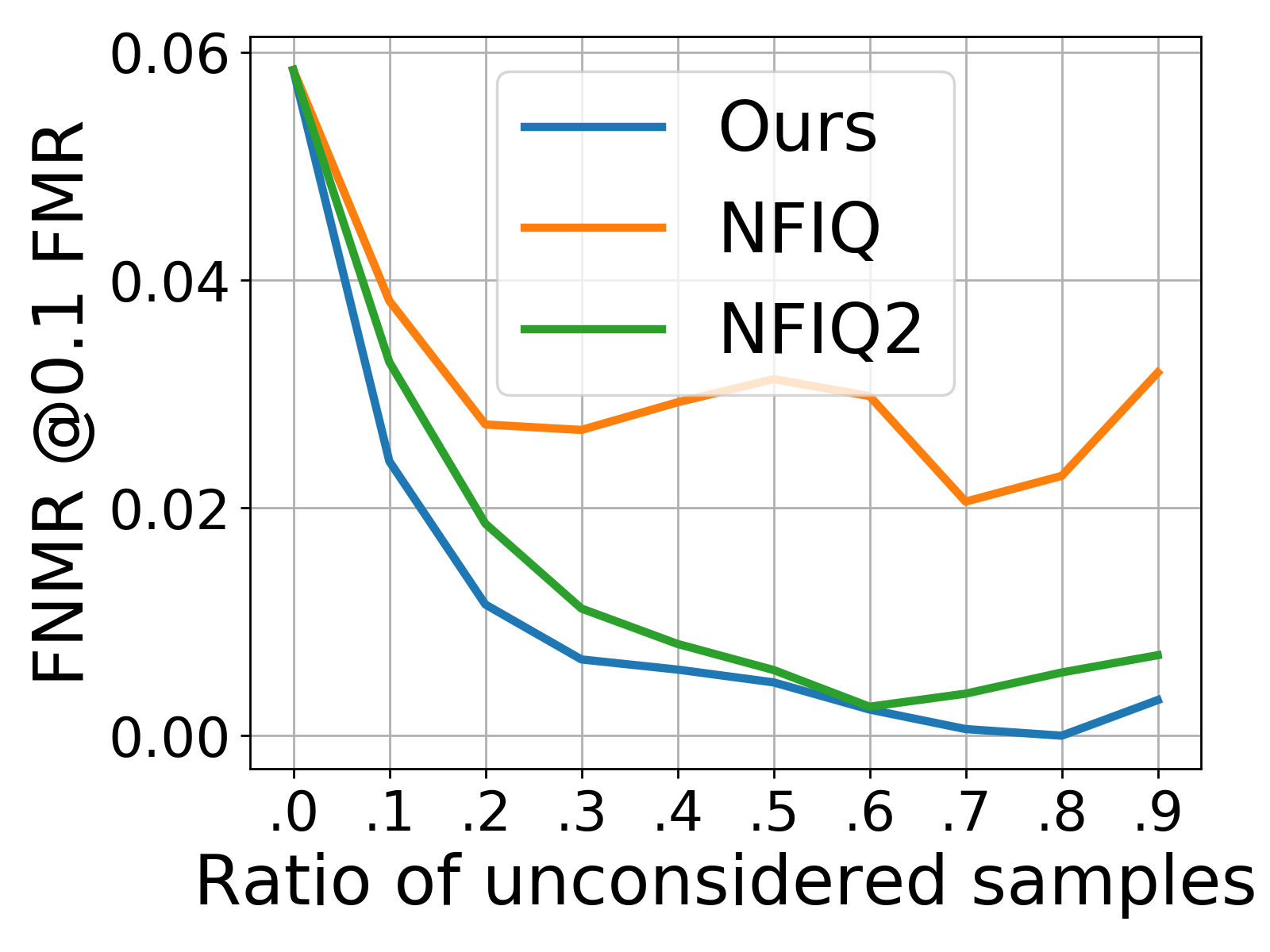}} 
\subfloat[DB4 (synthetic data) \label{fig:FPQ_DB4_1_MCC}]{%
       \includegraphics[width=0.20\textwidth]{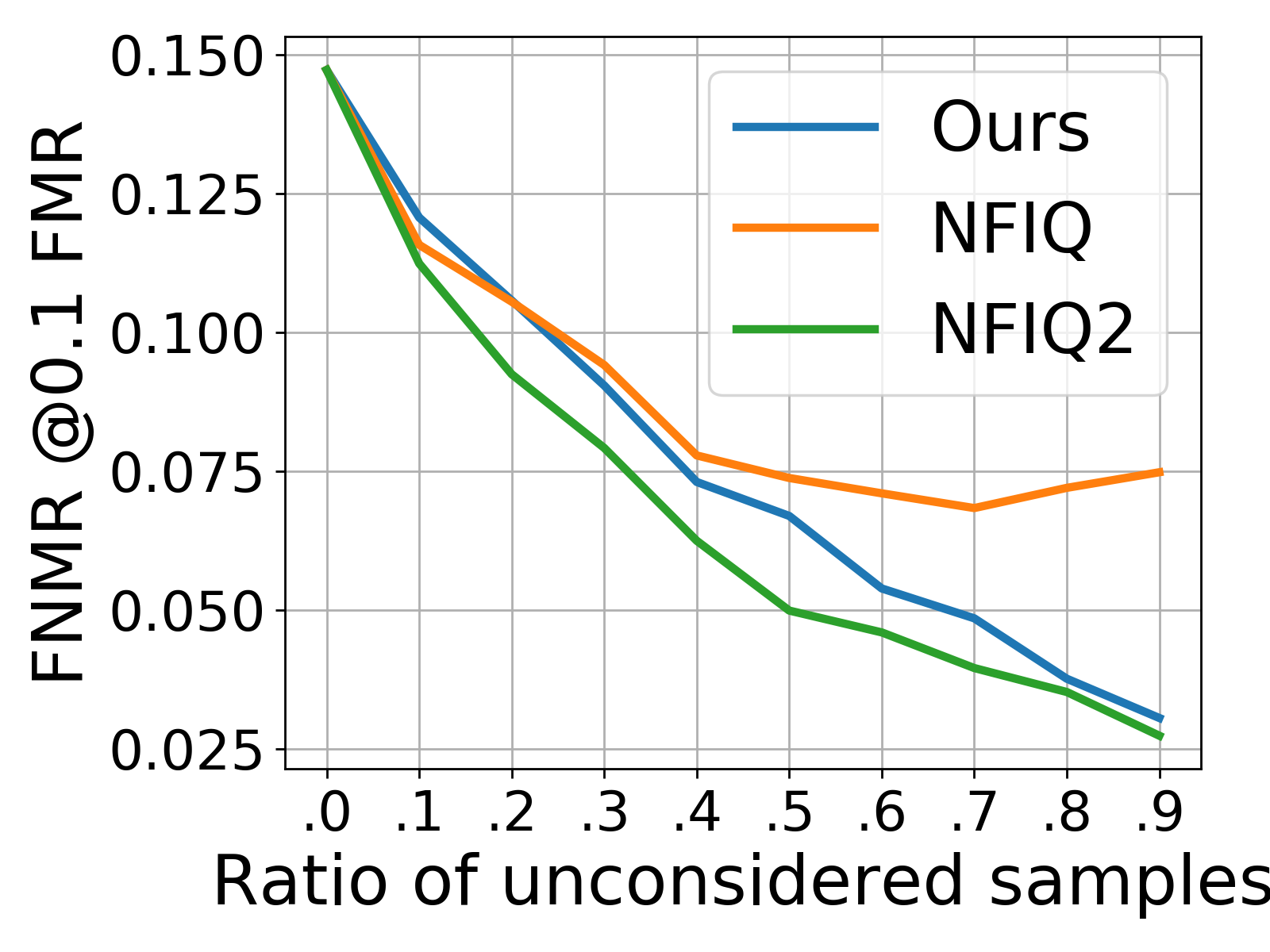}}
       
\subfloat[DB1 (electric field sensor)\label{fig:FPQ_DB1_01_MCC}]{%
       \includegraphics[width=0.20\textwidth]{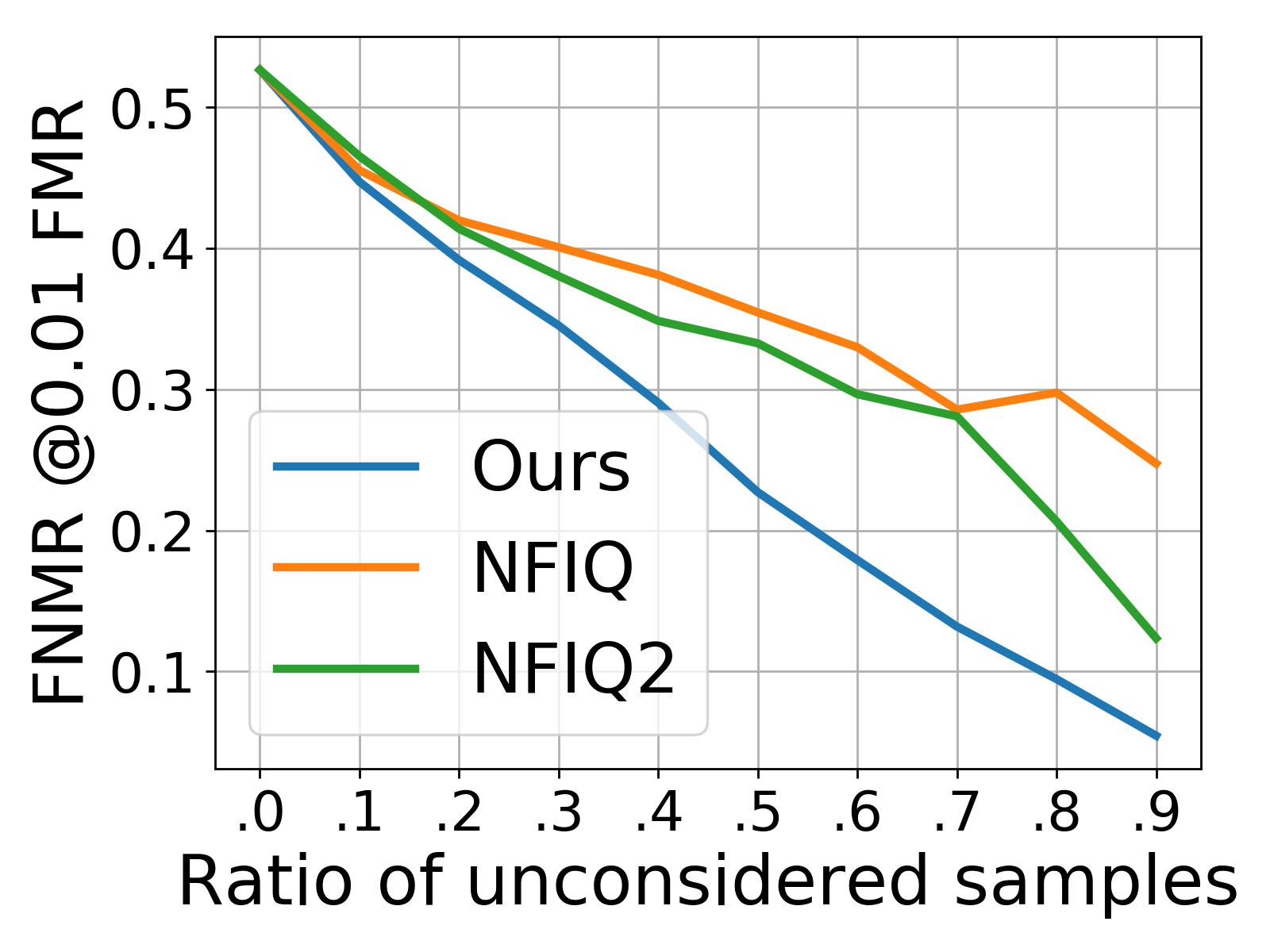}} 
\subfloat[DB2 (optical sensor) \label{fig:FPQ_DB2_01_MCC}]{%
       \includegraphics[width=0.20\textwidth]{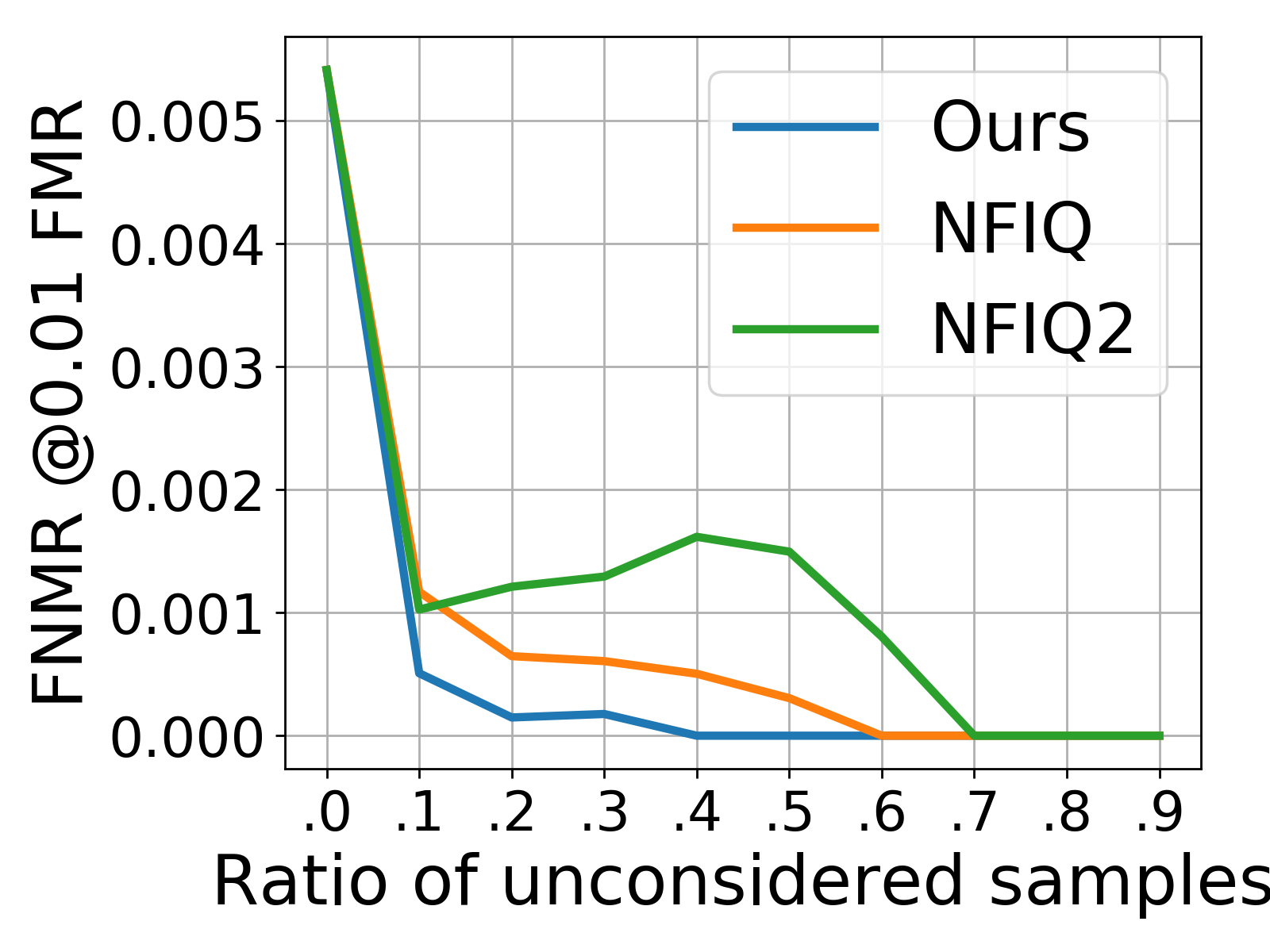}}   
\subfloat[DB3 (thermal sensor) \label{fig:FPQ_DB3_01_MCC}]{%
       \includegraphics[width=0.20\textwidth]{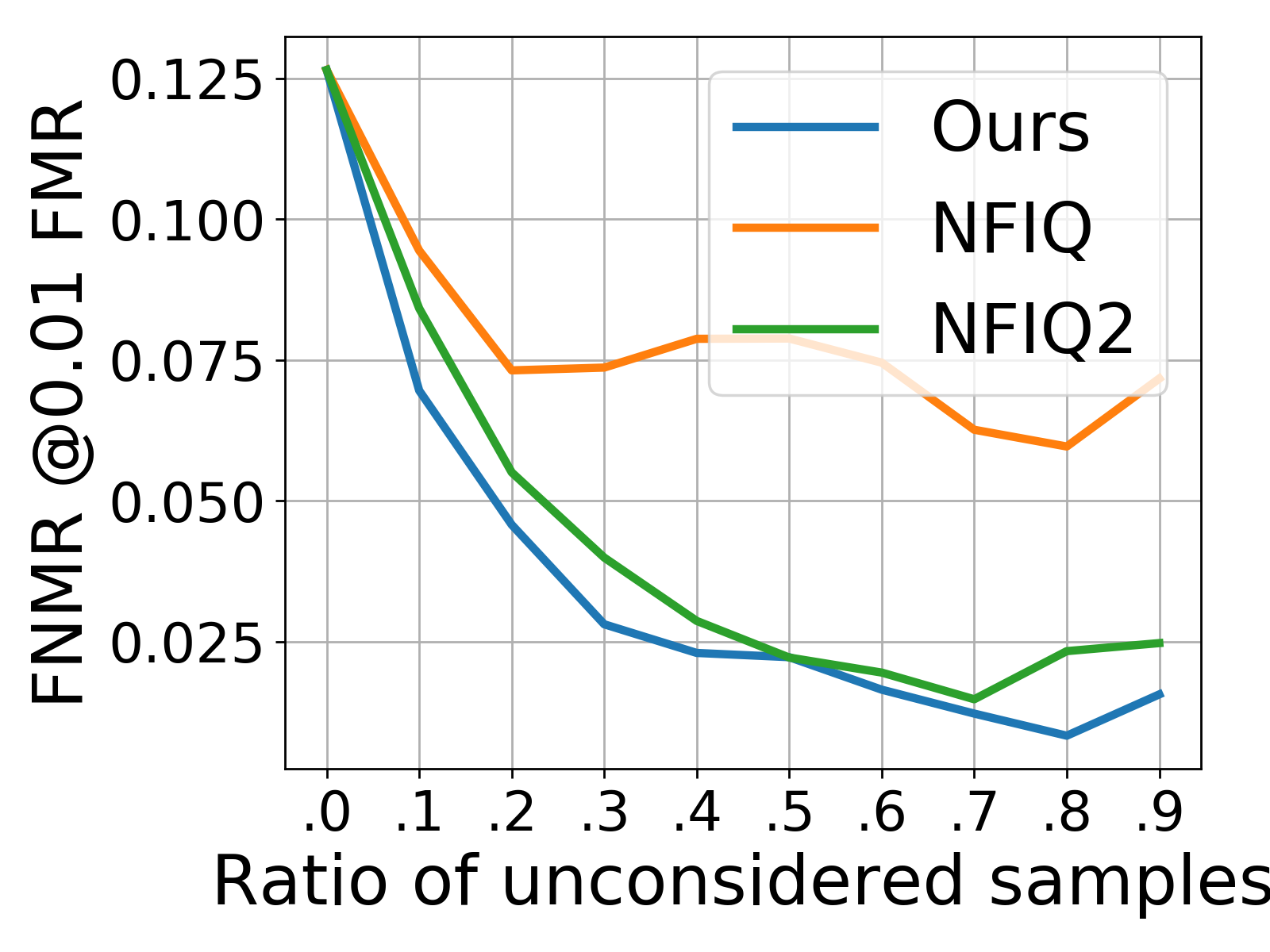}} 
\subfloat[DB4 (synthetic data) \label{fig:FPQ_DB4_01_MCC}]{%
       \includegraphics[width=0.20\textwidth]{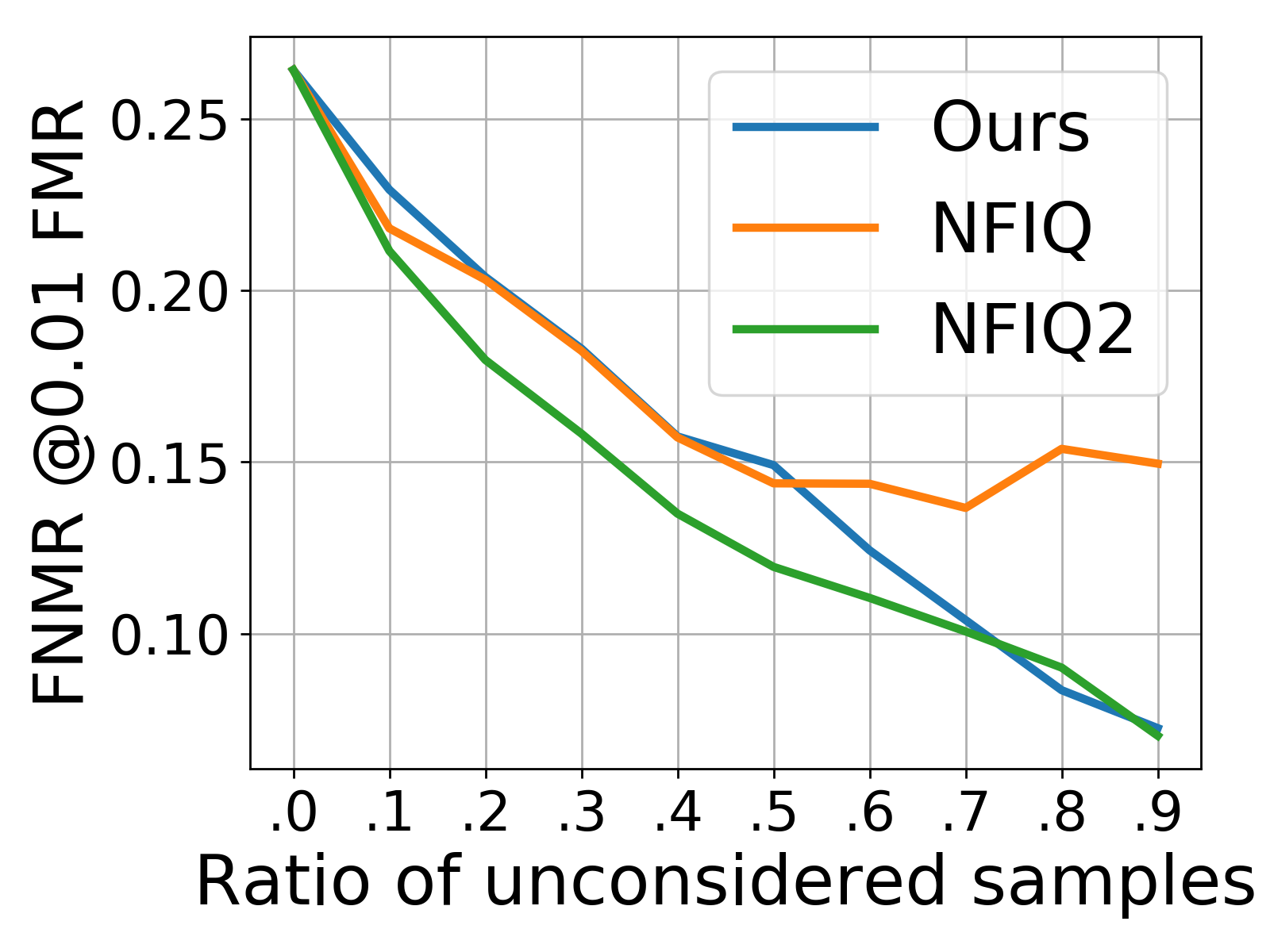}}

\subfloat[DB1 (electric field sensor)\label{fig:FPQ_DB1_001_MCC}]{%
       \includegraphics[width=0.20\textwidth]{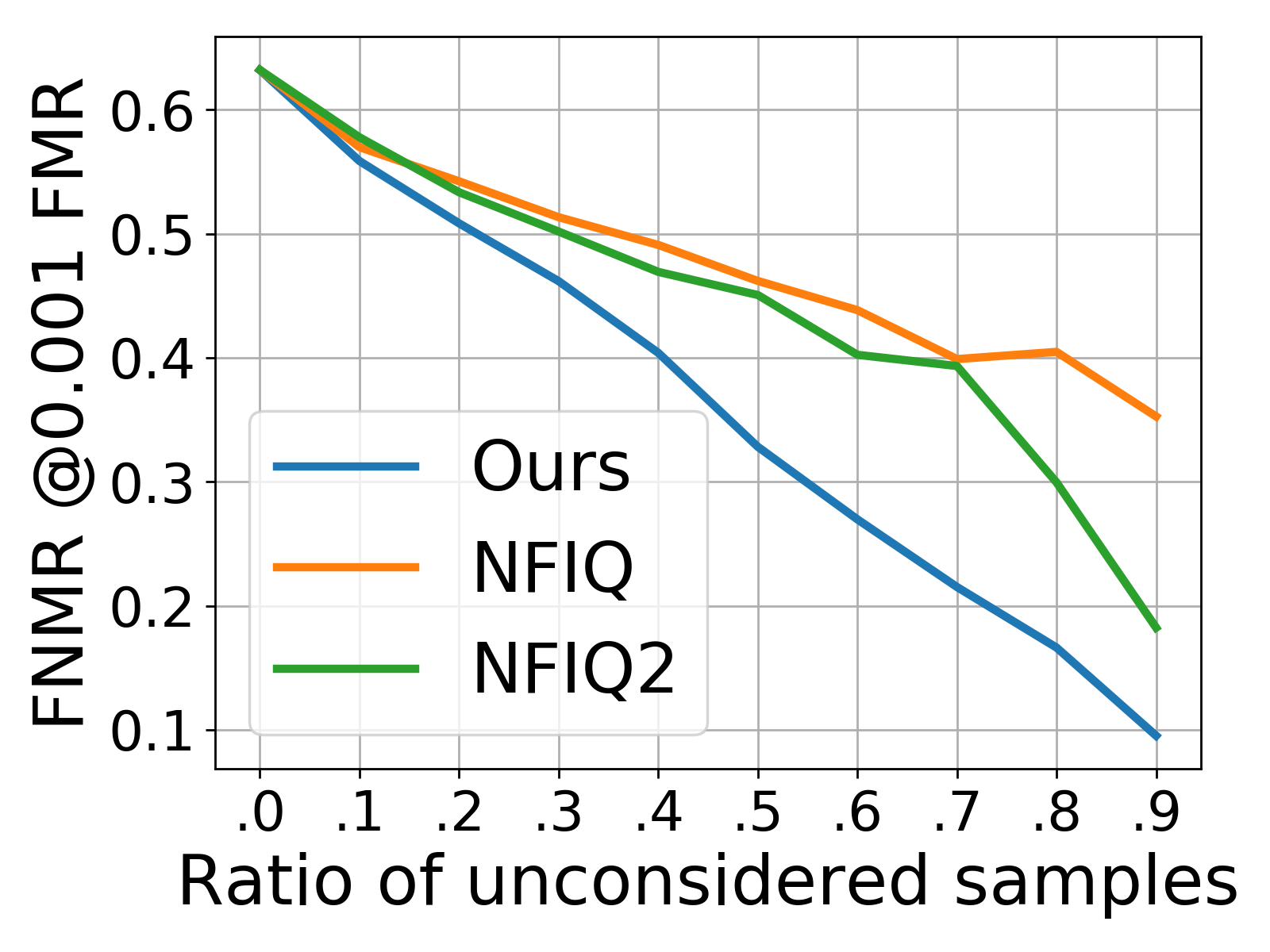}} 
\subfloat[DB2 (optical sensor) \label{fig:FPQ_DB2_001_MCC}]{%
       \includegraphics[width=0.20\textwidth]{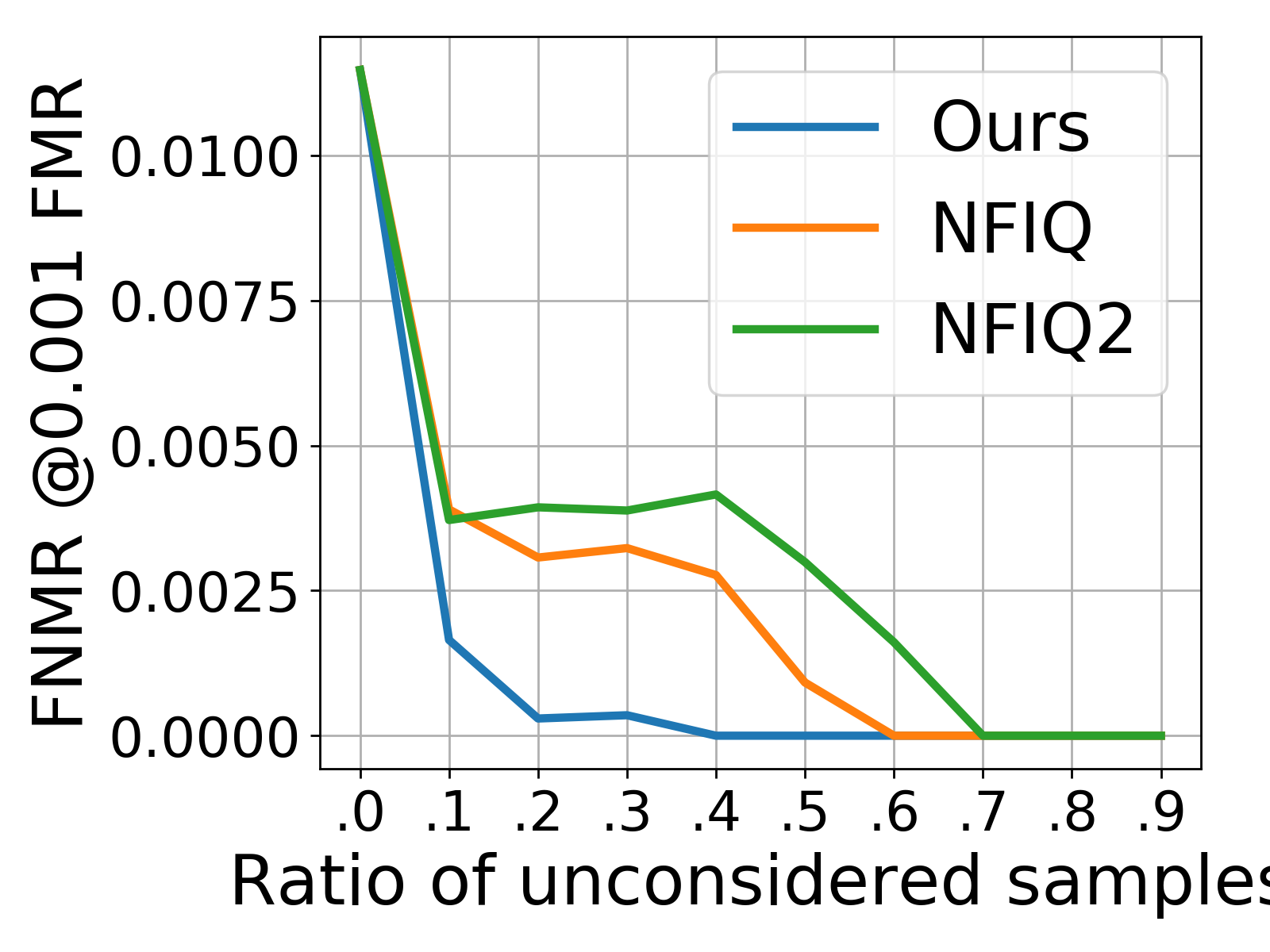}}   
\subfloat[DB3 (thermal sensor) \label{fig:FPQ_DB3_001_MCC}]{%
       \includegraphics[width=0.20\textwidth]{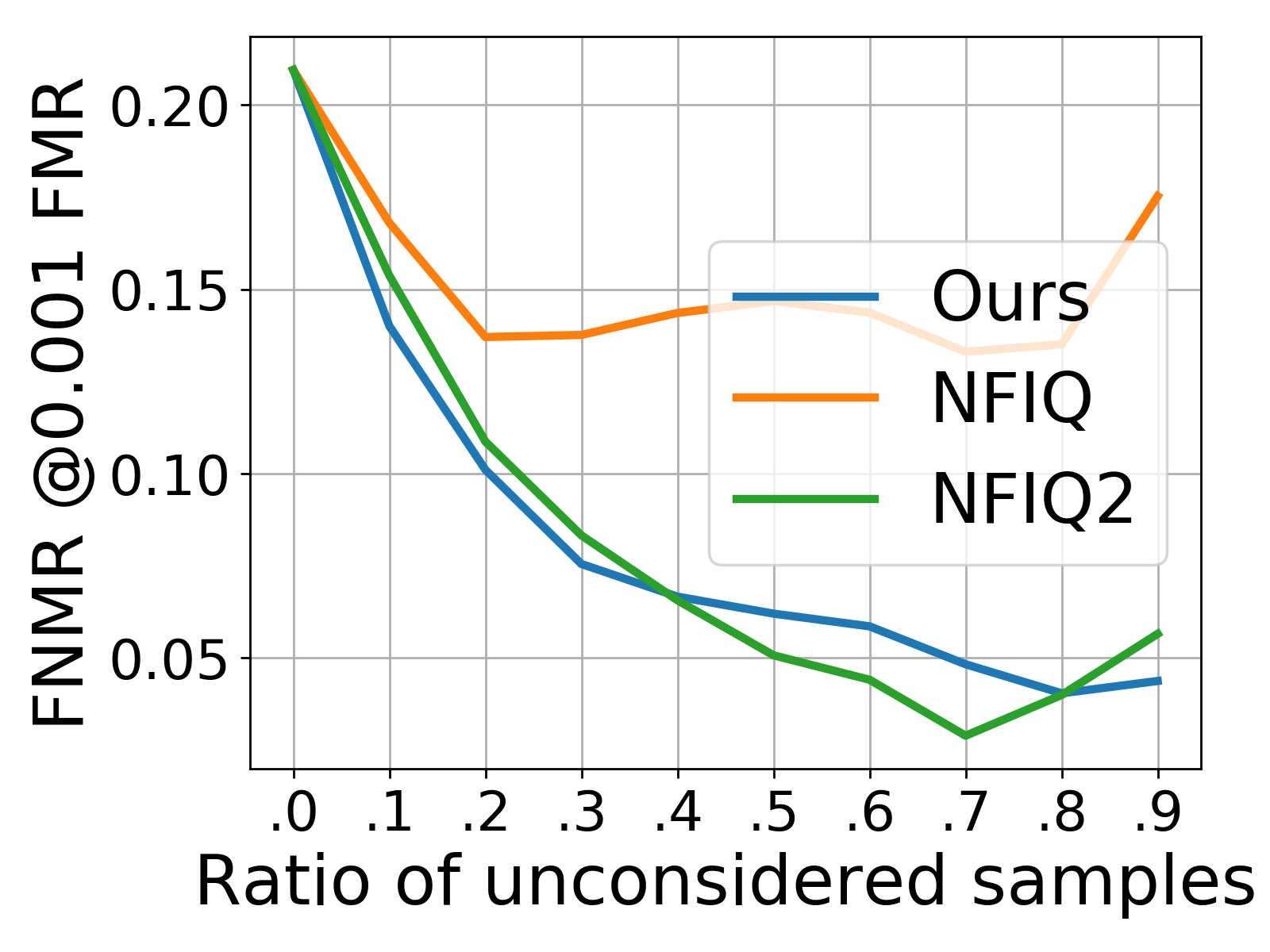}} 
\subfloat[DB4 (synthetic data) \label{fig:FPQ_DB4_001_MCC}]{%
       \includegraphics[width=0.20\textwidth]{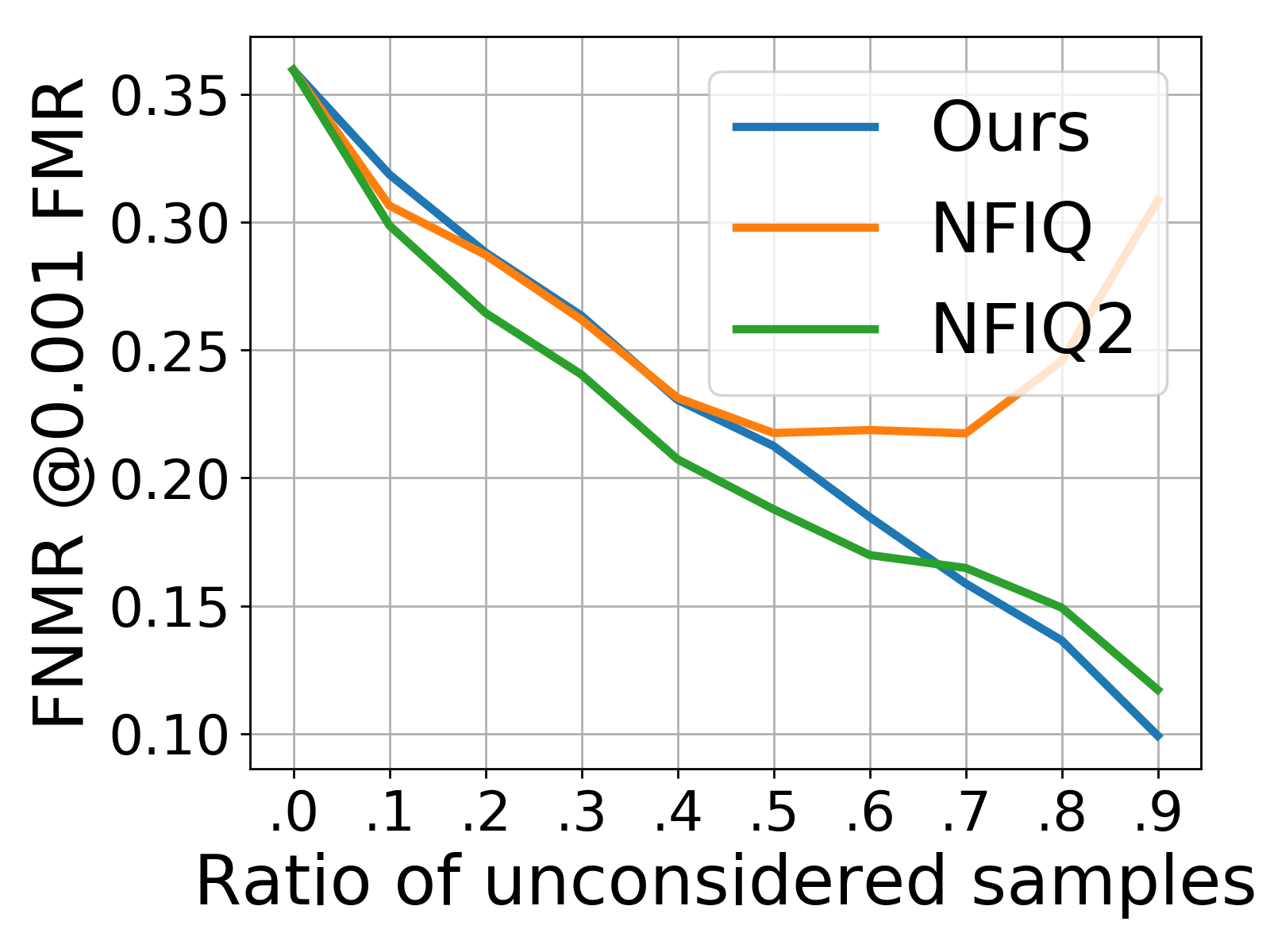}}

\caption{Fingerprint quality assessment on the MCC \cite{DBLP:journals/pami/CappelliFM10} matcher.
Each row represents the recognition error at a different FMR ($10^{-1}$, $10^{-2}$, and $10^{-3}$). Especially on the real-world sensor data, the proposed approach outperforms the widely-used NFIQ and NFIQ2 baselines. This holds true for all investigated sensor-types.}
\label{fig:FingerprintQualityMCC} \vspace{-3mm}
\end{figure*}

Figures \ref{fig:FingerprintQuality_Bozorth3} and \ref{fig:FingerprintQualityMCC} compare the quality assessment performance of our proposed approach with NFIQ and NFIQ2 on the Bozorth3 (Figure \ref{fig:FingerprintQuality_Bozorth3}) and MCC (Figure \ref{fig:FingerprintQualityMCC}) matchers.
The quality assessment performance is shown as error-vs-reject curves as described in Section \ref{sec:EvaluationMetrics}.
The corresponding verification error is shown in terms of FNMR at an FMR of $10^{-1}$ (a-d), $10^{-2}$ (e-h), and $10^{-3}$ (i-l).
Moreover, the effectiveness of the different quality assessment solutions is analysed on three different sensor types (electric field, optical, and thermal) and synthetic data.

In all investigated scenarios, all quality assessment solutions show a solid and (relatively) stable performance.
Our proposed method based on the combination of minutia detection confidences outperforms the NIST's NFIQ and NFIQ2 in all cases involving real fingerprints.
Just on the less-relevant synthetically generated data using the MCC matcher, NFIQ2 shows better quality estimates.
However, on data captured from real fingerprints, the proposed methodology constantly shows a significantly stronger quality assessment performance than the baselines.
Especially for electric field and optical sensors, the proposed approach outperforms the baselines by a large margin.

\vspace{-1mm}
\section{Conclusion}
\vspace{-1mm}

Previous works on the quality assessment of single minutiae or full fingerprint captures are either learning models based on error-prone quality labels or make use of naive hand-crafted features.
Although many fingerprint recognition systems are still based on minutiae information, previous works on quality assessment neglected the minutia extraction process.
In this work, we proposed a novel concept of measuring the quality of fingerprints and single minutiae based on an arbitrary minutia extraction network.
The proposed method, MiDeCon, uses the minutia detection confidence of the utilized minutia extractor as the quality indicator for this minutia.
This confidence is based on the agreement of stochastic variants of the extraction network.
By combining the highest minutia qualities of a fingerprint, it successfully determines the fingerprint quality of this capture.
Experiments on the publicly available FVC 2006 databases demonstrate that the proposed approach strongly outperforms the baselines, such as Mindtct and CoarseNet for quality assessment of single minutia and NIST's widely-used fingerprint image software NFIQ1 and NFIQ2 for fingerprint quality assessment.
The strong performance of MiDeCon is especially observed when facing real-world sensor data, such as electric field, optical, and thermal sensors. 
In contrast to previous works, MiDeCon jointly (a) does not rely on quality labels for training, (b) produces continuous quality values, (c) considers difficulties in the minutiae extraction process, and (d) assesses the quality of both, full fingerprints and single minutia.


\vspace{-3mm}
\paragraph{Acknowledgement} 
This research work has been funded by the German Federal Ministry of Education and Research and the Hessen State Ministry for Higher Education, Research and the Arts within their joint support of the National Research Center for Applied Cybersecurity ATHENE.

{\small
\bibliographystyle{ieee}
\bibliography{egbib_short}
}

\newpage
\clearpage

\section*{Supplementary}

In this section, we show an ablation study on our proposed approach that did not fit in the main paper due to the conference page-limit.
First, we show that both components for the FIQ estimation are needed for a stable and accurate quality assessment.
Second, we demonstrate that the proposed approach is very stable concerning the choice of the number of considered minutia $n$.

\section*{Analysing the different components of the MiDeCon quality}

\begin{figure*}[]
\captionsetup[subfloat]{farskip=5pt,captionskip=1pt}
\centering

\subfloat[DB1 (electric field sensor)\label{fig:Components_DB1_1_Bo}]{%
       \includegraphics[width=0.20\textwidth]{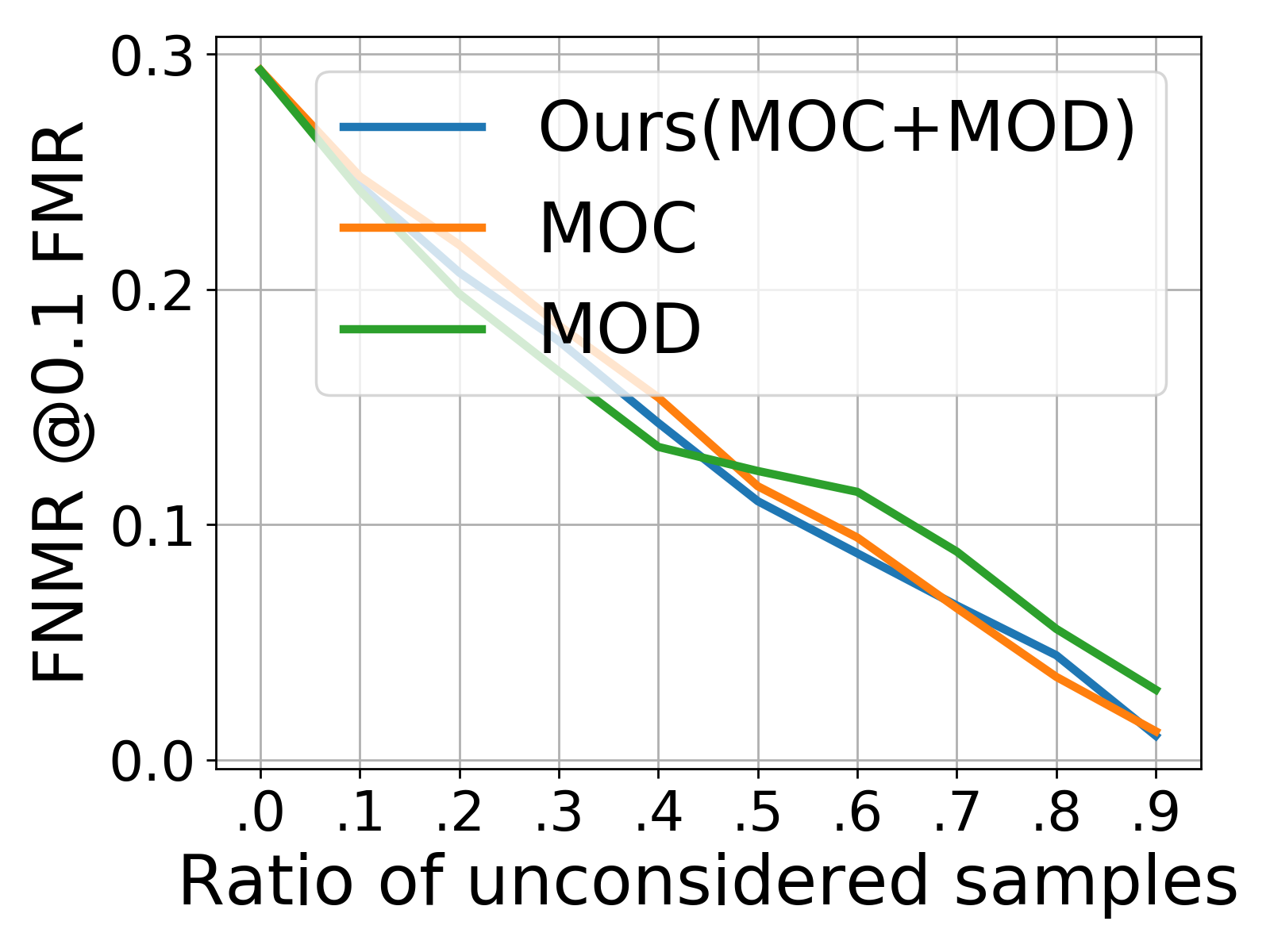}} 
\subfloat[DB2 (optical sensor)\label{fig:Components_DB2_1_Bo}]{%
       \includegraphics[width=0.20\textwidth]{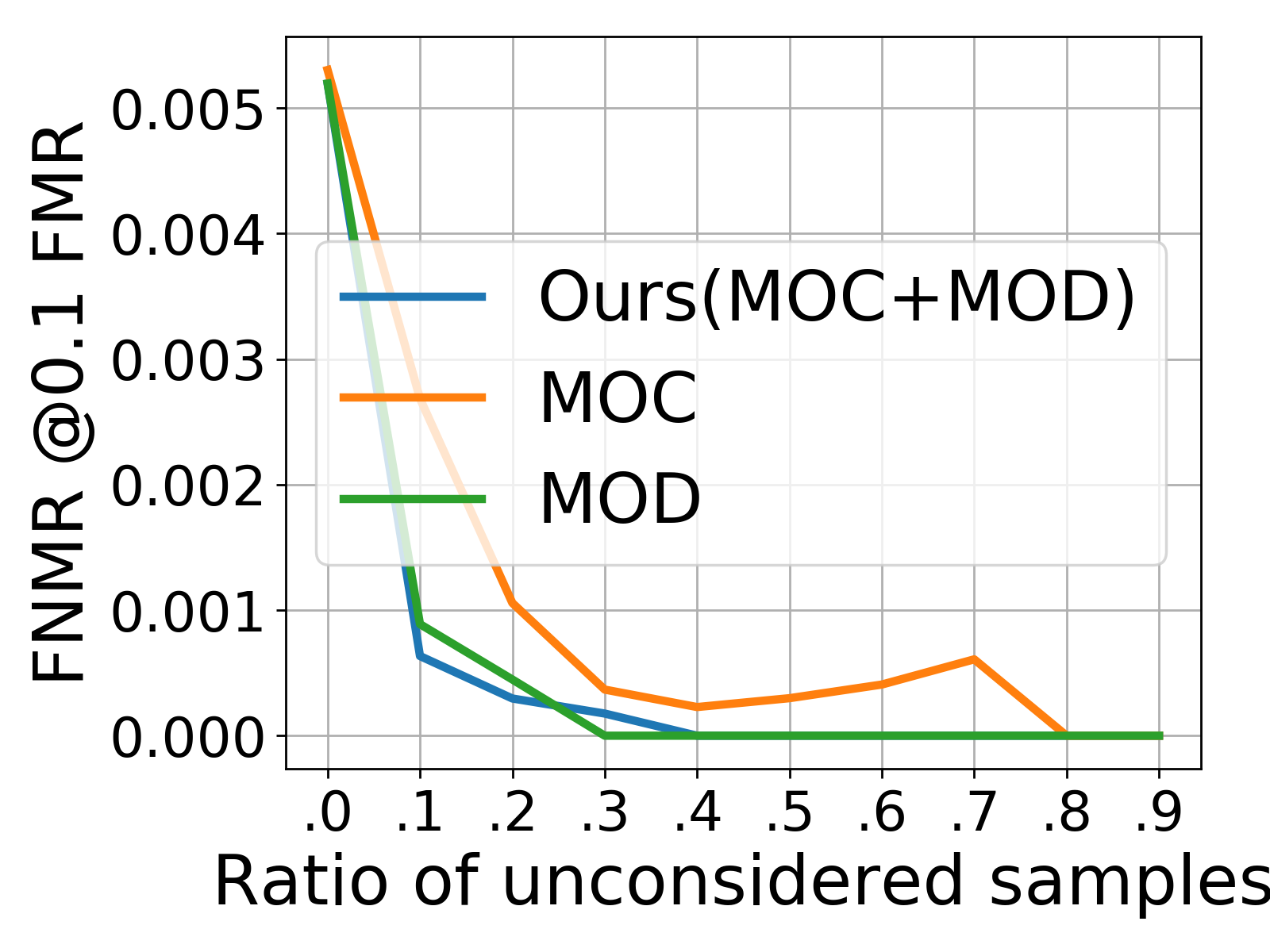}}    
\subfloat[DB3 (thermal sensor)\label{fig:Components_DB3_1_Bo}]{%
       \includegraphics[width=0.20\textwidth]{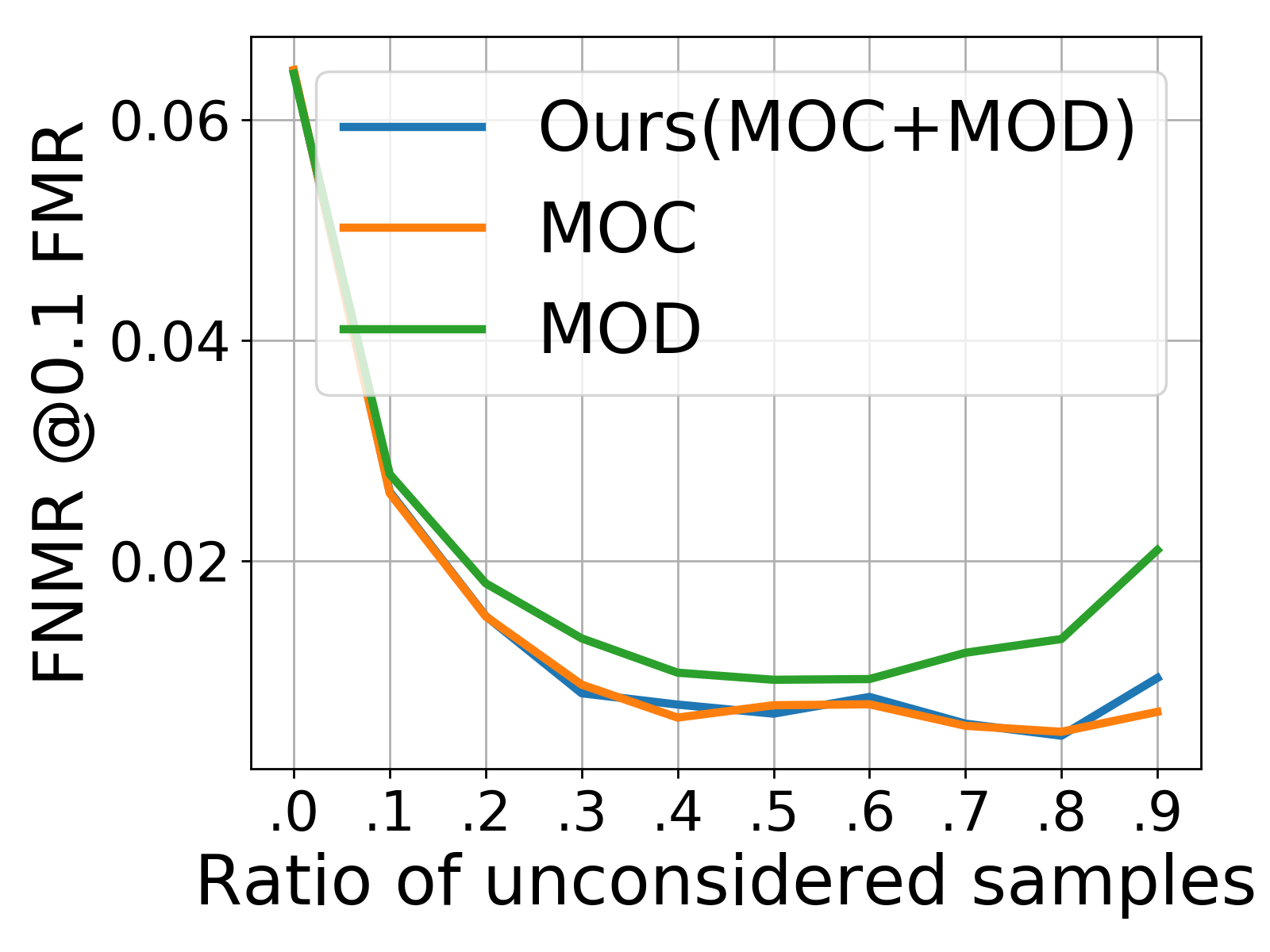}} 
\subfloat[DB4 (synthetic data)\label{fig:Components_DB4_1_Bo}]{%
       \includegraphics[width=0.20\textwidth]{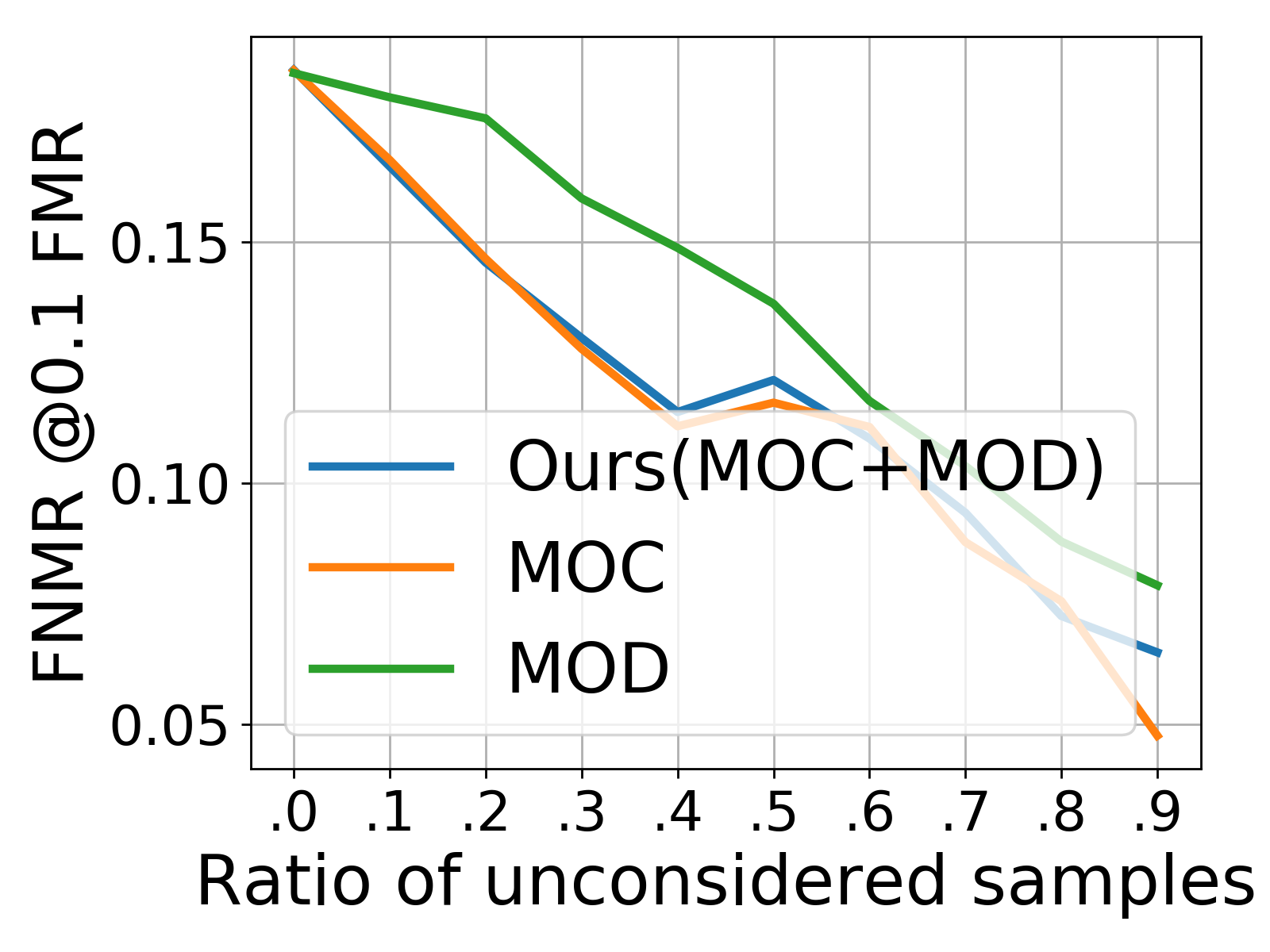}} 
               
\subfloat[DB1 (electric field sensor)\label{fig:Components_DB1_01_Bo}]{%
       \includegraphics[width=0.20\textwidth]{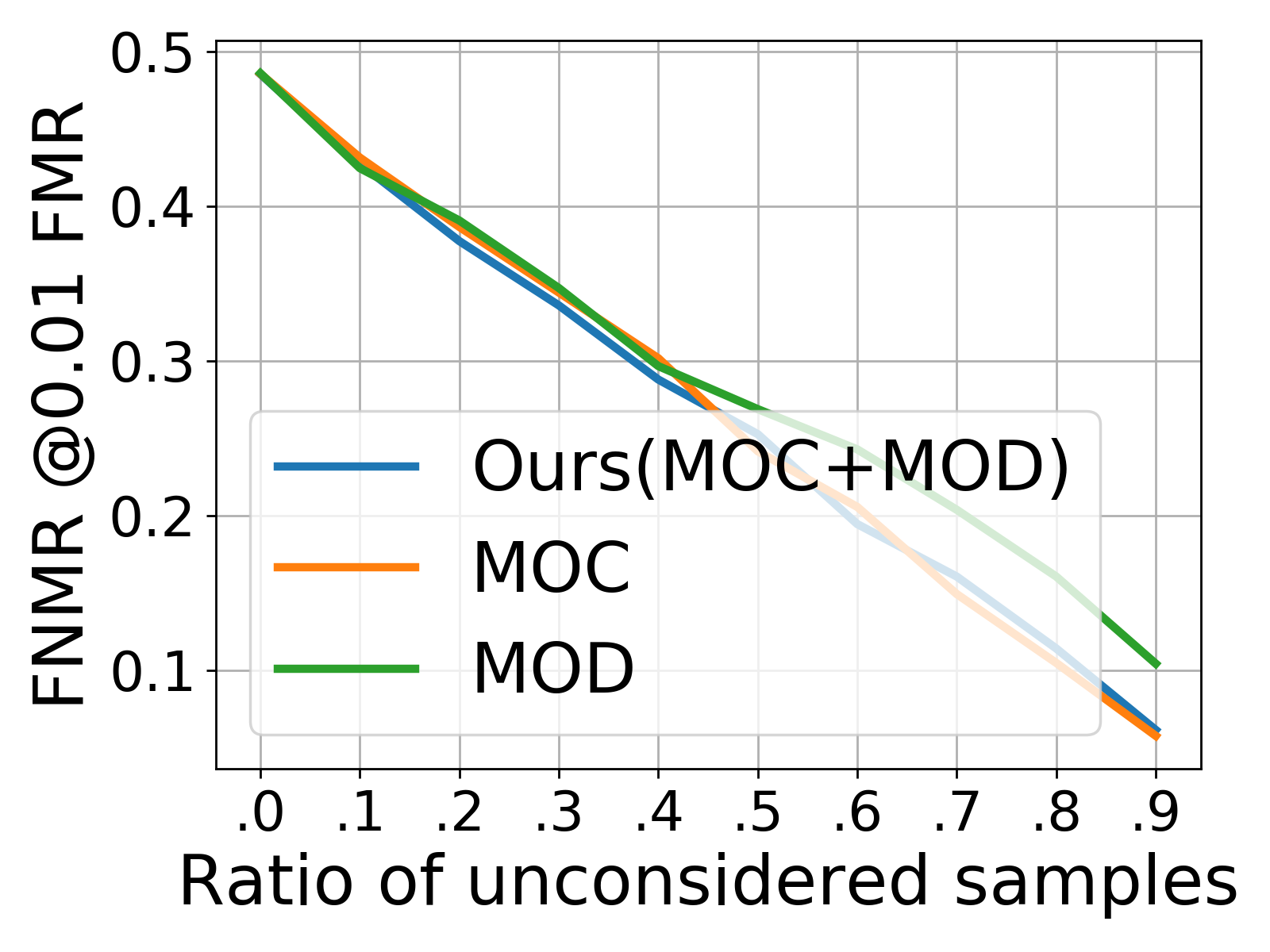}} 
\subfloat[DB2 (optical sensor)\label{fig:Components_DB2_01_Bo}]{%
       \includegraphics[width=0.20\textwidth]{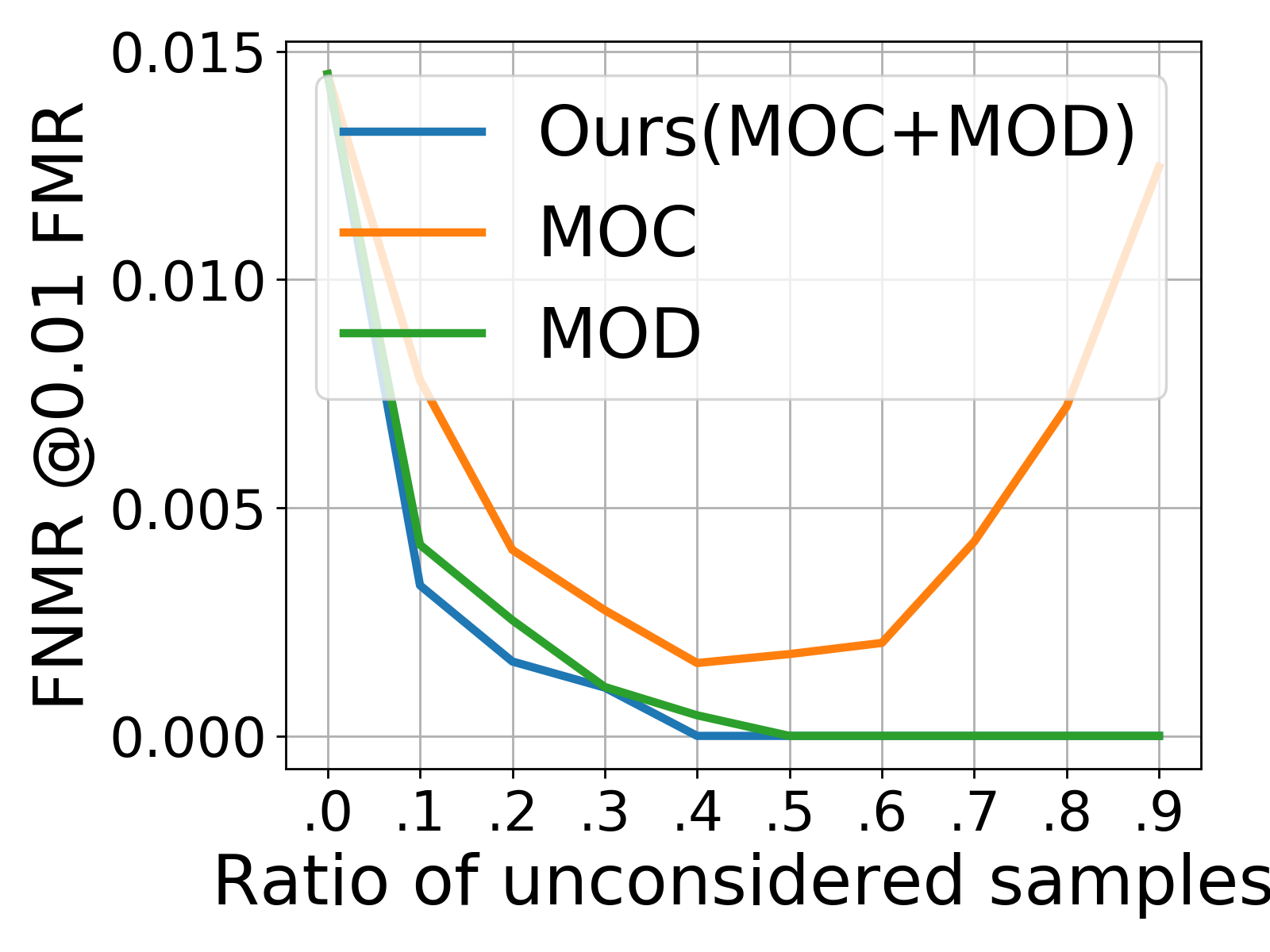}}    
\subfloat[DB3 (thermal sensor)\label{fig:Components_DB3_01_Bo}]{%
       \includegraphics[width=0.20\textwidth]{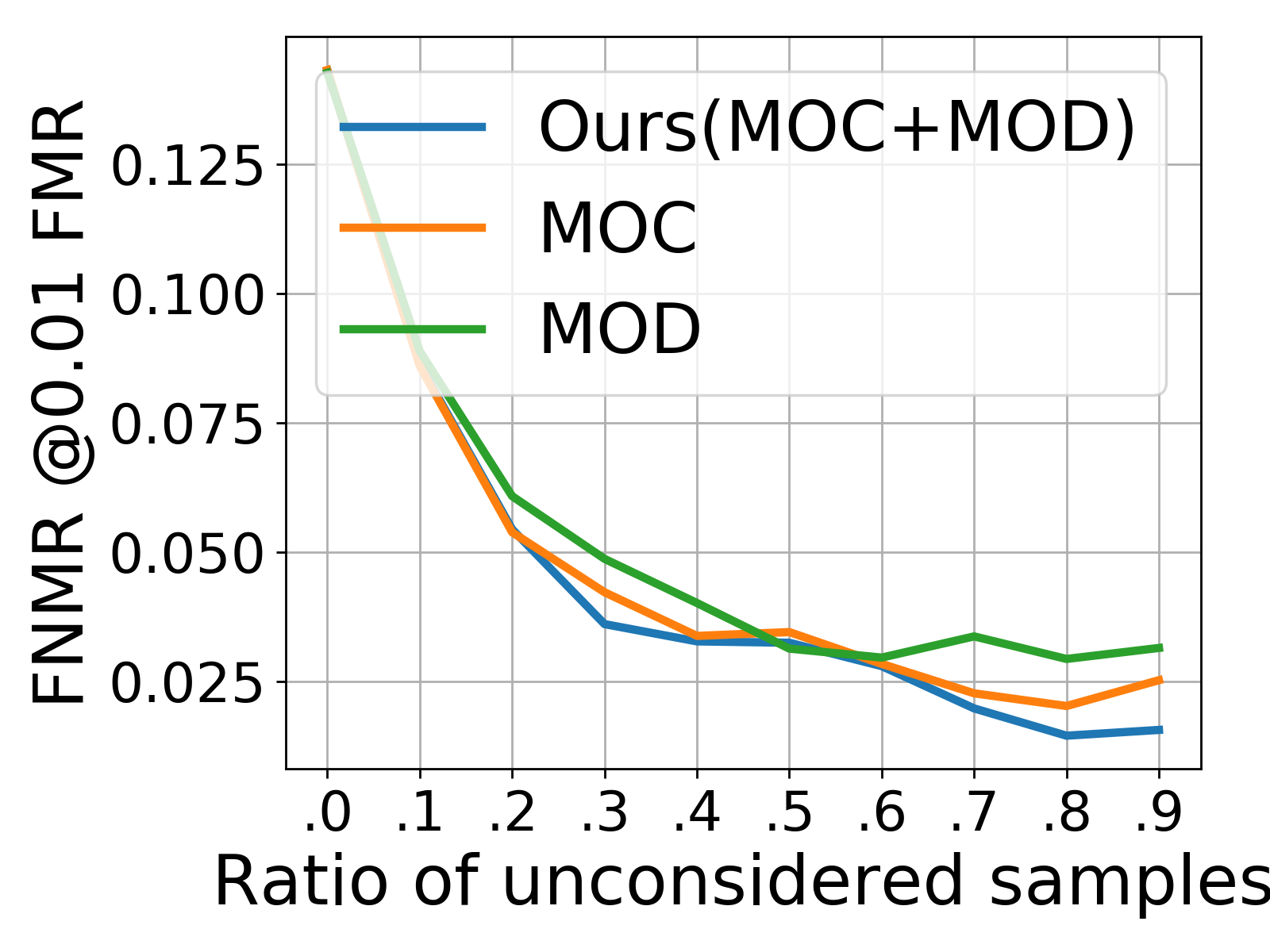}} 
\subfloat[DB4 (synthetic data)\label{fig:Components_DB4_01_Bo}]{%
       \includegraphics[width=0.20\textwidth]{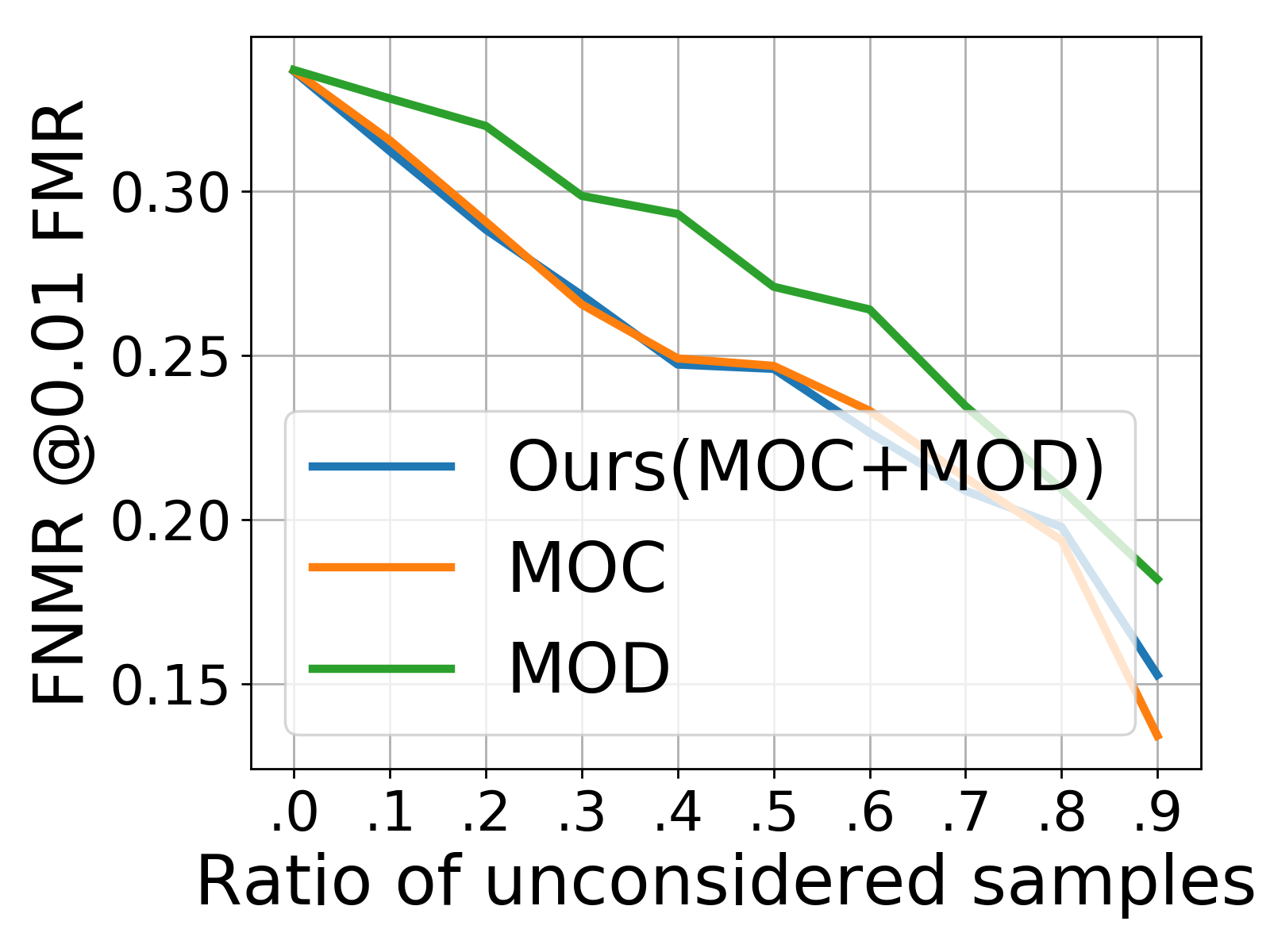}} 

\subfloat[DB1 (electric field sensor)\label{fig:Components_DB1_001_Bo}]{%
       \includegraphics[width=0.20\textwidth]{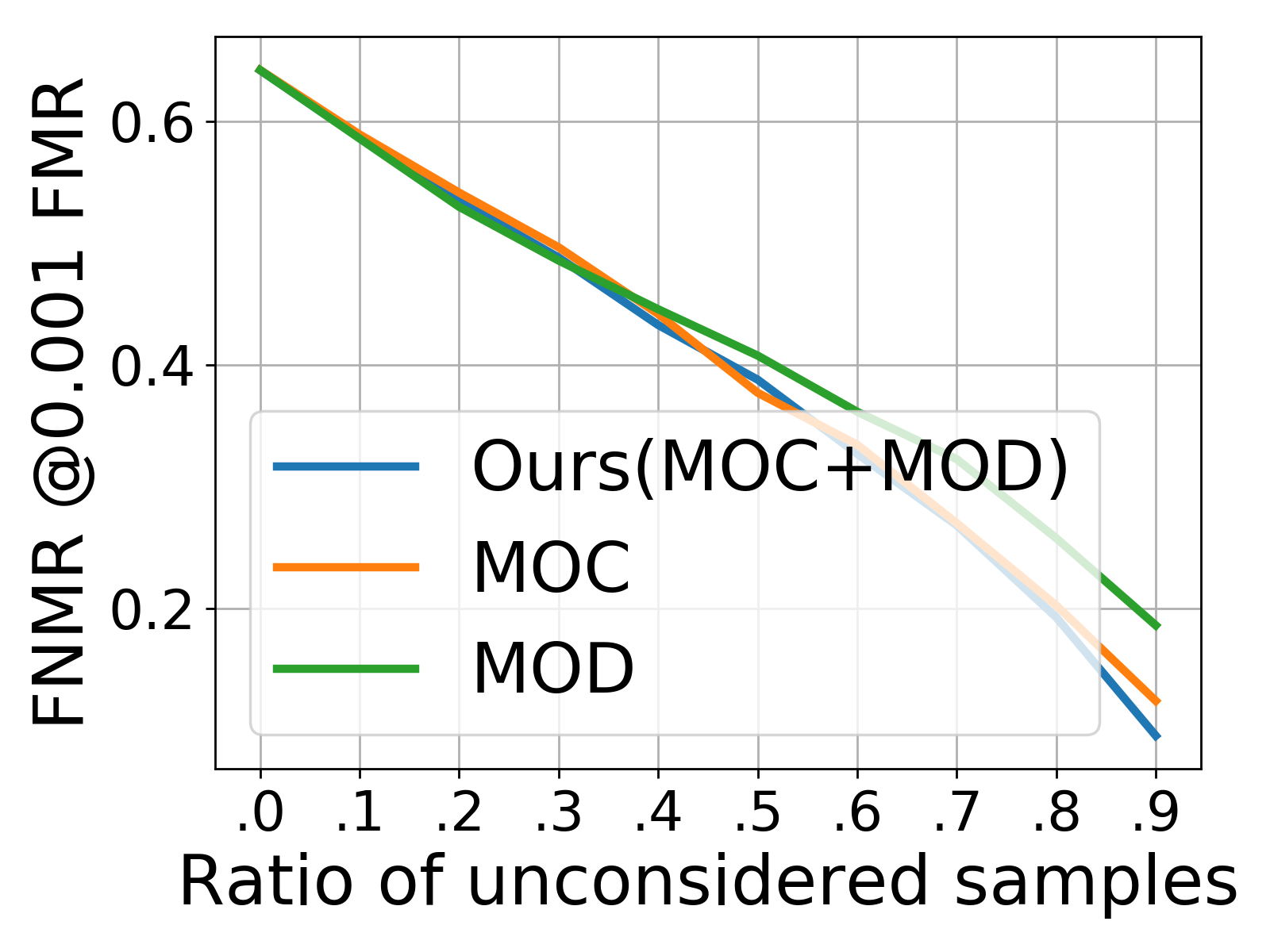}} 
\subfloat[DB2 (optical sensor)\label{fig:Components_DB2_001_Bo}]{%
       \includegraphics[width=0.20\textwidth]{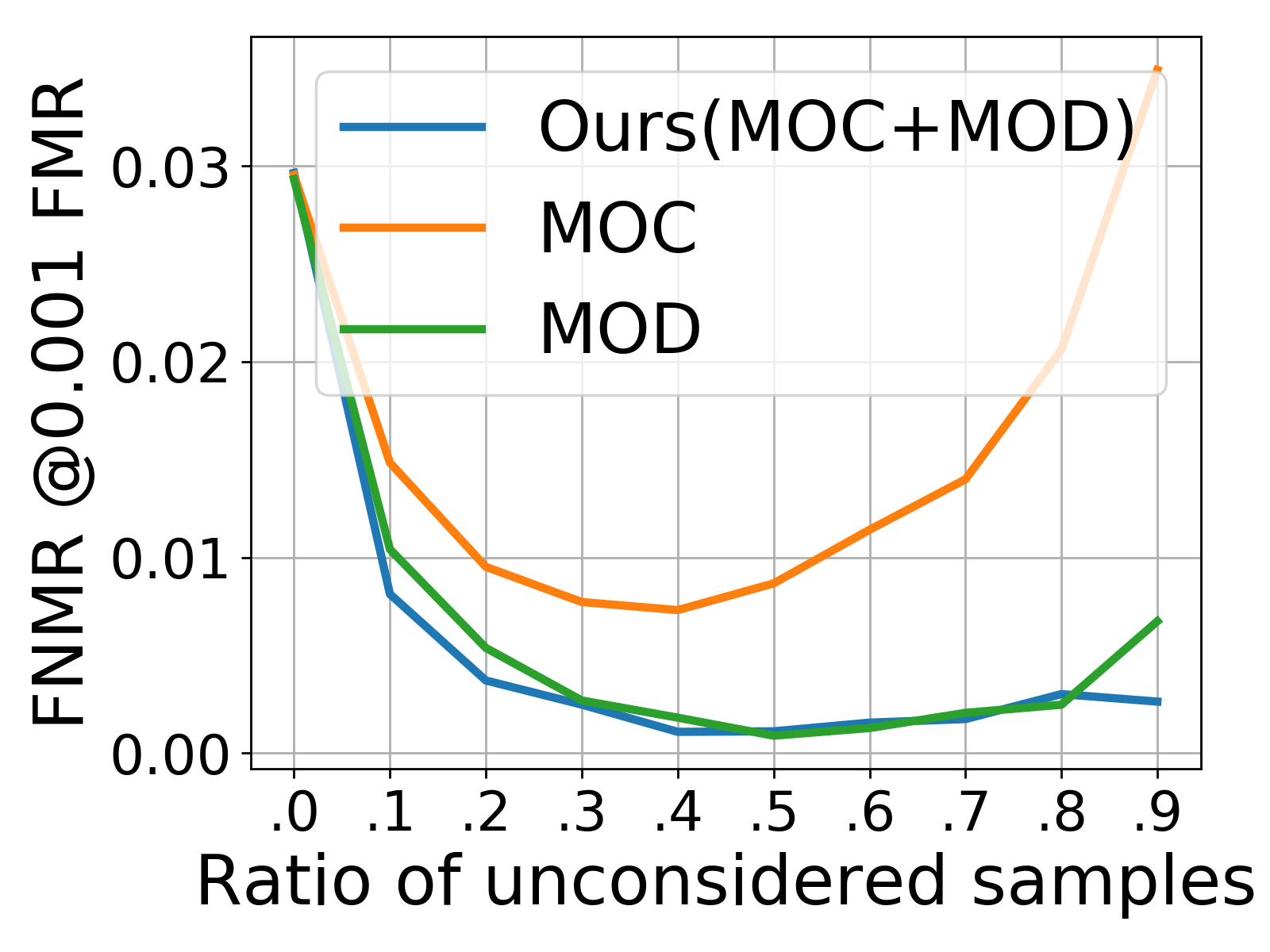}}    
\subfloat[DB3 (thermal sensor)\label{fig:Components_DB3_001_Bo}]{%
       \includegraphics[width=0.20\textwidth]{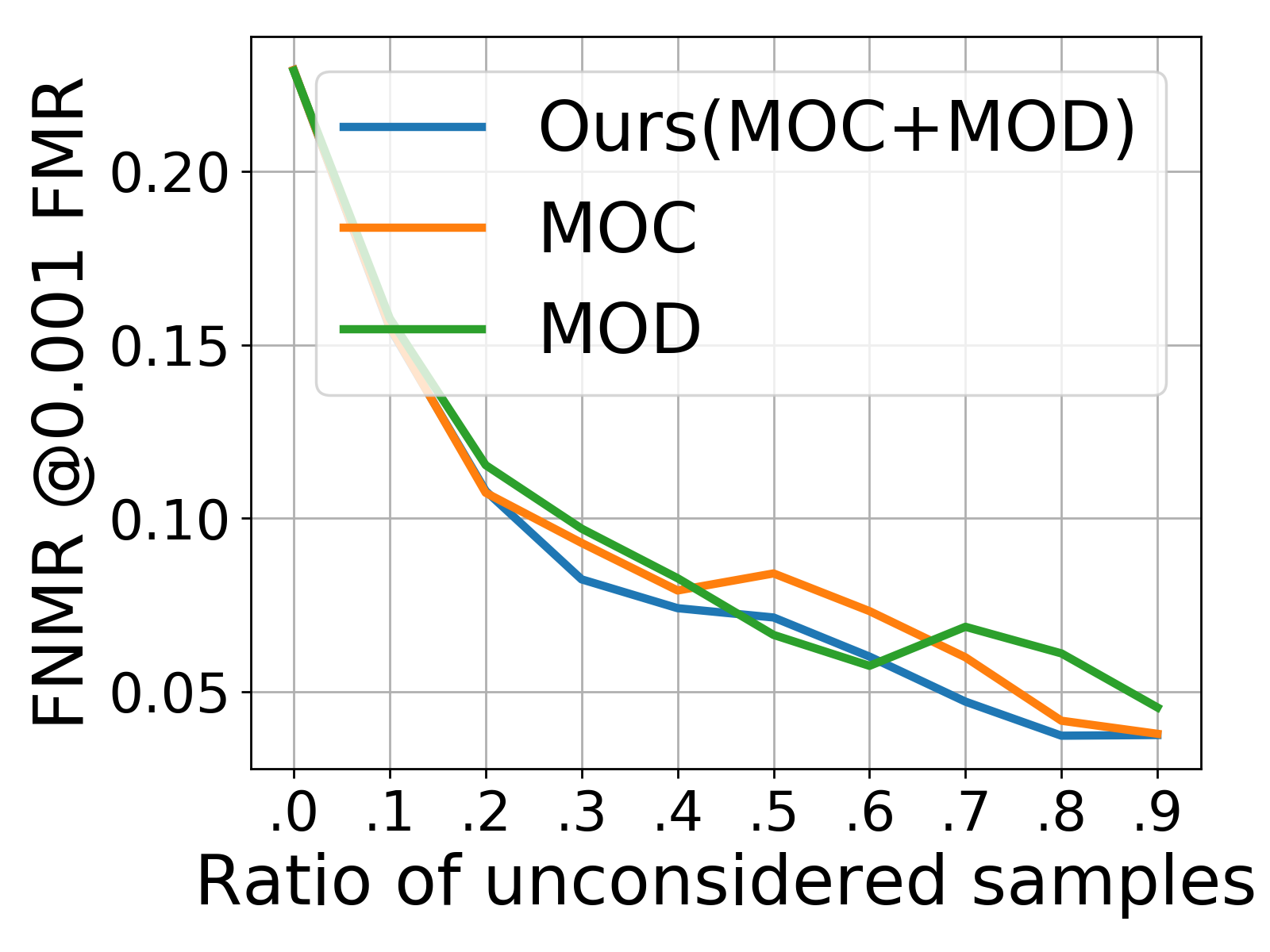}} 
\subfloat[DB4 (synthetic data)\label{fig:Components_DB4_001_Bo}]{%
       \includegraphics[width=0.20\textwidth]{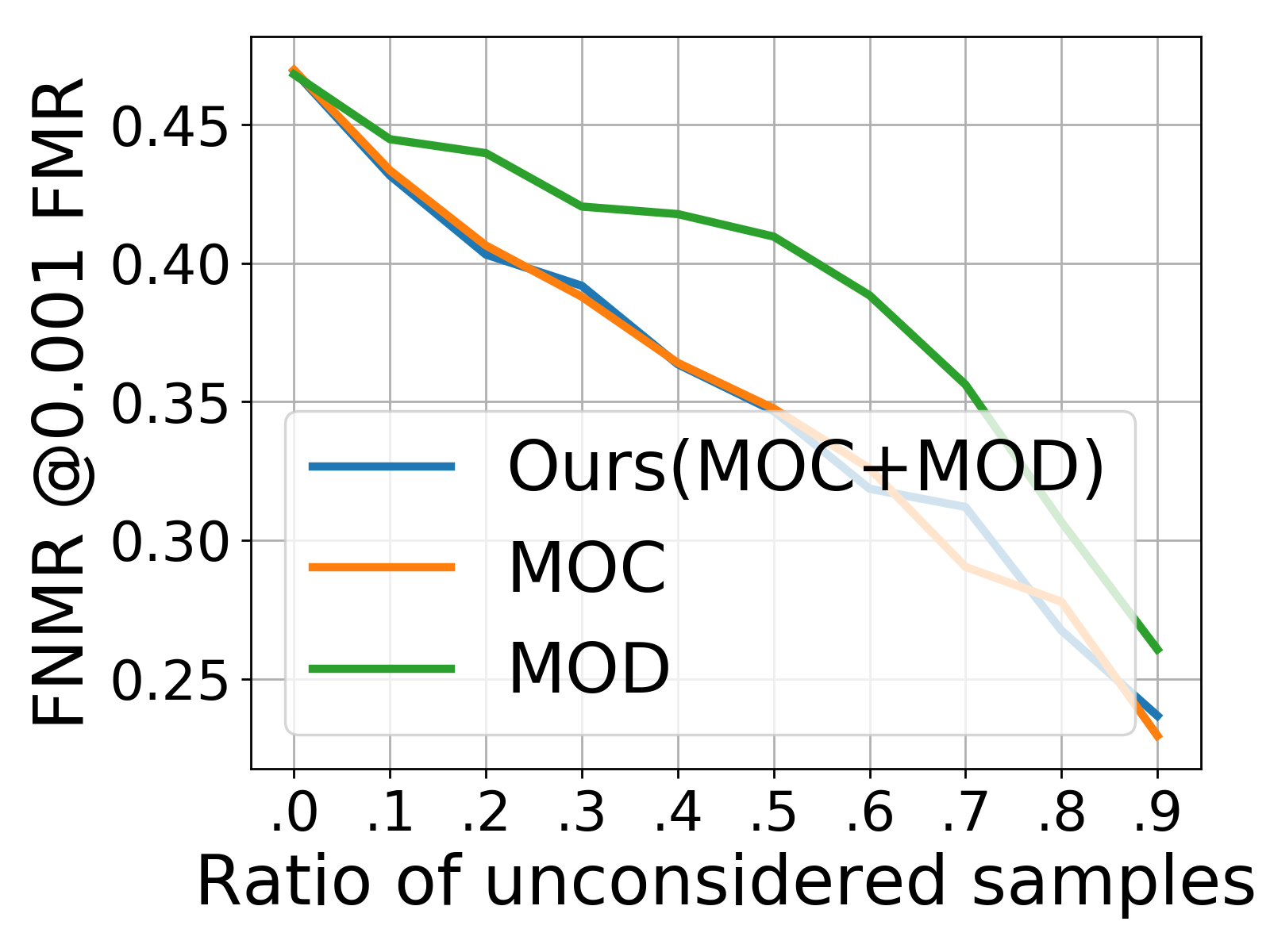}}

\caption{Component analysis of the quality assessment using Equation \ref{eq:MinuQuality}. The fingerprint quality assessment was evaluated on the Bozorth3 matcher.
While the proposed method combines MOC and MOD, here, we additionally analyse the performance when only one of these measures is considered for estimating quality.
Each row represents the recognition error at a different FMR ($10^{-1}$, $10^{-2}$, and $10^{-3}$). The results demonstrate that both, MOD and MOC, are necessary to ensure a stable and high performance across all scenarios.}
\label{fig:Components_Bozorth3}
\end{figure*}

\begin{figure*}[]
\captionsetup[subfloat]{farskip=5pt,captionskip=1pt}
\centering

\subfloat[DB1 (electric field sensor)\label{fig:Components_DB1_1_MCC}]{%
       \includegraphics[width=0.20\textwidth]{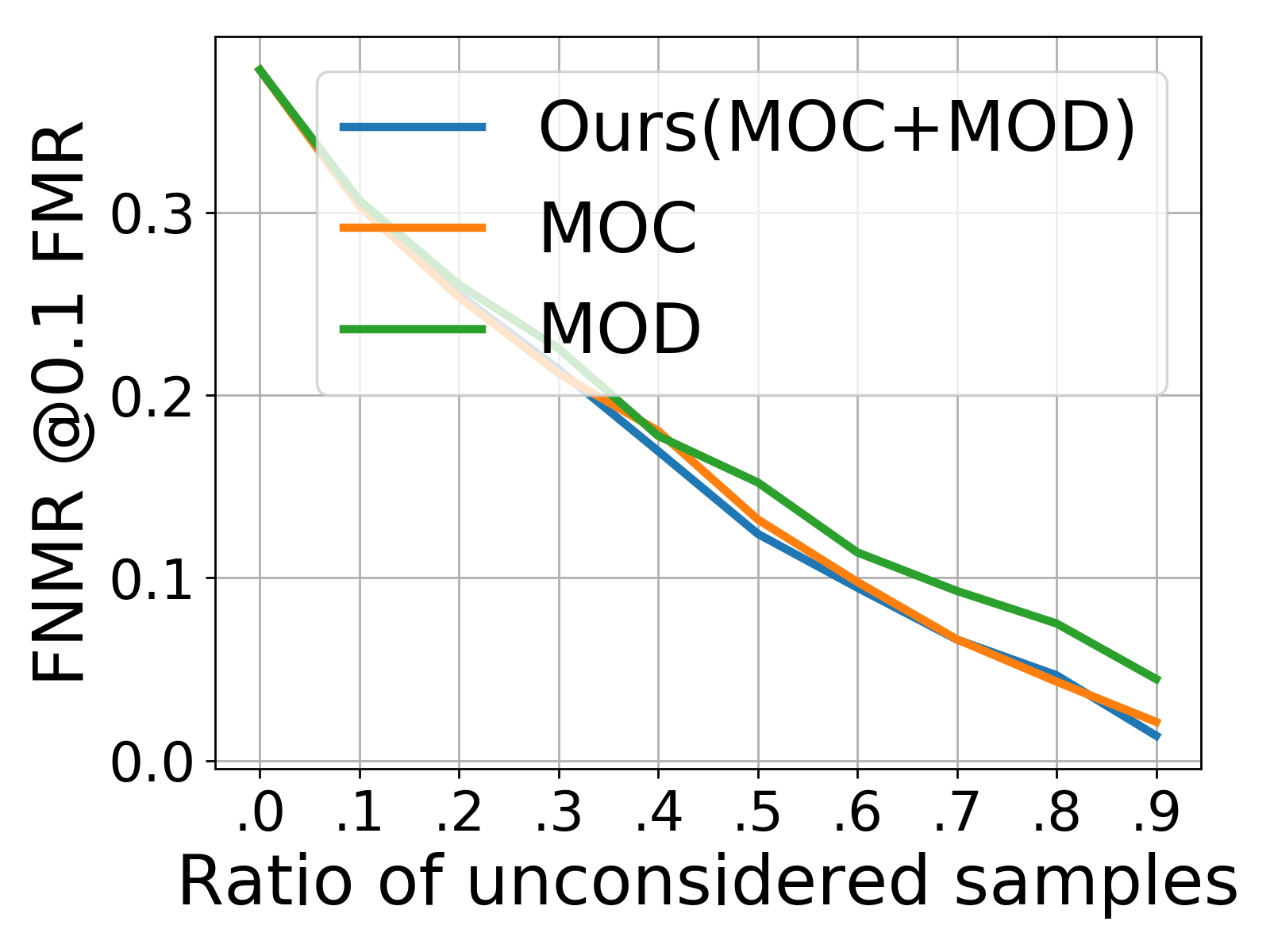}} 
\subfloat[DB2 (optical sensor)\label{fig:Components_DB2_1_MCC}]{%
       \includegraphics[width=0.20\textwidth]{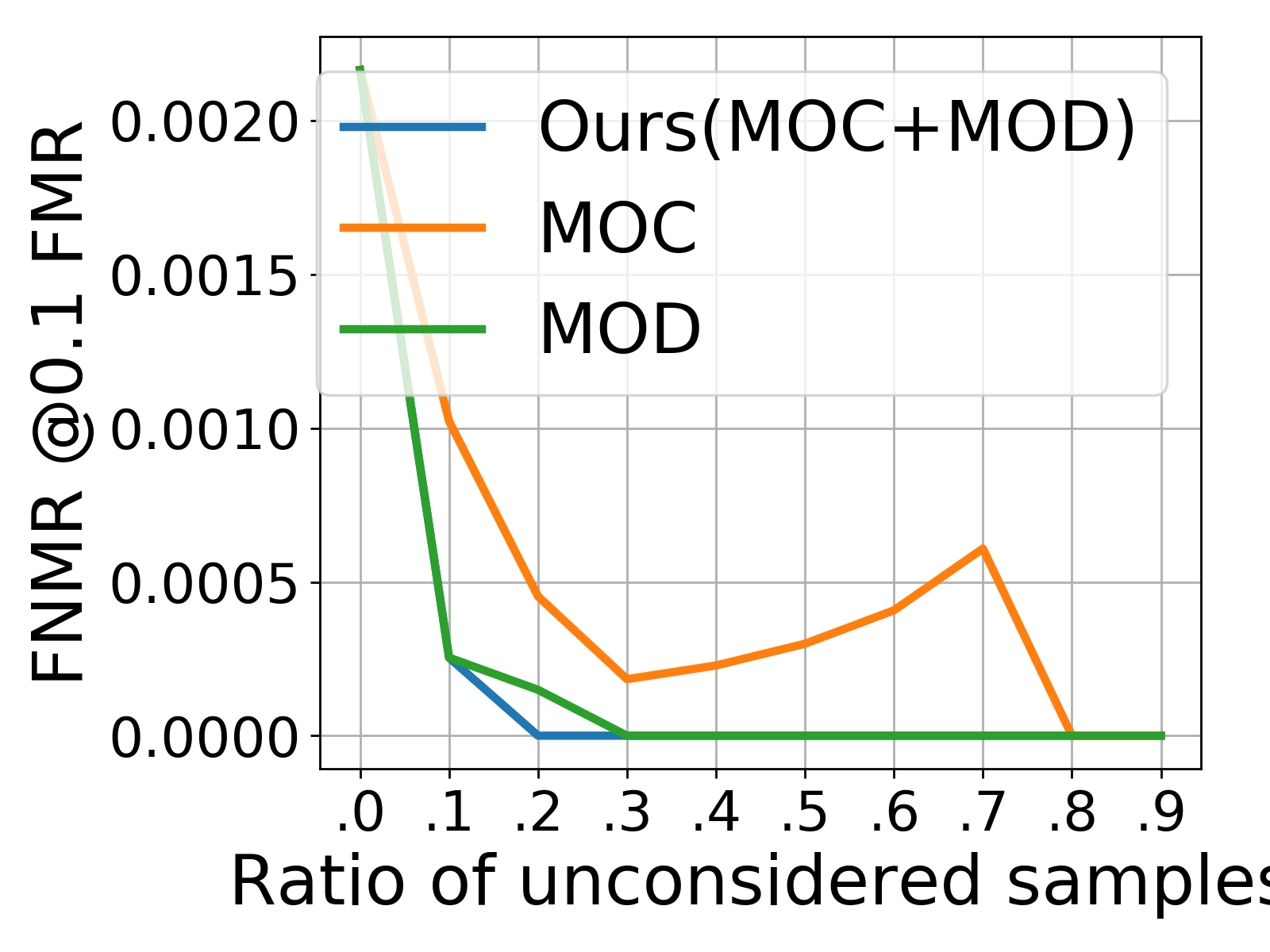}}    
\subfloat[DB3 (thermal sensor)\label{fig:Components_DB3_1_MCC}]{%
       \includegraphics[width=0.20\textwidth]{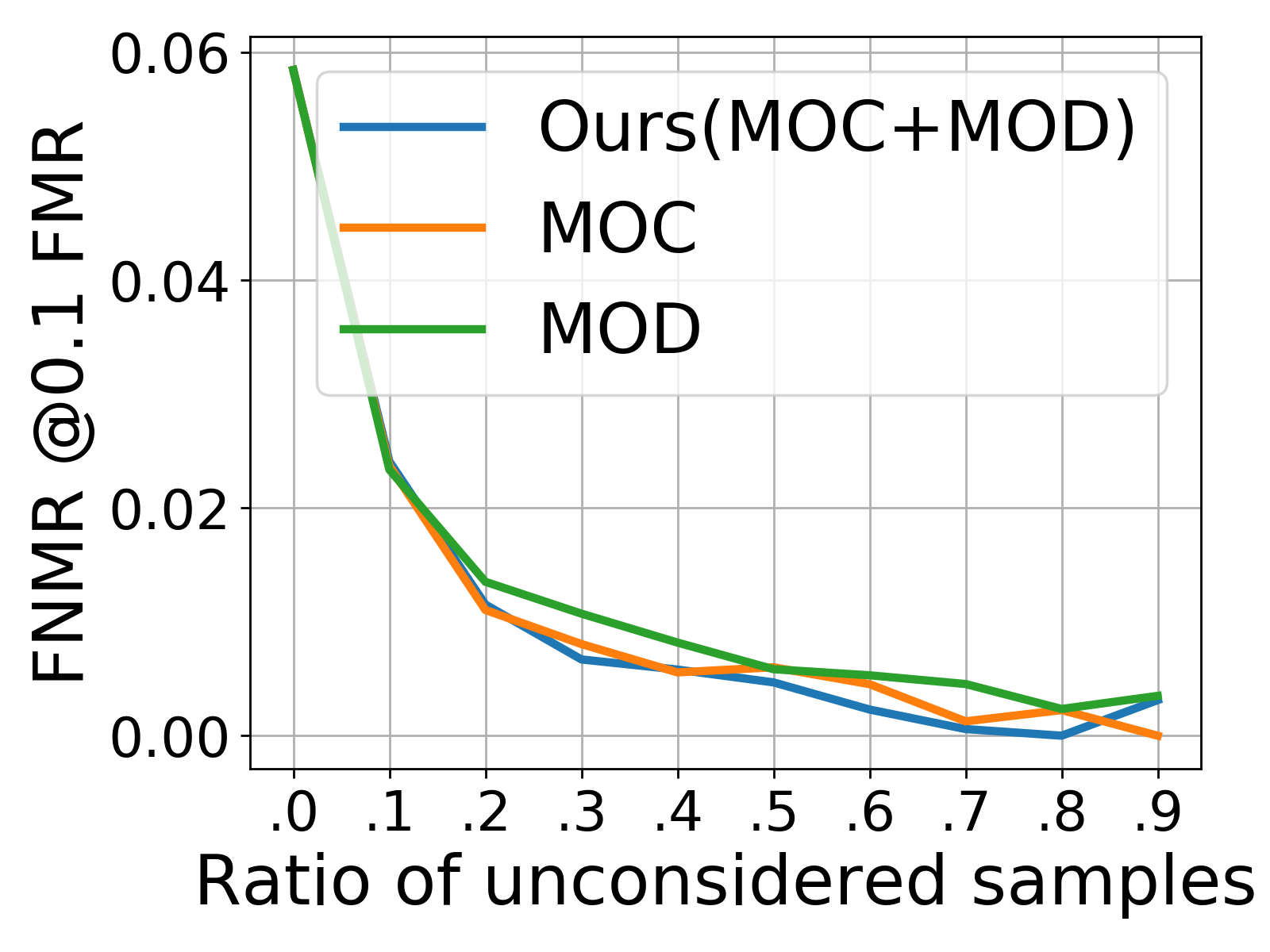}} 
\subfloat[DB4 (synthetic data)\label{fig:Components_DB4_1_MCC}]{%
       \includegraphics[width=0.20\textwidth]{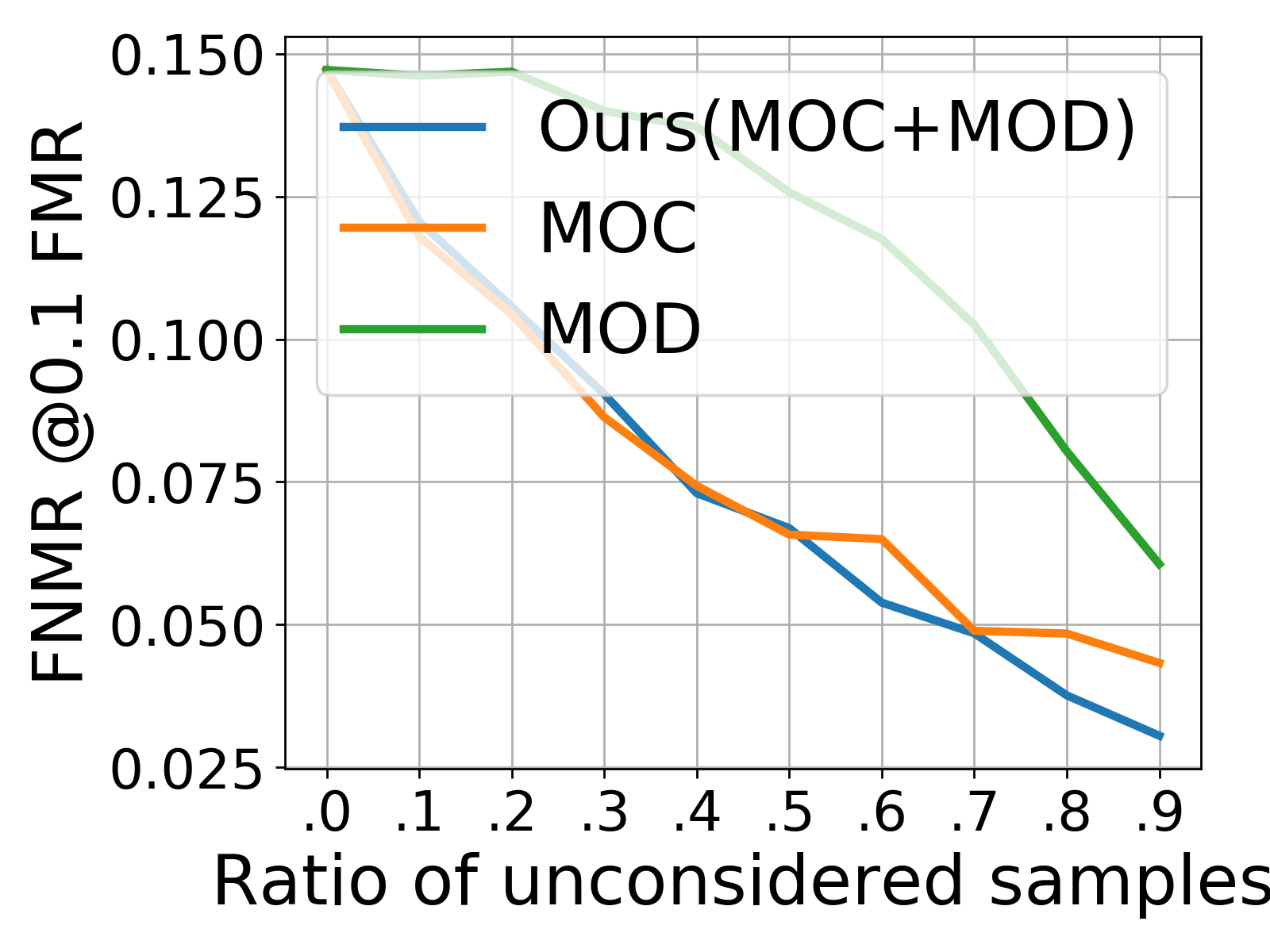}} 
               
\subfloat[DB1 (electric field sensor)\label{fig:Components_DB1_01_MCC}]{%
       \includegraphics[width=0.20\textwidth]{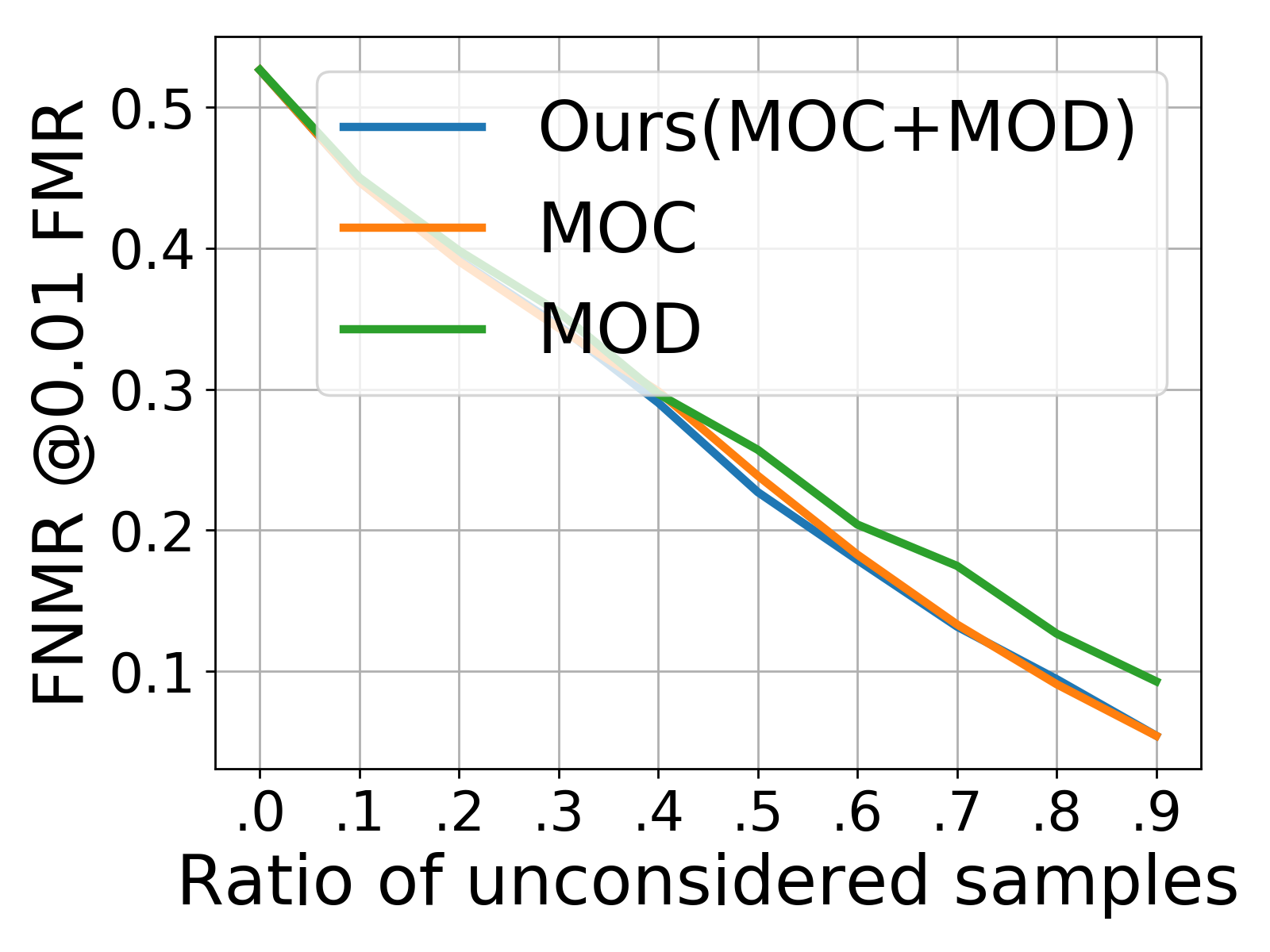}} 
\subfloat[DB2 (optical sensor)\label{fig:Components_DB2_01_MCC}]{%
       \includegraphics[width=0.20\textwidth]{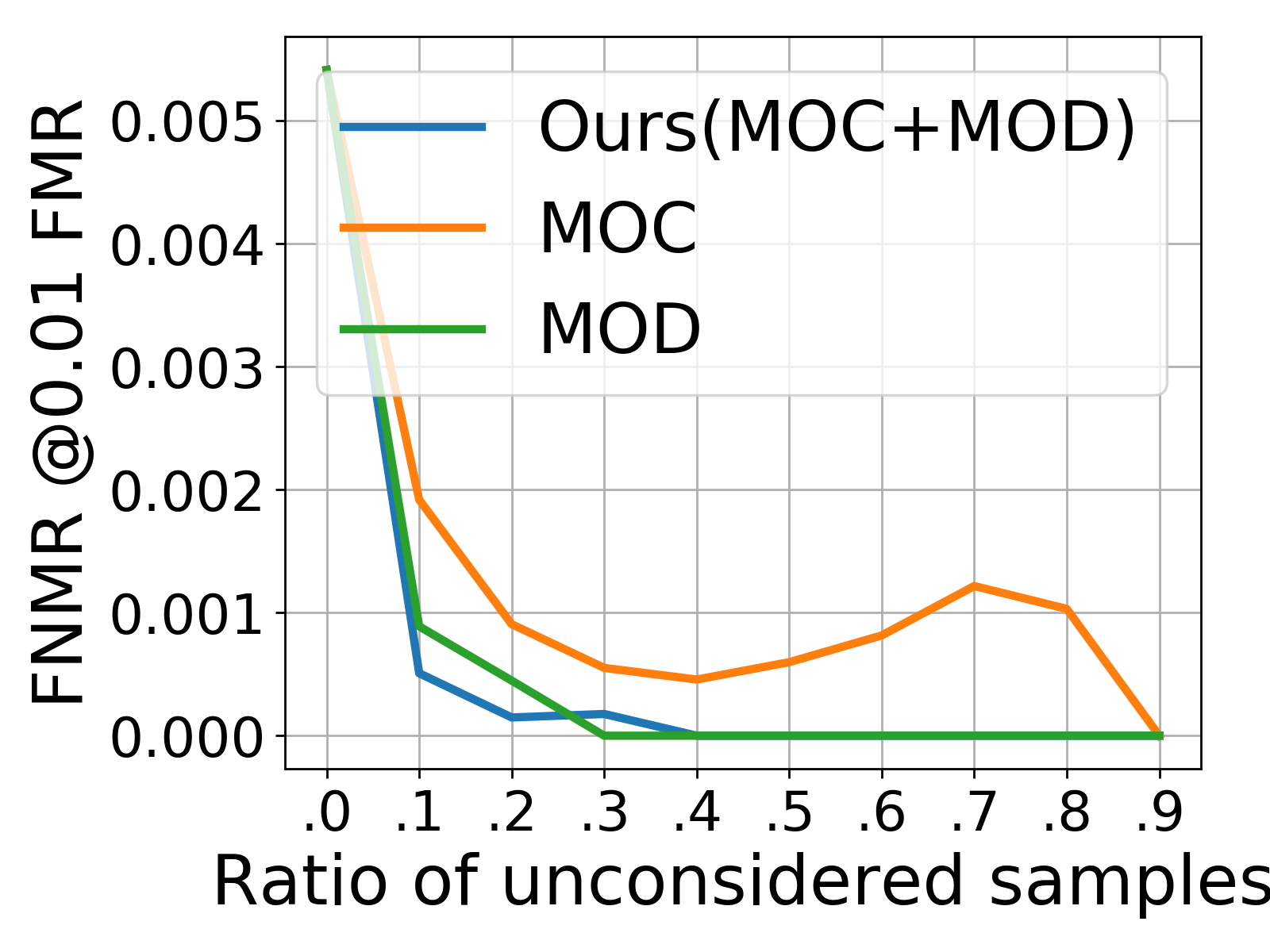}}    
\subfloat[DB3 (thermal sensor)\label{fig:Components_DB3_01_MCC}]{%
       \includegraphics[width=0.20\textwidth]{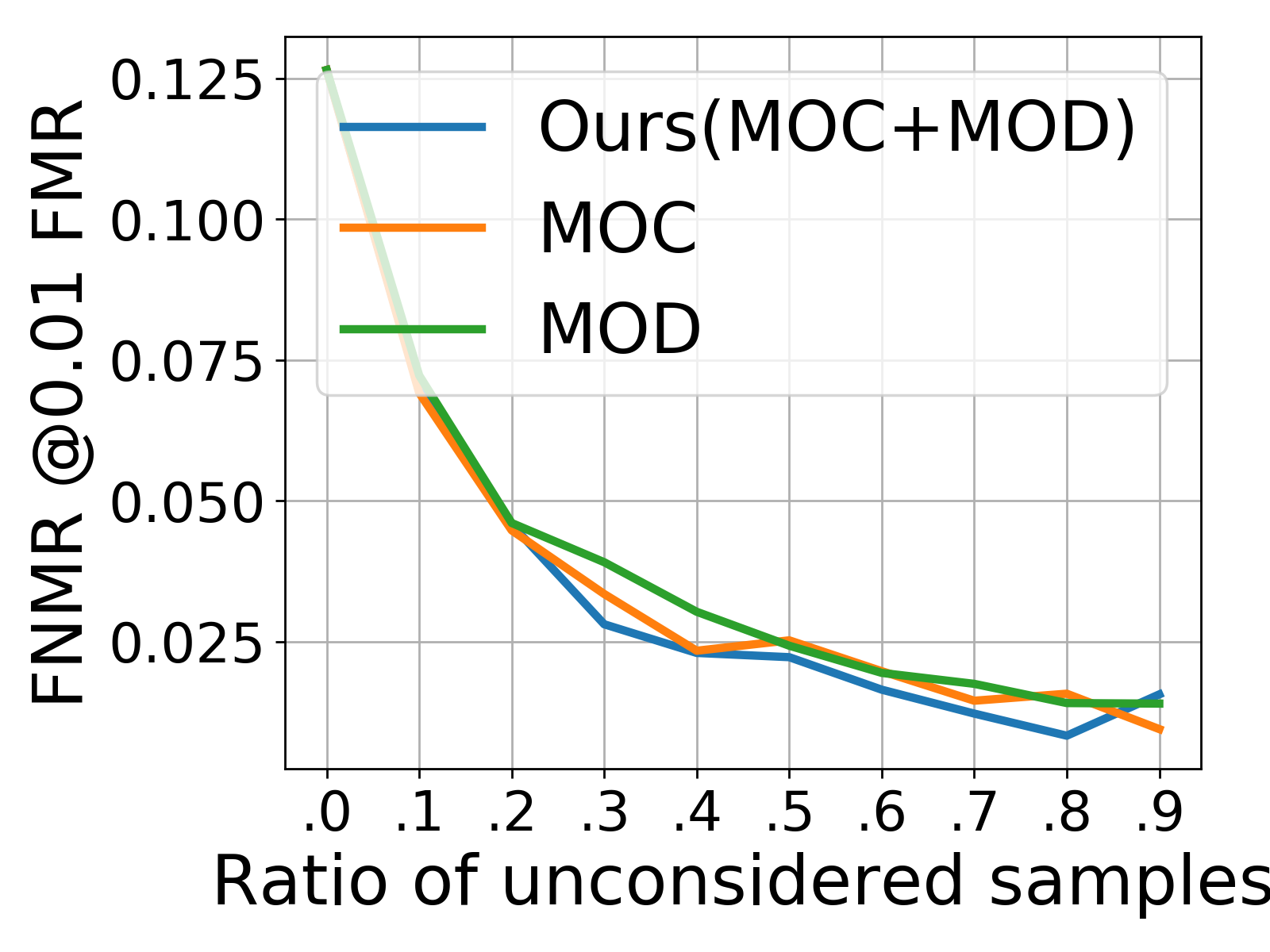}} 
\subfloat[DB4 (synthetic data)\label{fig:Components_DB4_01_MCC}]{%
       \includegraphics[width=0.20\textwidth]{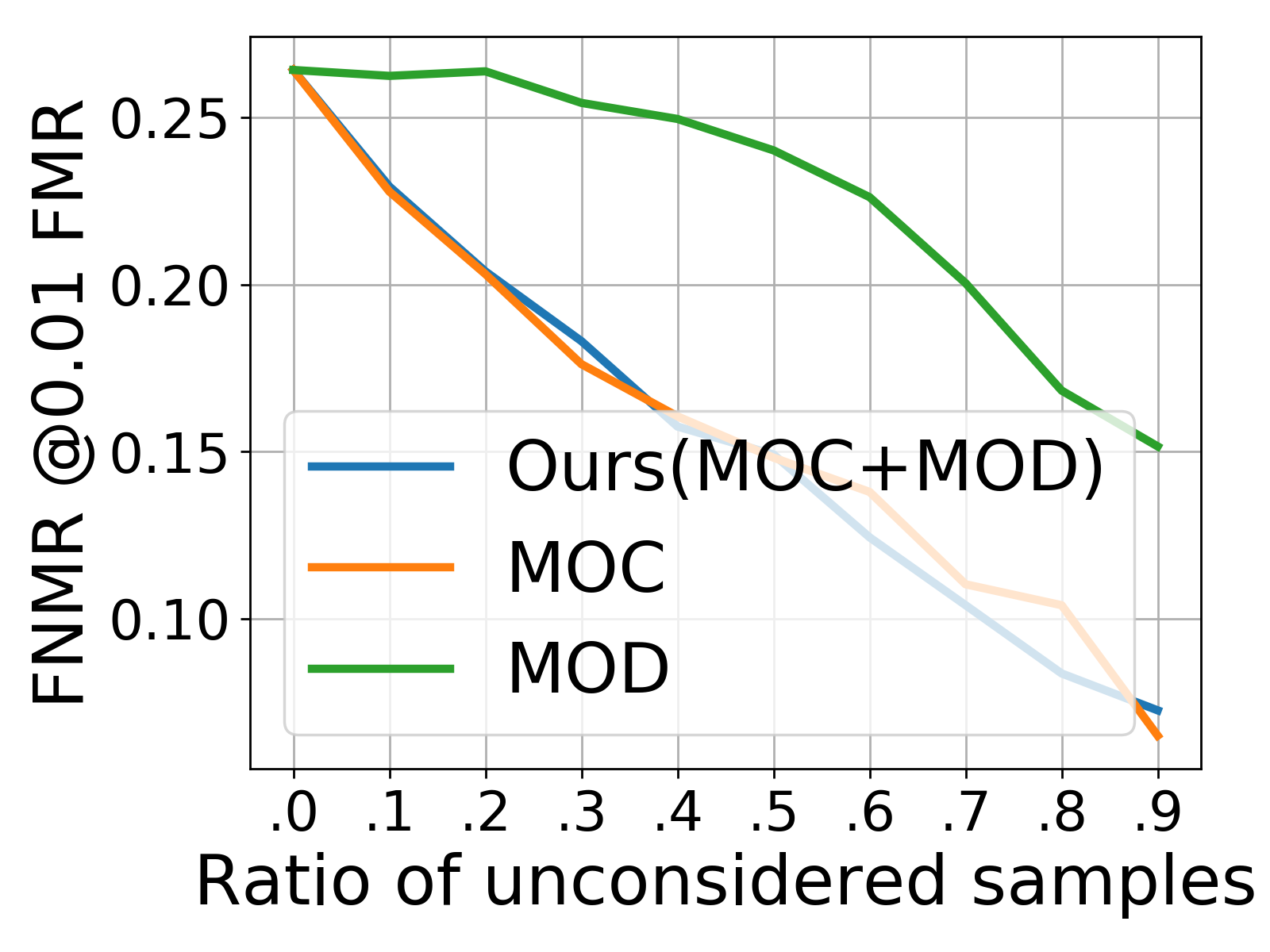}} 

\subfloat[DB1 (electric field sensor)\label{fig:Components_DB1_001_MCC}]{%
       \includegraphics[width=0.20\textwidth]{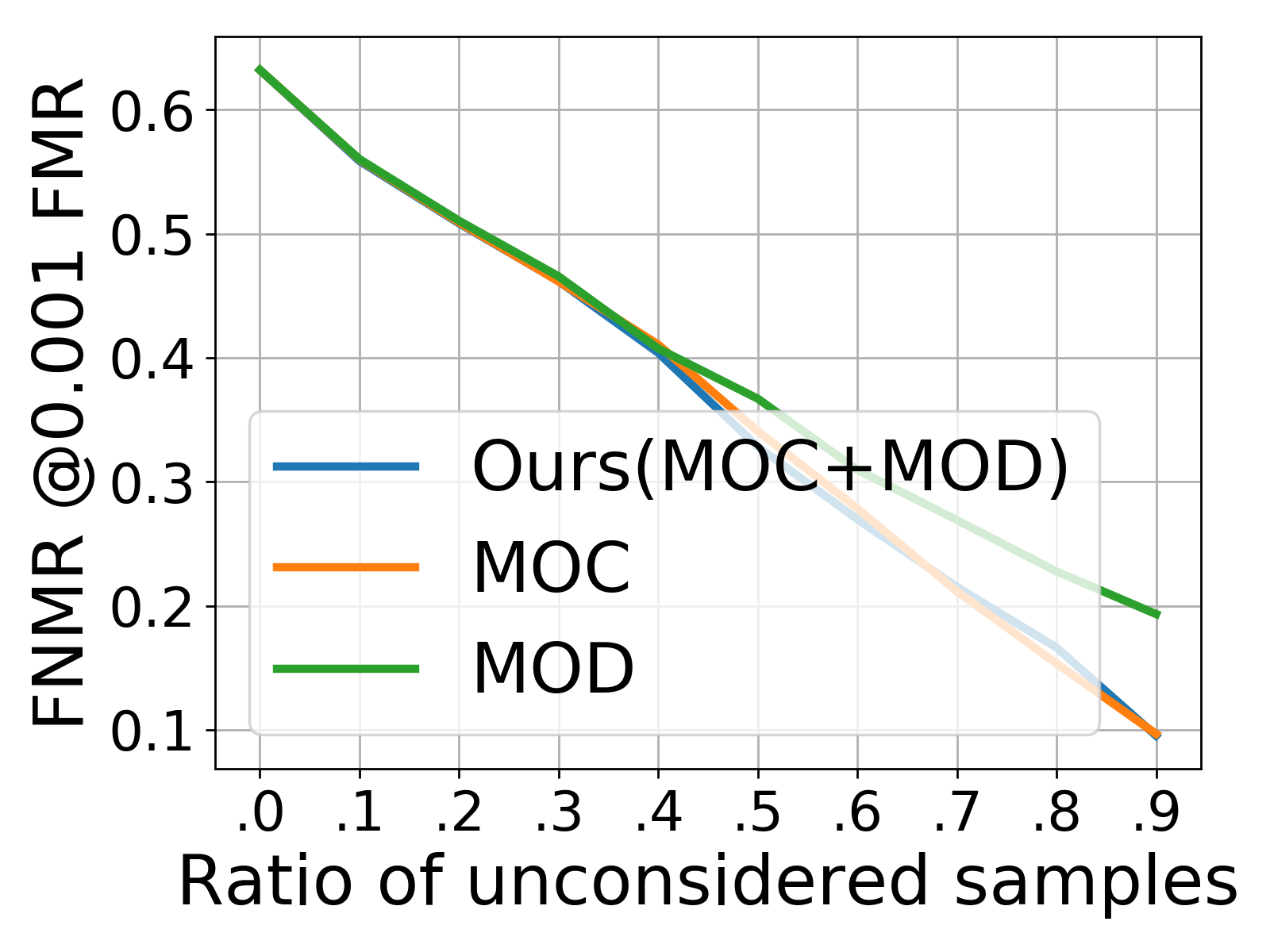}} 
\subfloat[DB2 (optical sensor)\label{fig:Components_DB2_001_MCC}]{%
       \includegraphics[width=0.20\textwidth]{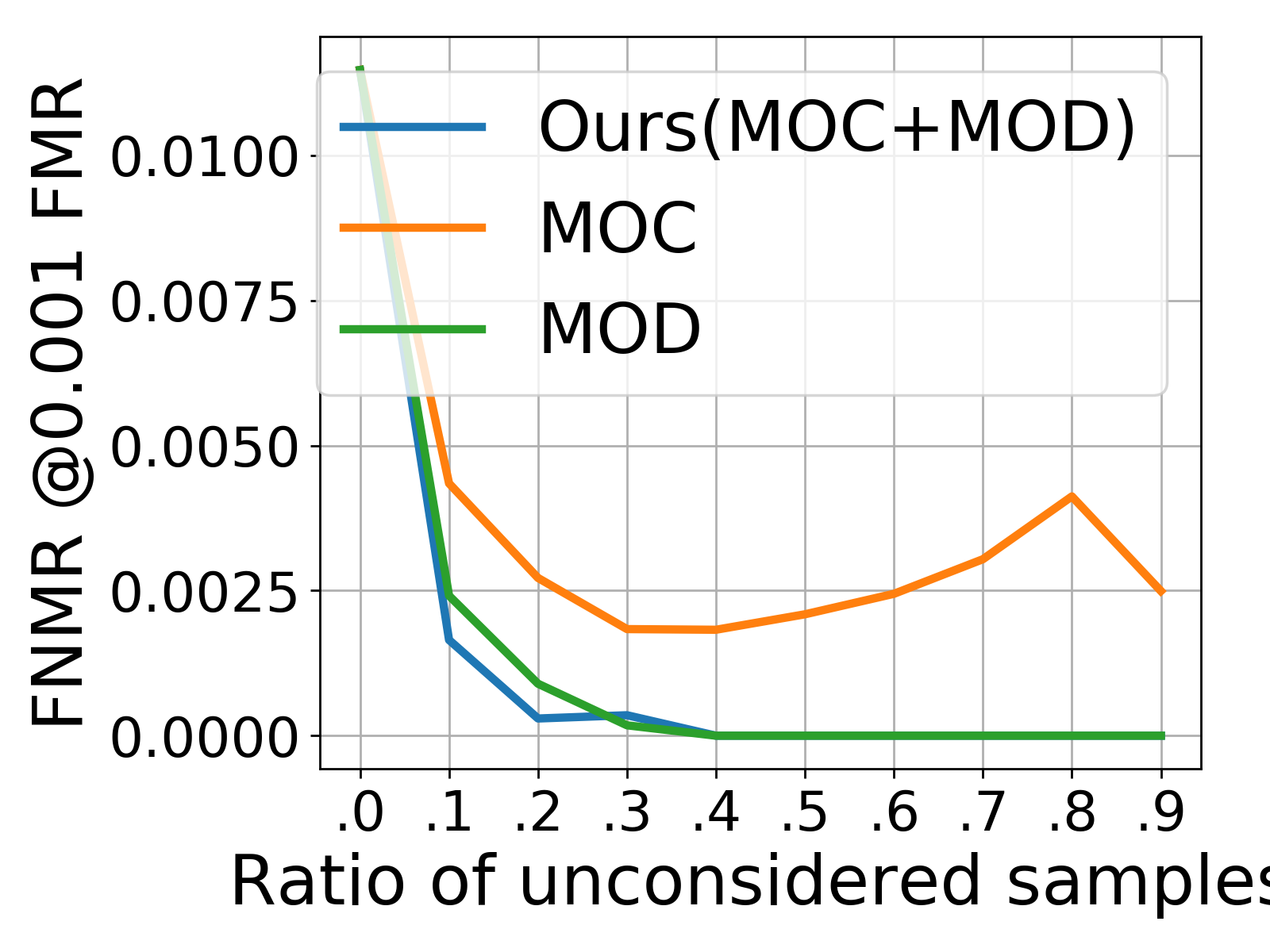}}    
\subfloat[DB3 (thermal sensor)\label{fig:Components_DB3_001_MCC}]{%
       \includegraphics[width=0.20\textwidth]{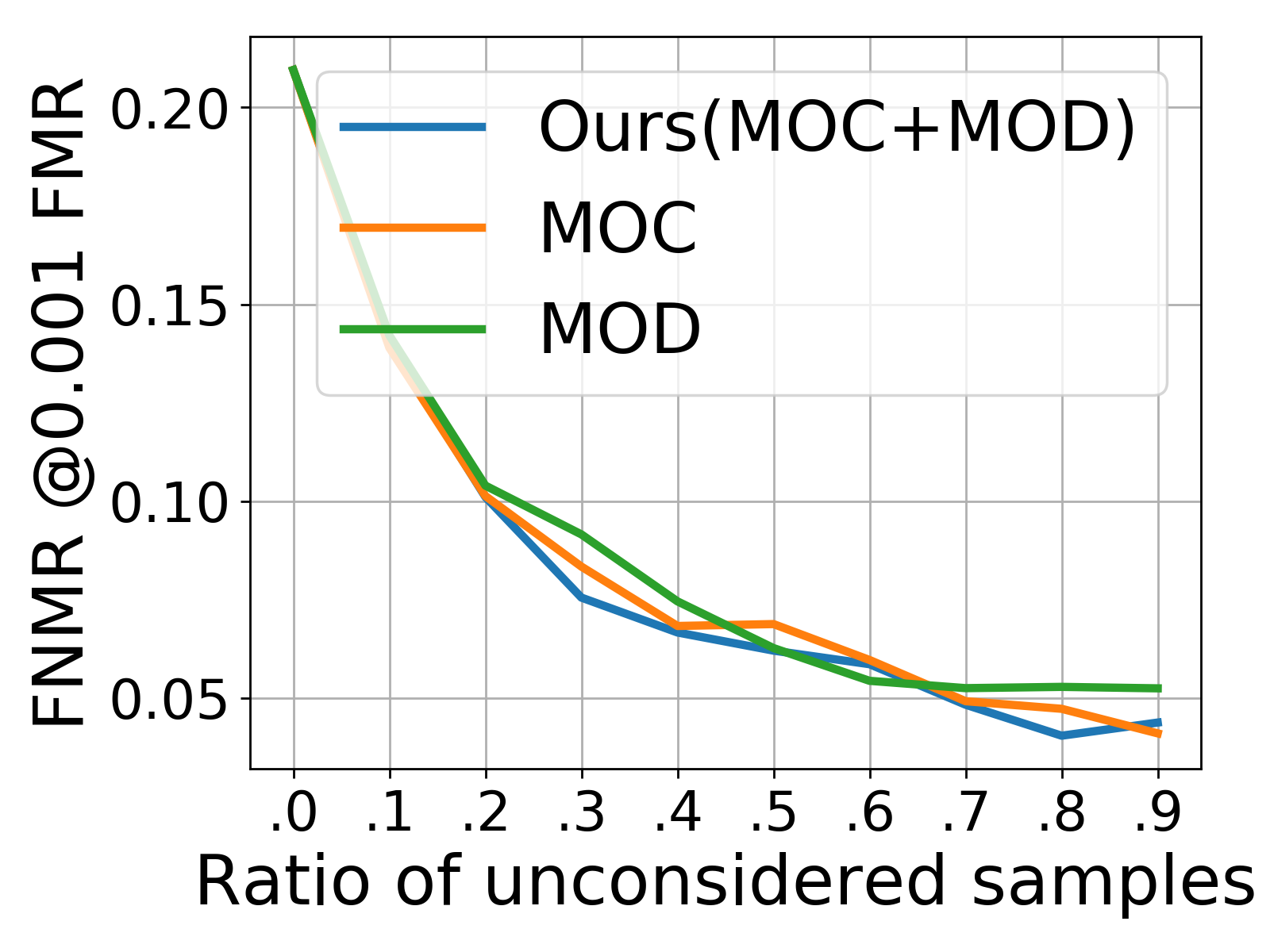}} 
\subfloat[DB4 (synthetic data)\label{fig:Components_DB4_001_MCC}]{%
       \includegraphics[width=0.20\textwidth]{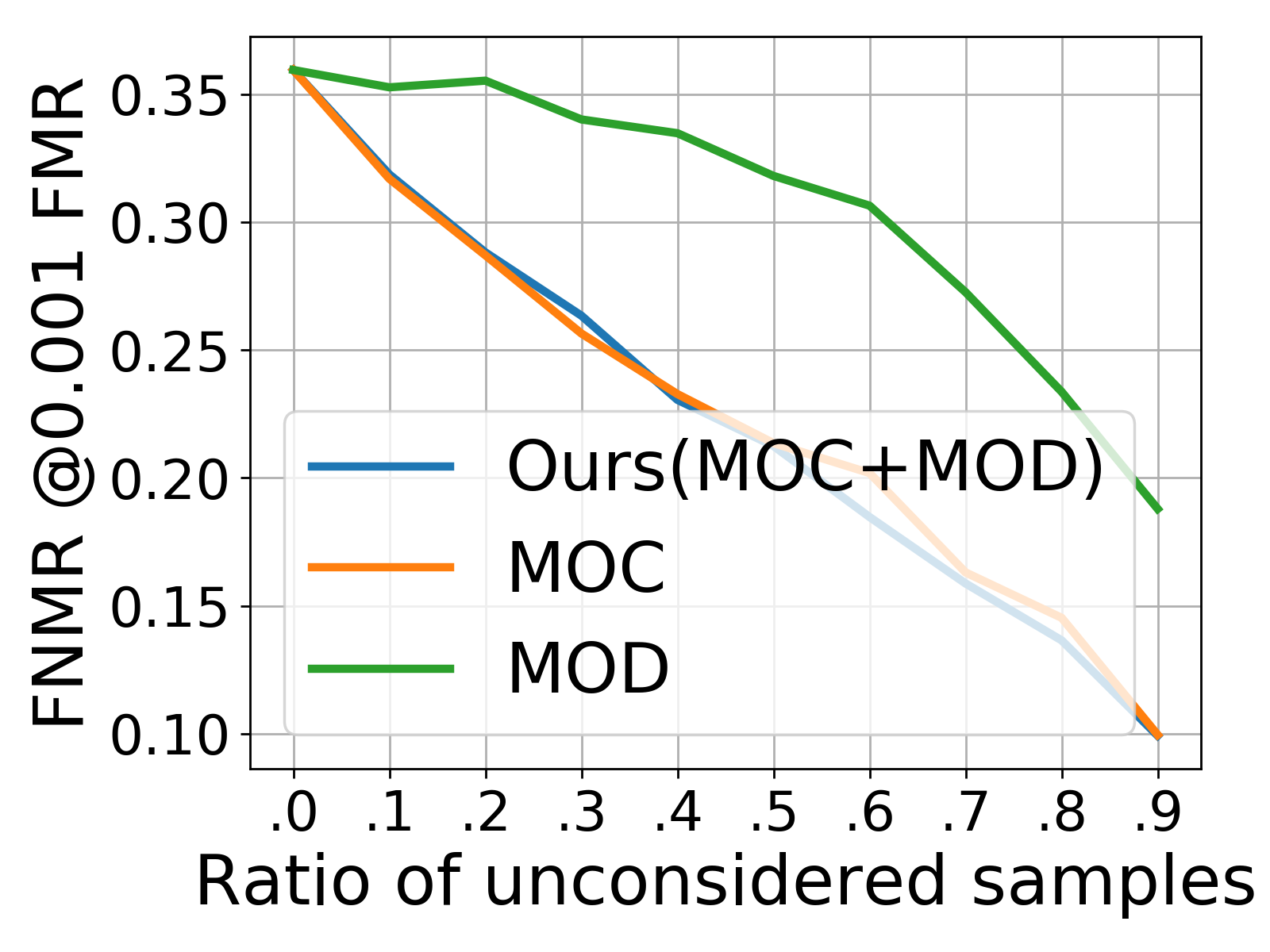}}

\caption{Component analysis of the quality assessment using Equation \ref{eq:MinuQuality}. The fingerprint quality assessment was evaluated on the MCC matcher.
While the proposed method combines MOC and MOD, here, we additionally analyse the performance when only one of these measures is considered for estimating quality.
Each row represents the recognition error at a different FMR ($10^{-1}$, $10^{-2}$, and $10^{-3}$). The results demonstrate that both, MOD and MOC, are necessary to ensure a stable and high performance across all scenarios.}
\label{fig:Components_MCC}
\end{figure*}

As explained in Section \ref{sec:Methodology}, the proposed quality estimate (Equation \ref{eq:MinuQuality}) combines a measure of centrality (MOC) and a measure of dispersion (MOD).
In Figures \ref{fig:Components_Bozorth3} and \ref{fig:Components_MCC}, the fingerprint quality assessment performance is analysed using the Bozorth3 and MCC matcher.
The figures show the performance of the proposed approach compared to the performance when only MOC or MOD is used for the quality calculation.
While MOD and MOC show strong and weak performances depending on the investigated scenario, the proposed approach that combines both ensures accurate and stable over all FMRs and all sensor types.
This demonstrates the need for both components for a stable and accurate quality assessment.

\section*{Analysing the number of minutiae considered for FIQ assessment with MiDeCon}
\begin{figure*}[]
\captionsetup[subfloat]{farskip=5pt,captionskip=1pt}
\centering

\subfloat[DB1 (electric field sensor)\label{fig:nAnalysis_DB1_1_Bo}]{%
       \includegraphics[width=0.20\textwidth]{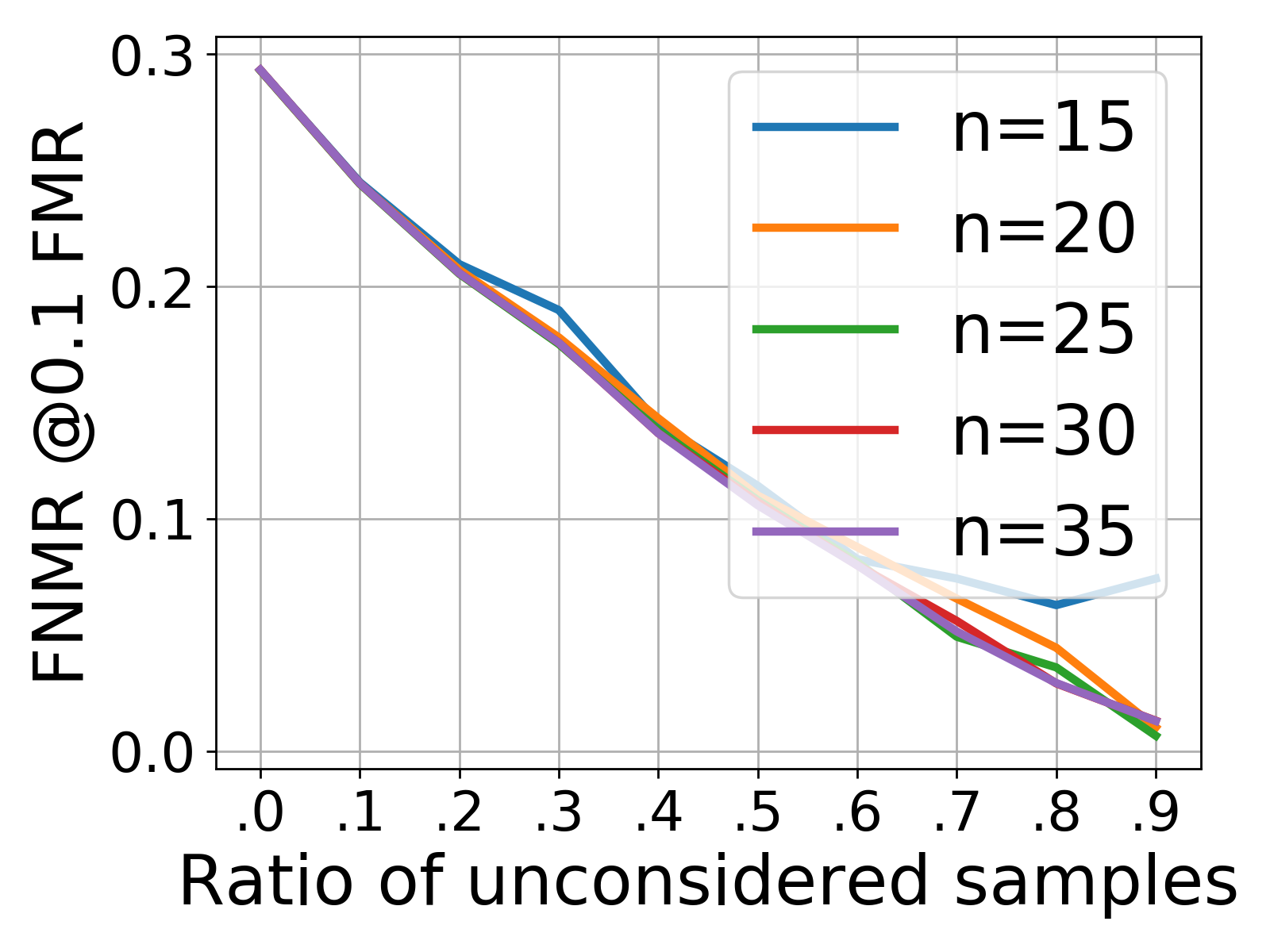}} 
\subfloat[DB2 (optical sensor)\label{fig:nAnalysis_DB2_1_Bo}]{%
       \includegraphics[width=0.20\textwidth]{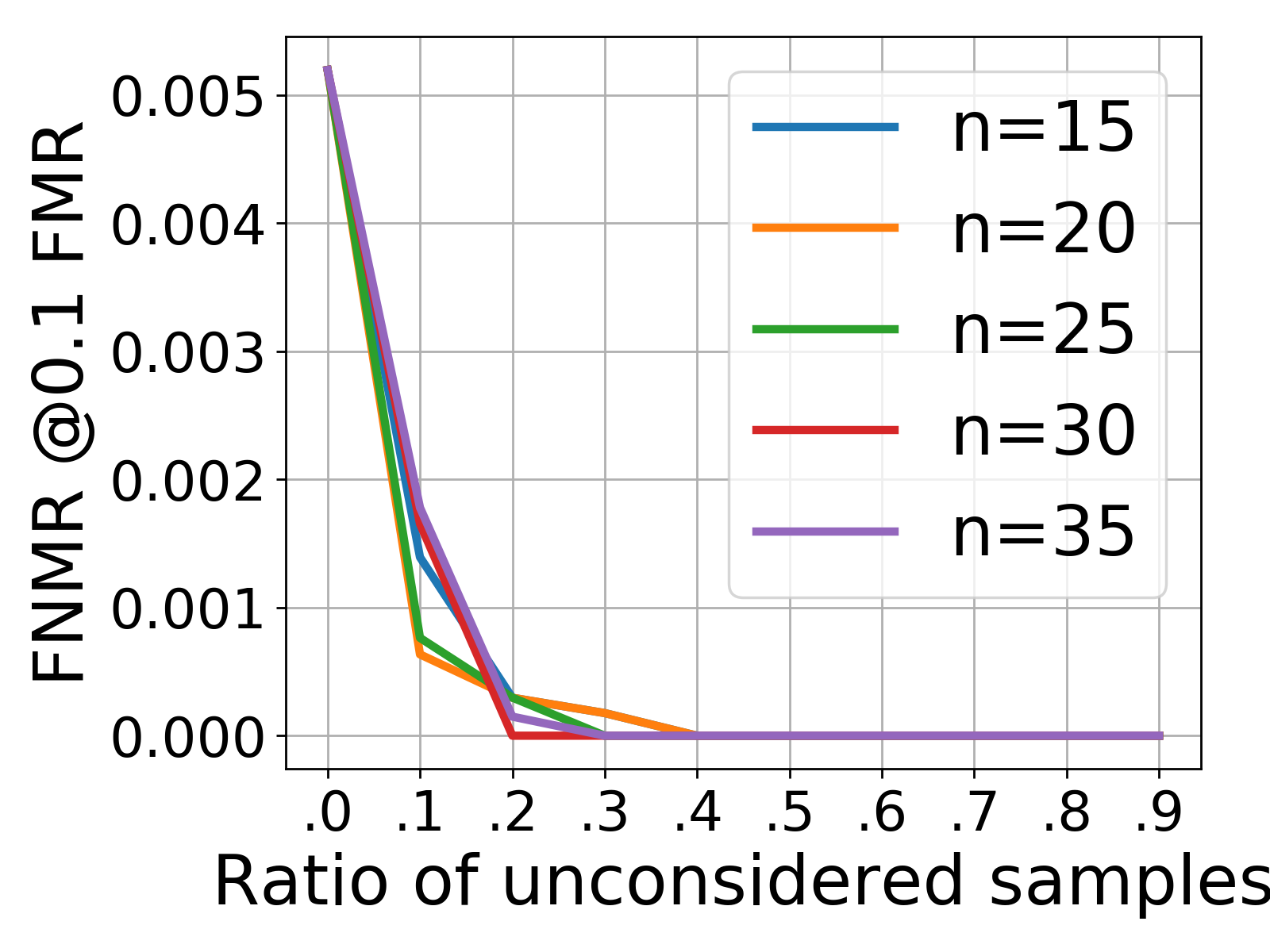}}    
\subfloat[DB3 (thermal sensor)\label{fig:nAnalysis_DB3_1_Bo}]{%
       \includegraphics[width=0.20\textwidth]{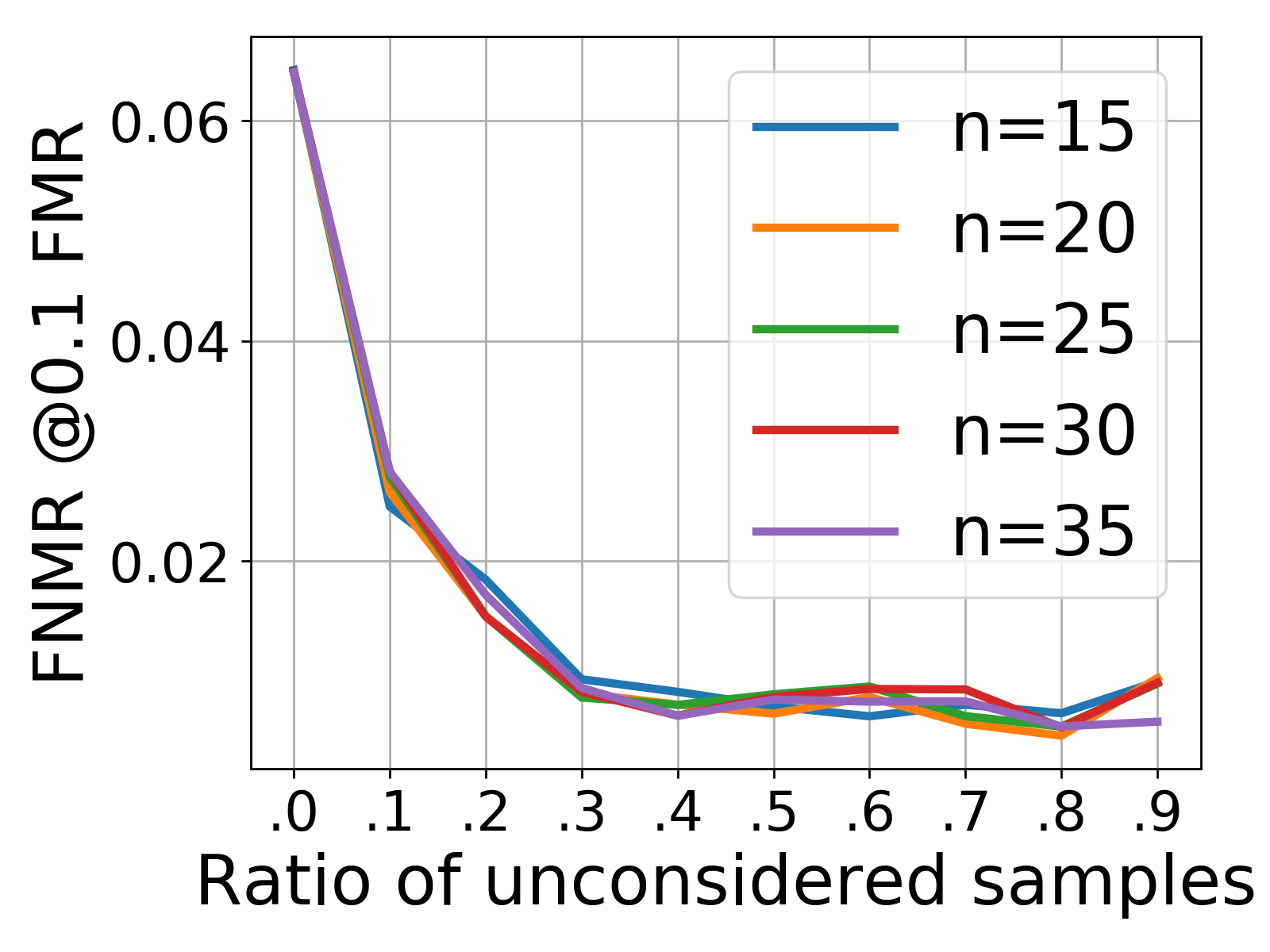}} 
\subfloat[DB4 (synthetic data)\label{fig:nAnalysis_DB4_1_Bo}]{%
       \includegraphics[width=0.20\textwidth]{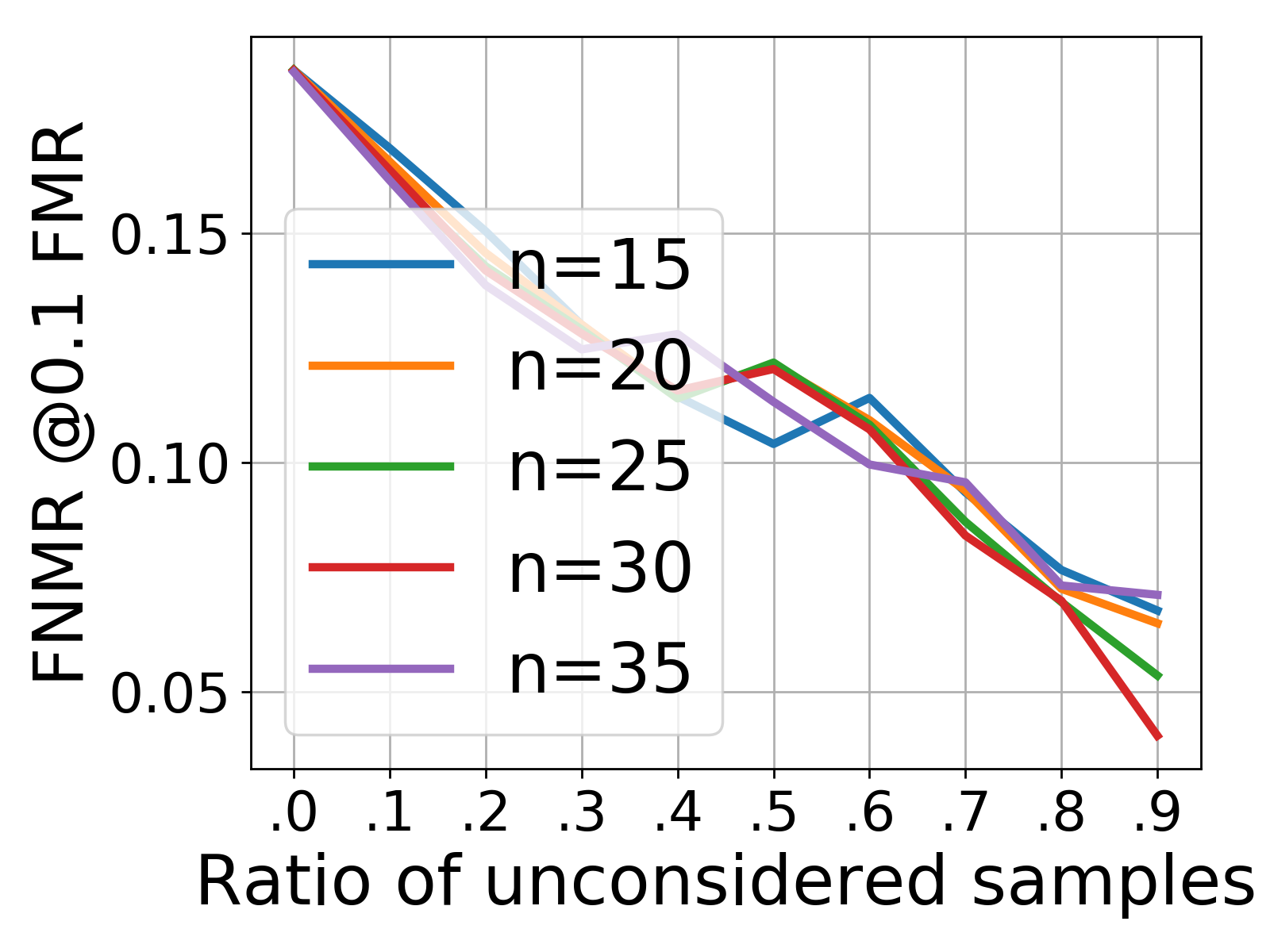}} 
               
\subfloat[DB1 (electric field sensor)\label{fig:nAnalysis_DB1_01_Bo}]{%
       \includegraphics[width=0.20\textwidth]{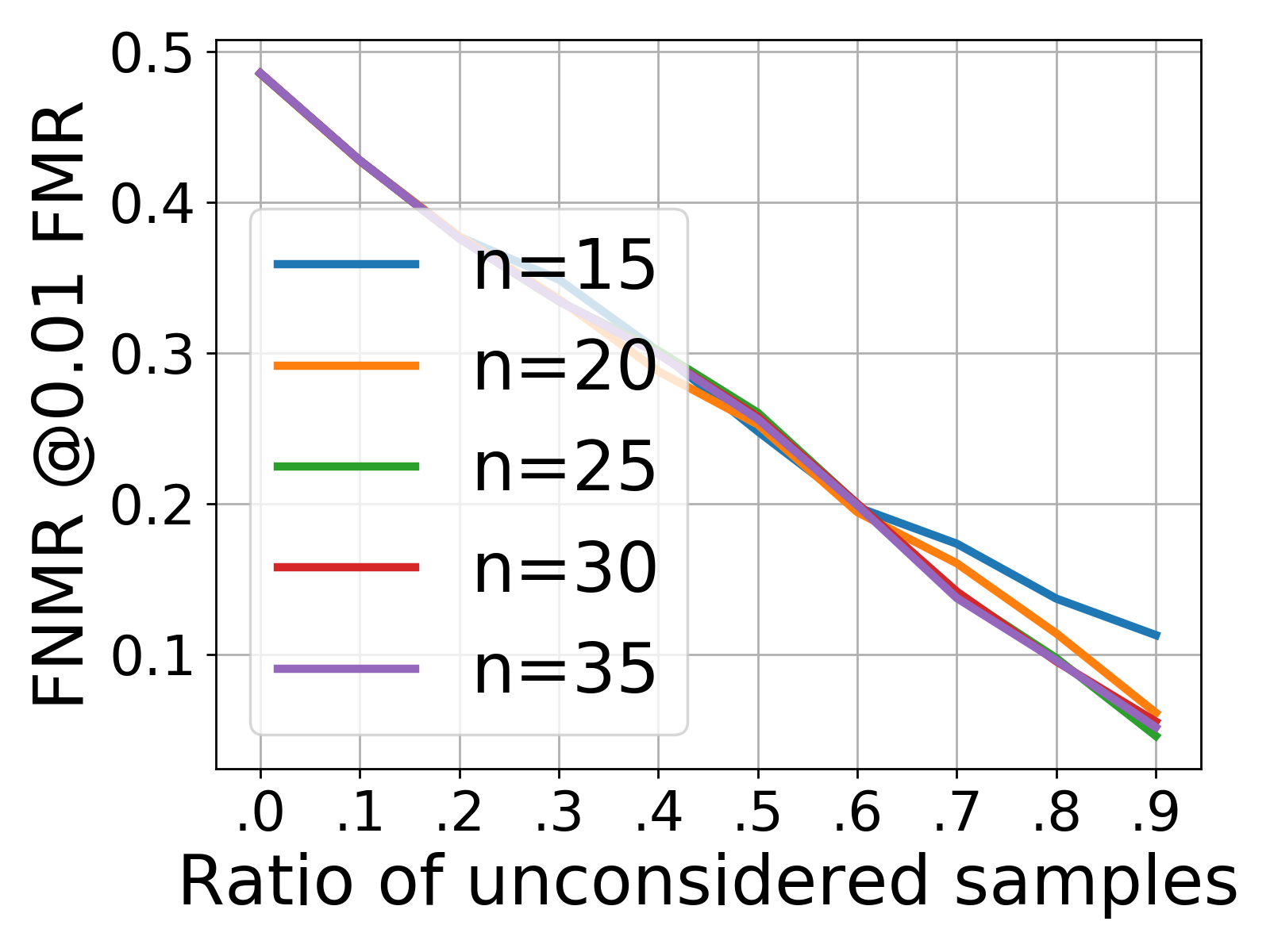}} 
\subfloat[DB2 (optical sensor)\label{fig:nAnalysis_DB2_01_Bo}]{%
       \includegraphics[width=0.20\textwidth]{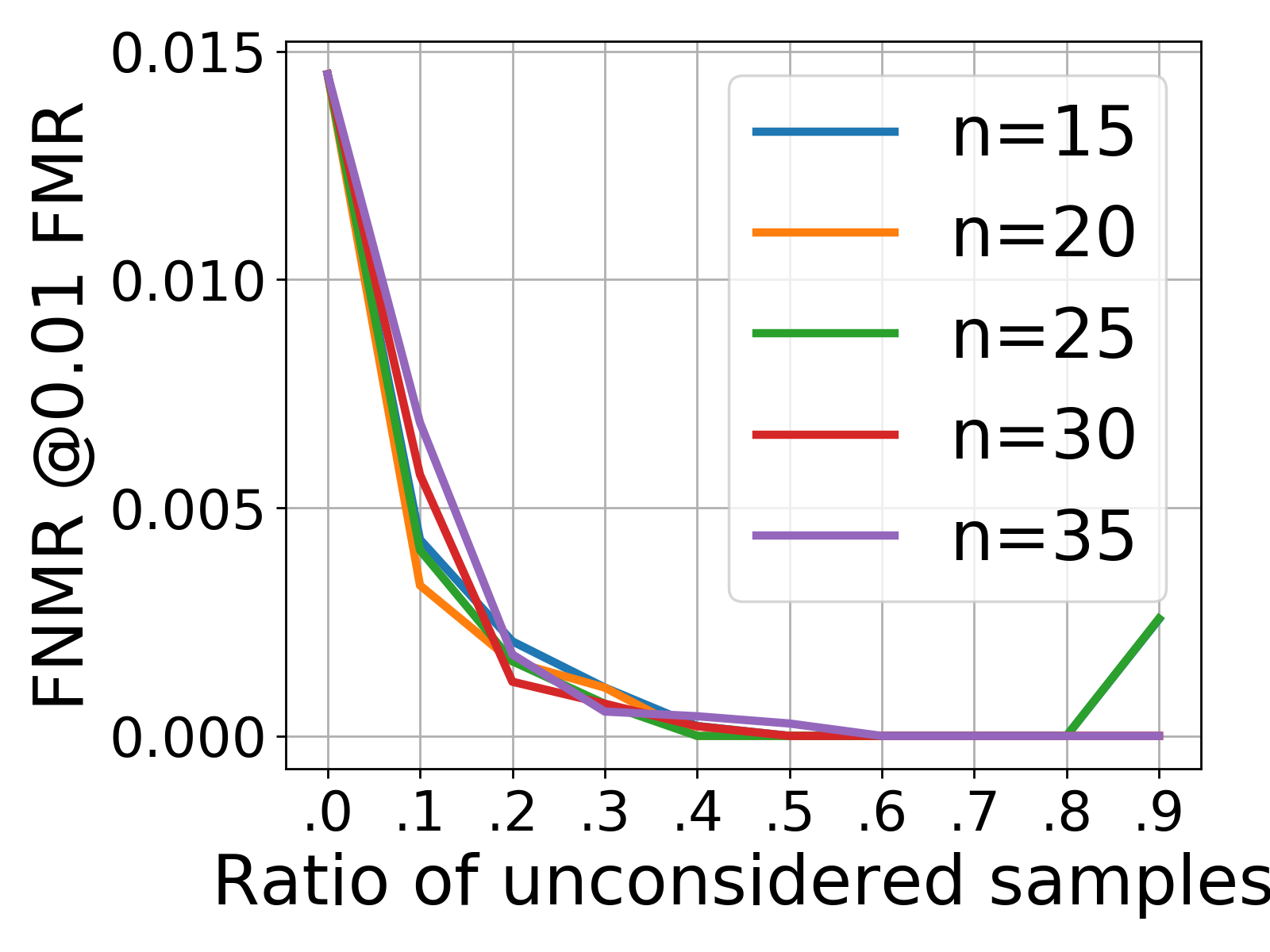}}    
\subfloat[DB3 (thermal sensor)\label{fig:nAnalysis_DB3_01_Bo}]{%
       \includegraphics[width=0.20\textwidth]{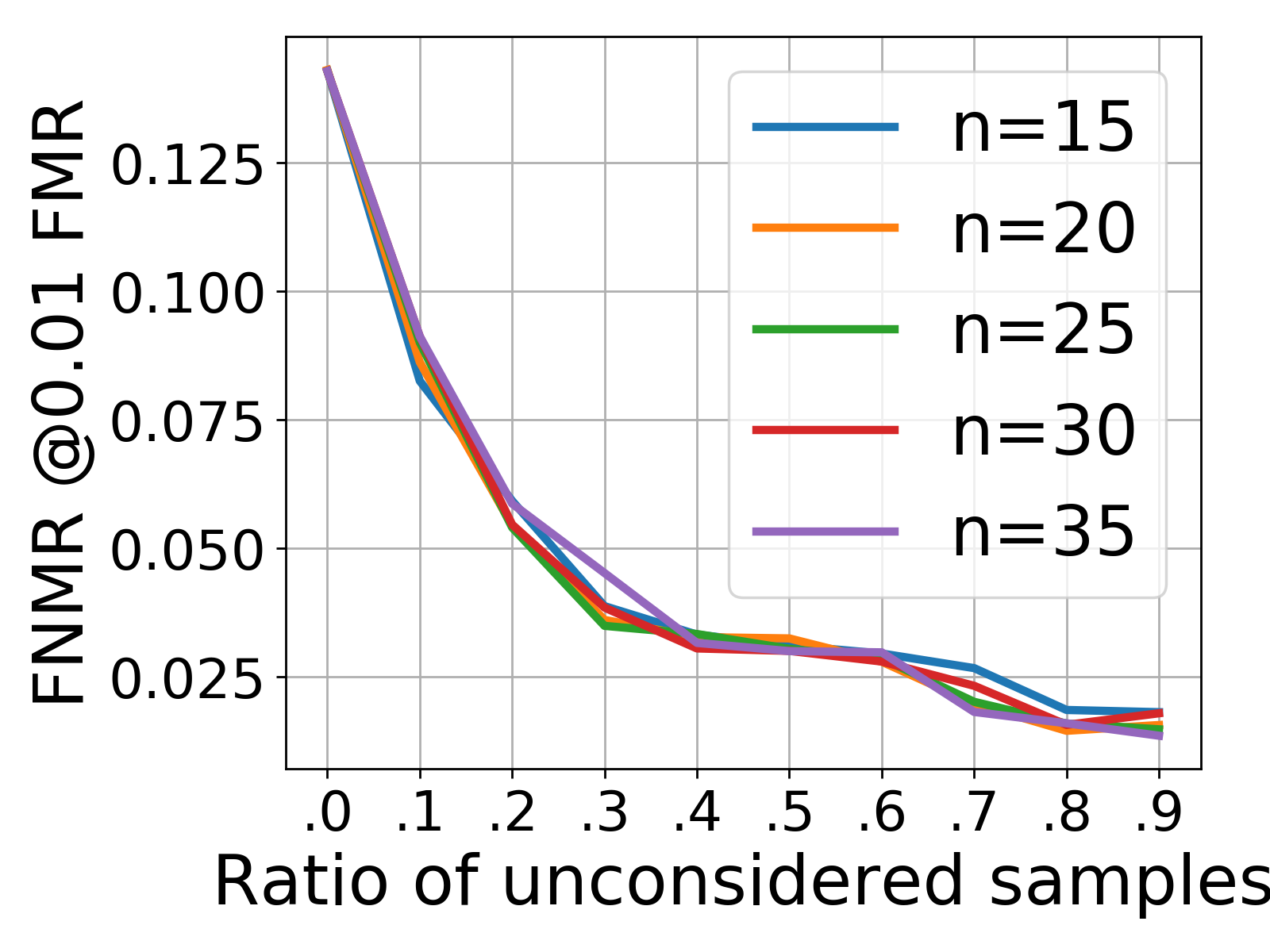}} 
\subfloat[DB4 (synthetic data)\label{fig:nAnalysis_DB4_01_Bo}]{%
       \includegraphics[width=0.20\textwidth]{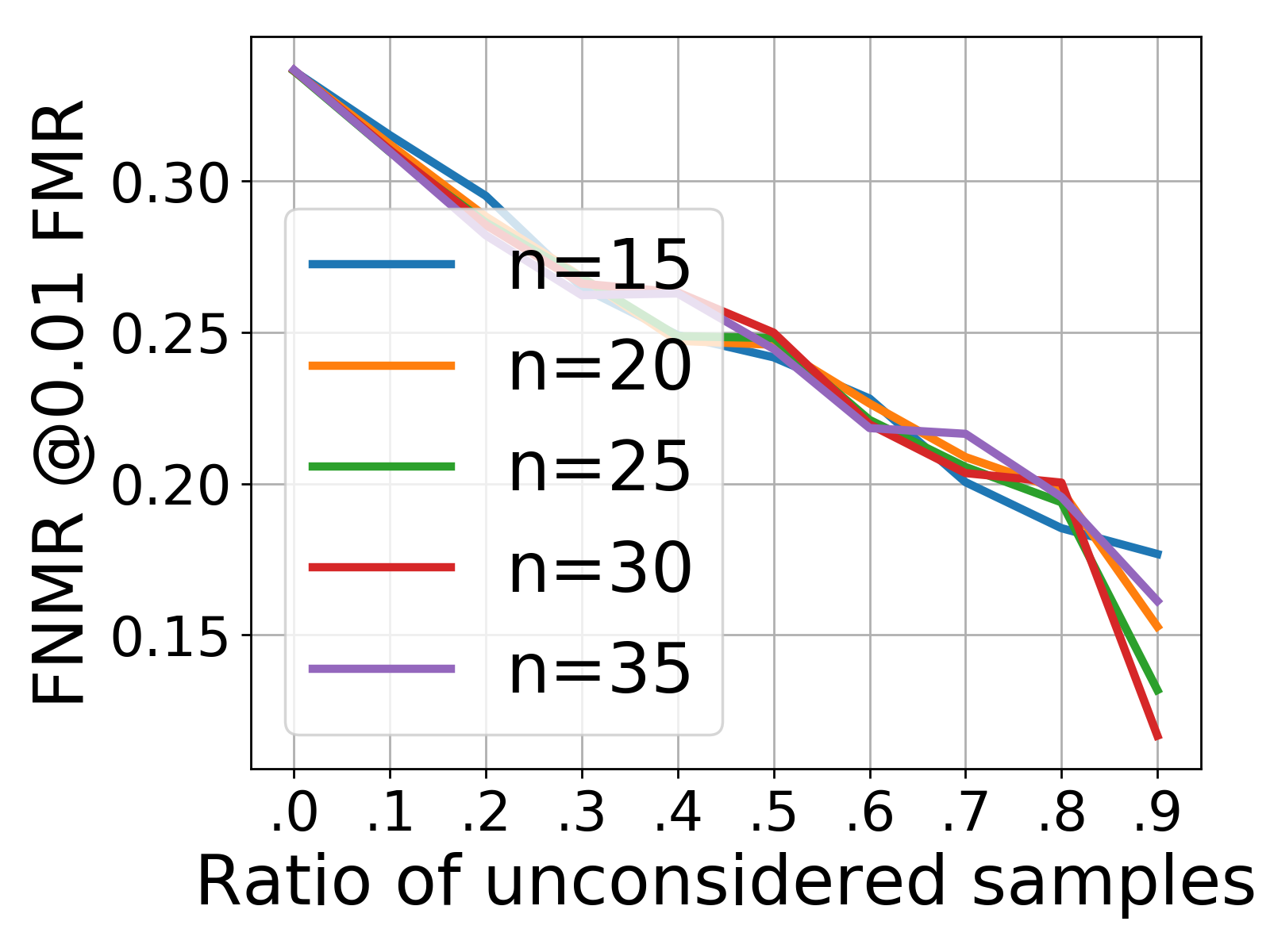}} 

\subfloat[DB1 (electric field sensor)\label{fig:nAnalysis_DB1_001_Bo}]{%
       \includegraphics[width=0.20\textwidth]{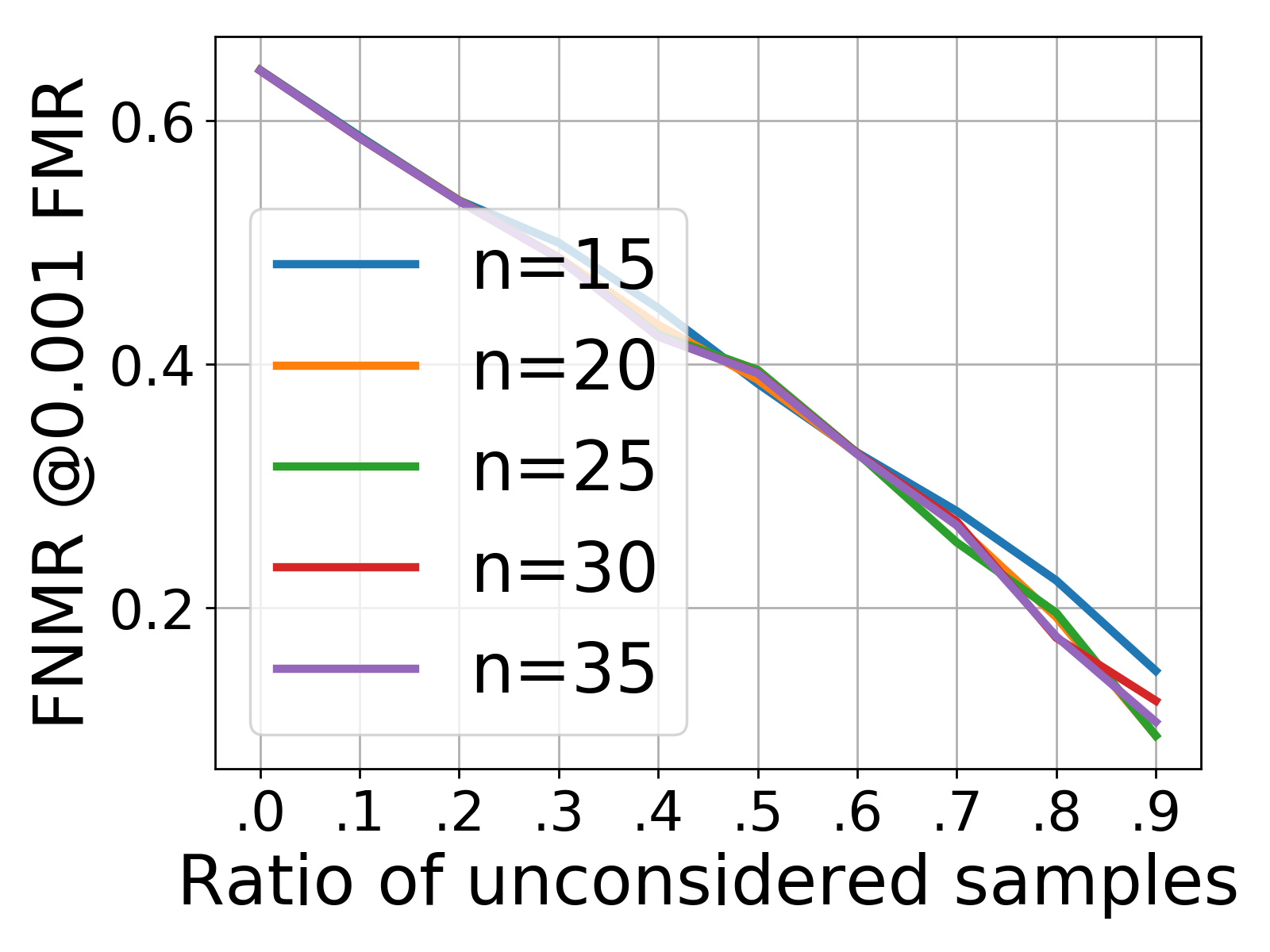}} 
\subfloat[DB2 (optical sensor)\label{fig:nAnalysis_DB2_001_Bo}]{%
       \includegraphics[width=0.20\textwidth]{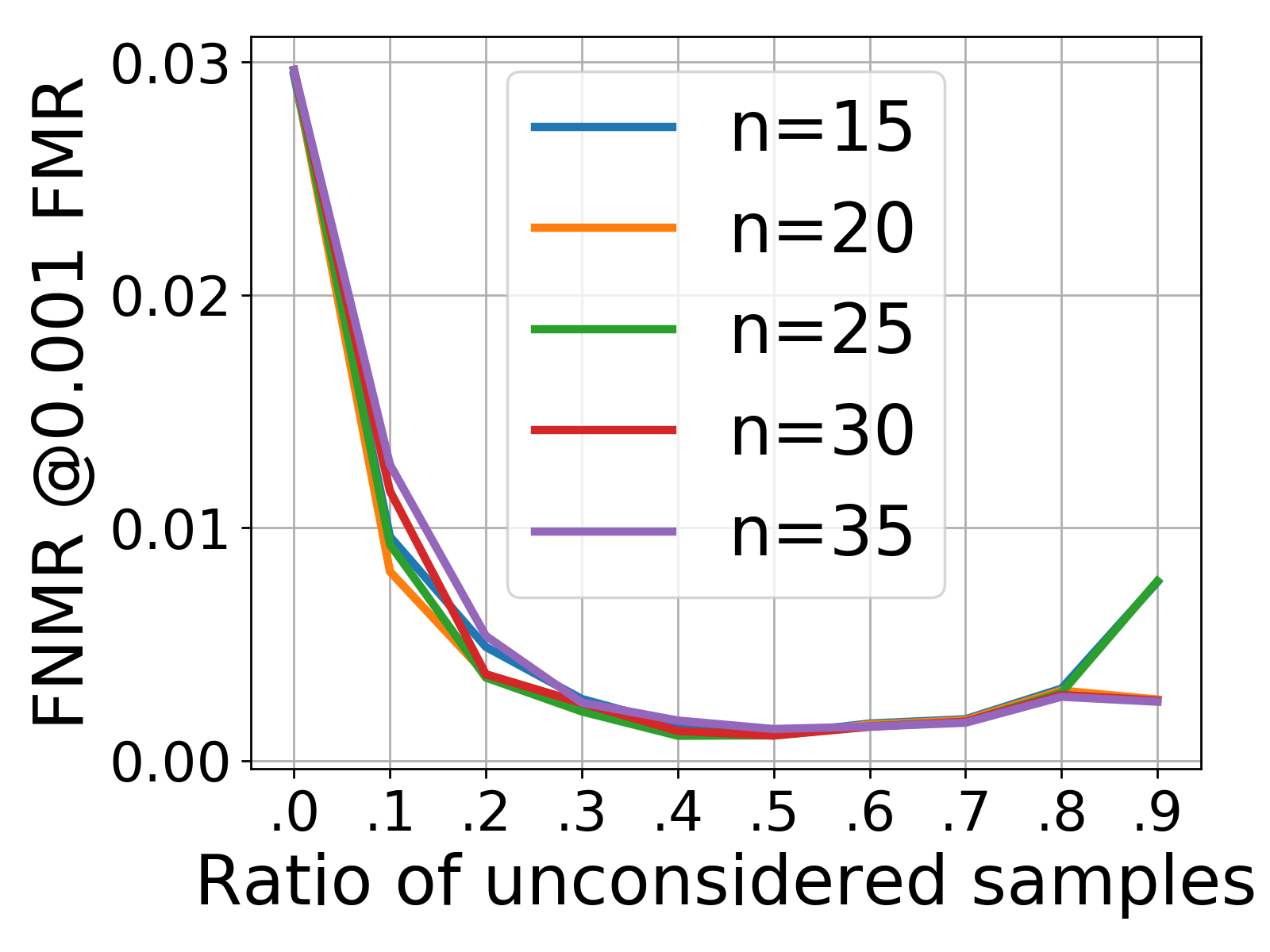}}    
\subfloat[DB3 (thermal sensor)\label{fig:nAnalysis_DB3_001_Bo}]{%
       \includegraphics[width=0.20\textwidth]{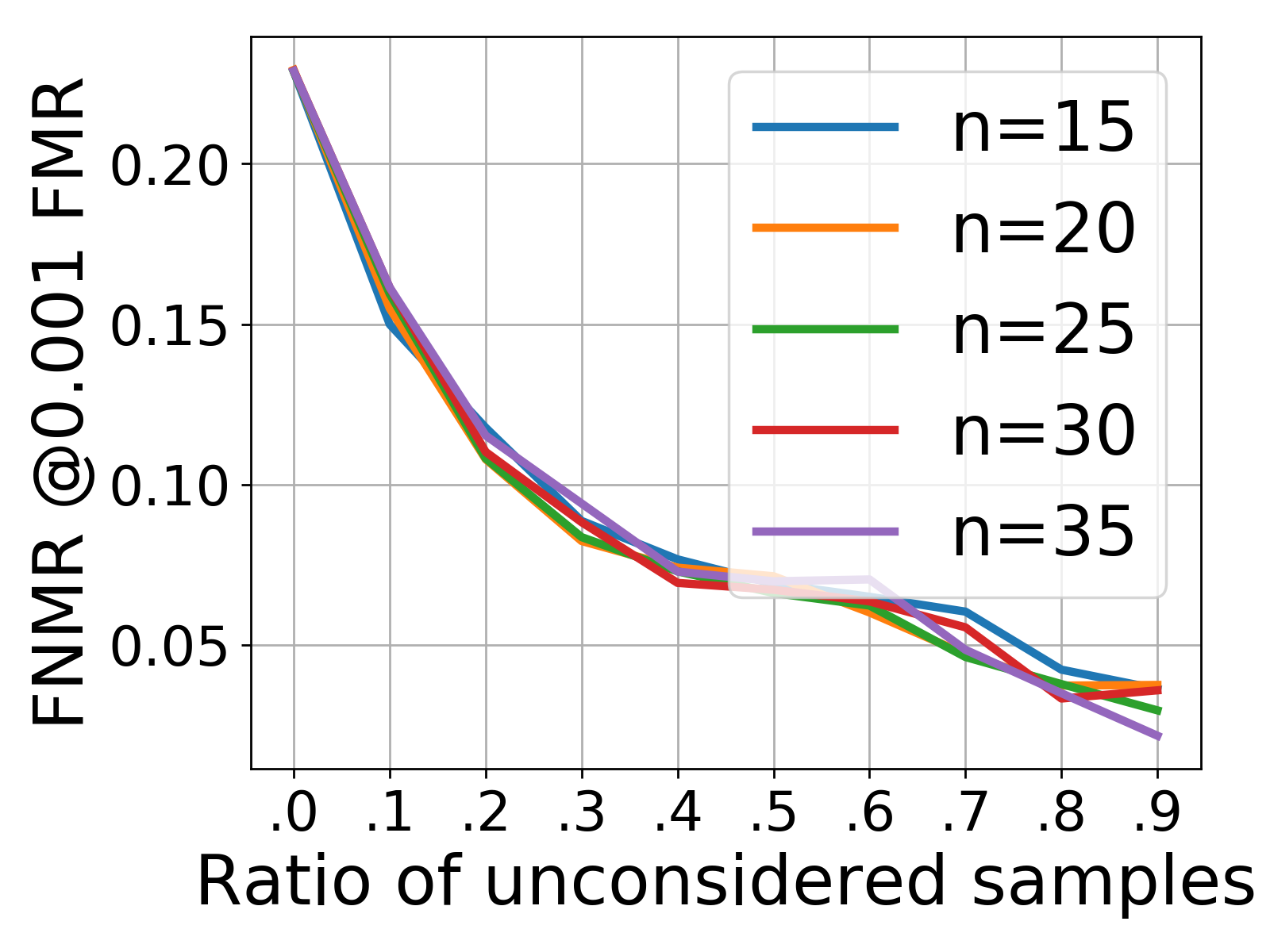}} 
\subfloat[DB4 (synthetic data)\label{fig:nAnalysis_DB4_001_Bo}]{%
       \includegraphics[width=0.20\textwidth]{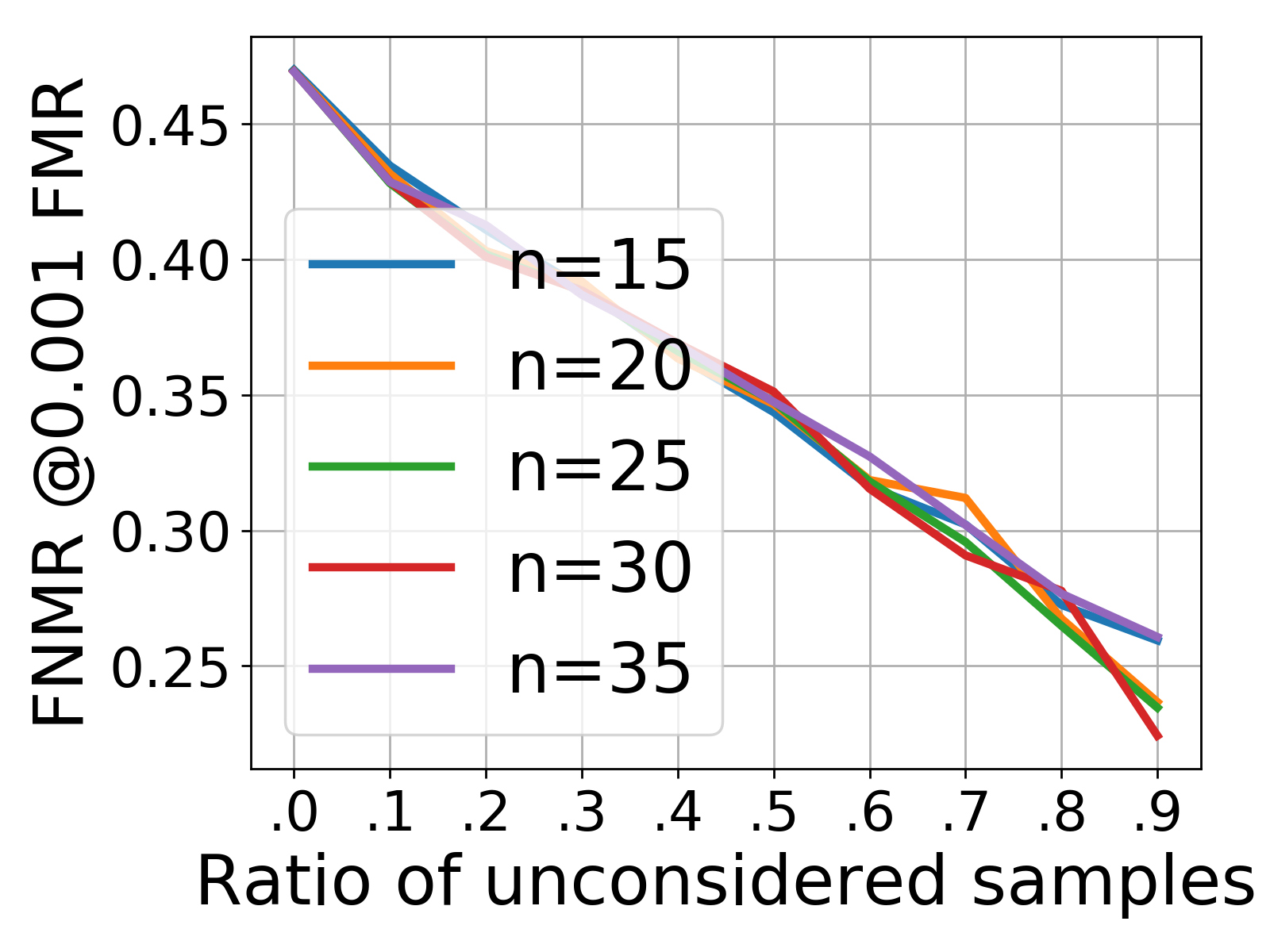}}

\caption{Fingerprint quality assessment performance for various $n$.
Performance is based on the Bozorth3 matcher.
Each row represents the recognition error at a different FMR ($10^{-1}$, $10^{-2}$, and $10^{-3}$). In the proposed method, we used $n=20$ to make the approach usable for partial fingerprints as well. The results demonstrate that the proposed FIQ approach is very stable concerning different choices of $n$.
}
\label{fig:nAnalysis_Bozorth3}
\end{figure*}

\begin{figure*}[]
\captionsetup[subfloat]{farskip=5pt,captionskip=1pt}
\centering

\subfloat[DB1 (electric field sensor)\label{fig:nAnalysis_DB1_1_MCC}]{%
       \includegraphics[width=0.20\textwidth]{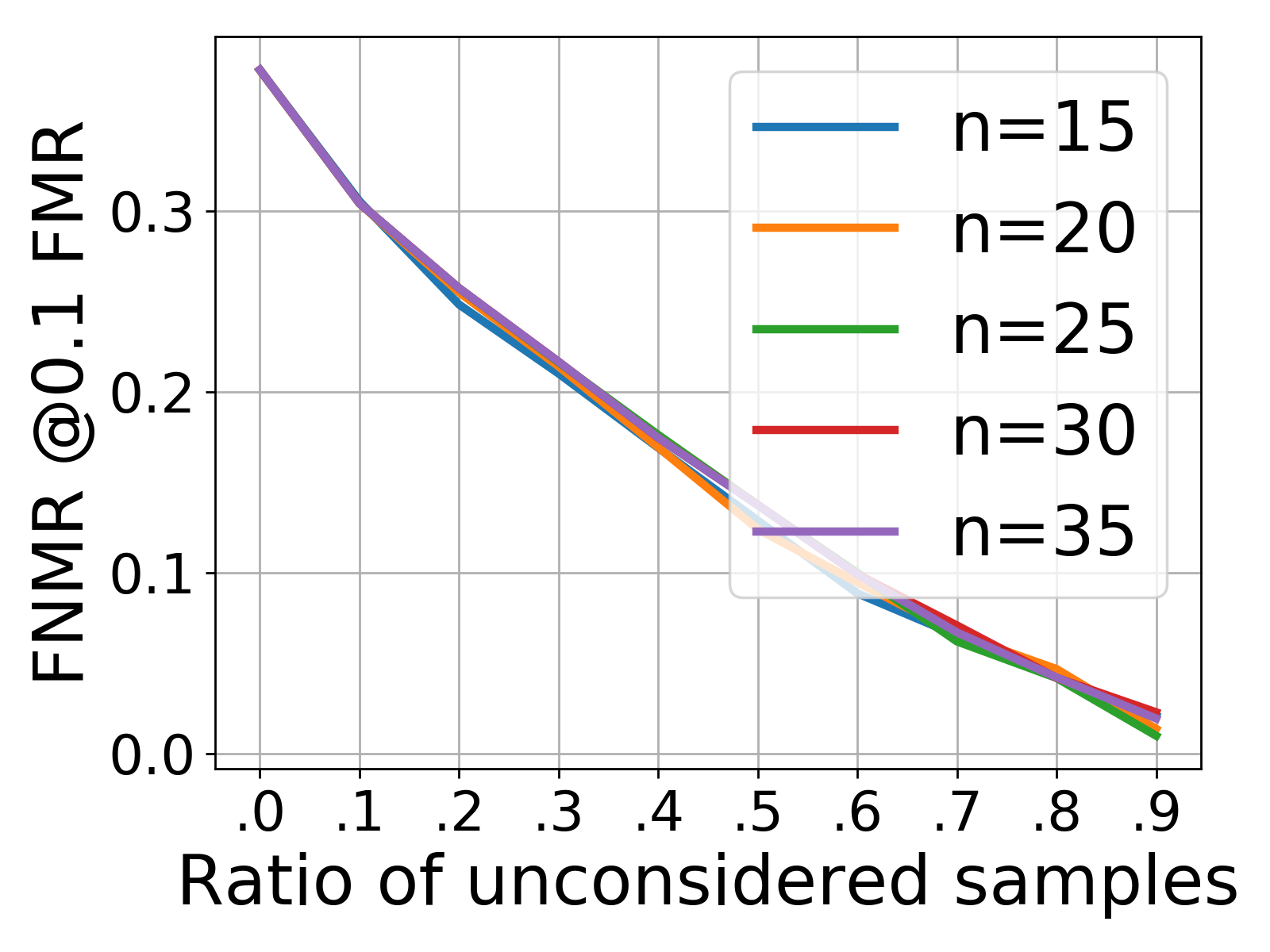}} 
\subfloat[DB2 (optical sensor)\label{fig:nAnalysis_DB2_1_MCC}]{%
       \includegraphics[width=0.20\textwidth]{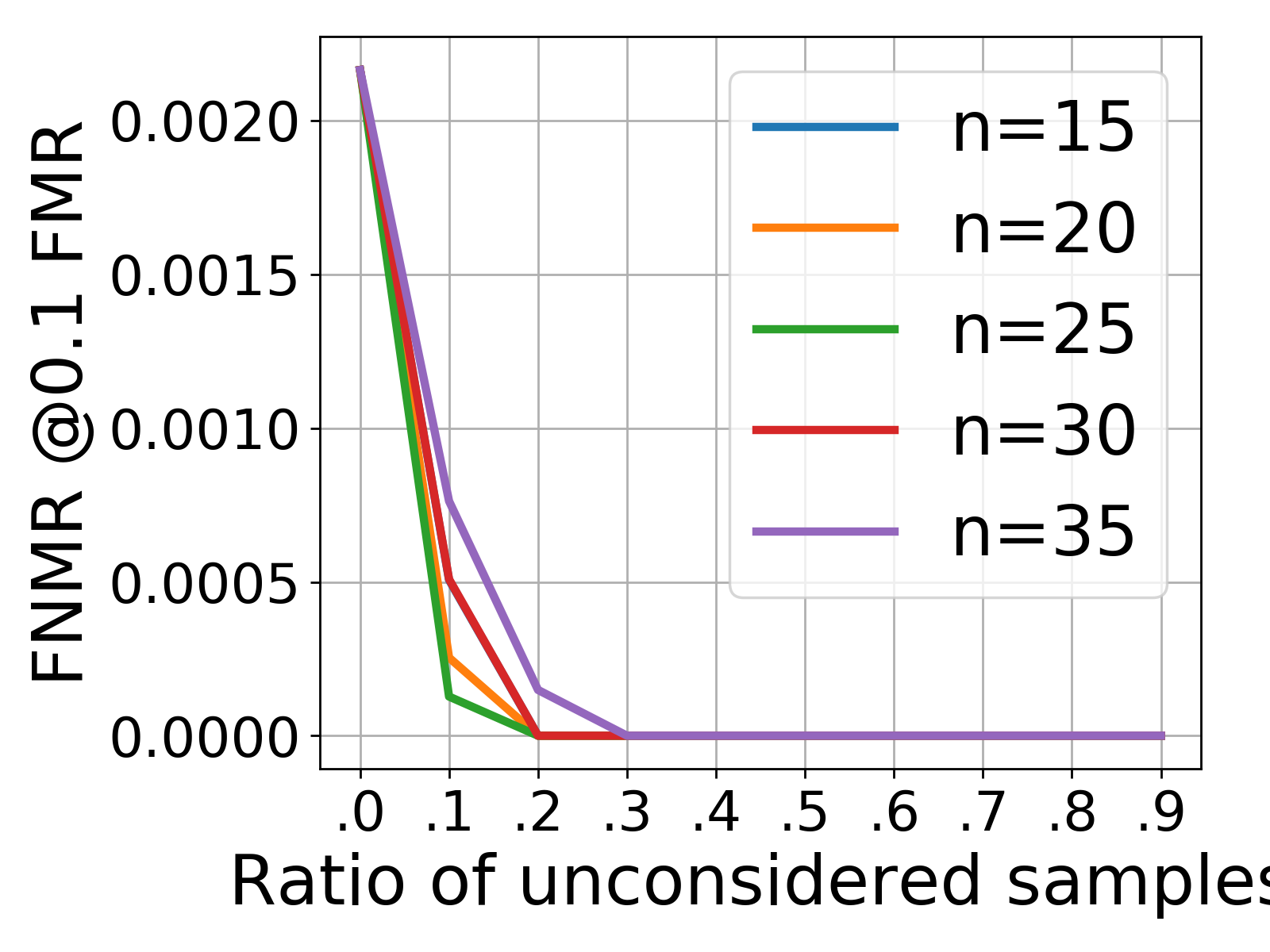}}    
\subfloat[DB3 (thermal sensor)\label{fig:nAnalysis_DB3_1_MCC}]{%
       \includegraphics[width=0.20\textwidth]{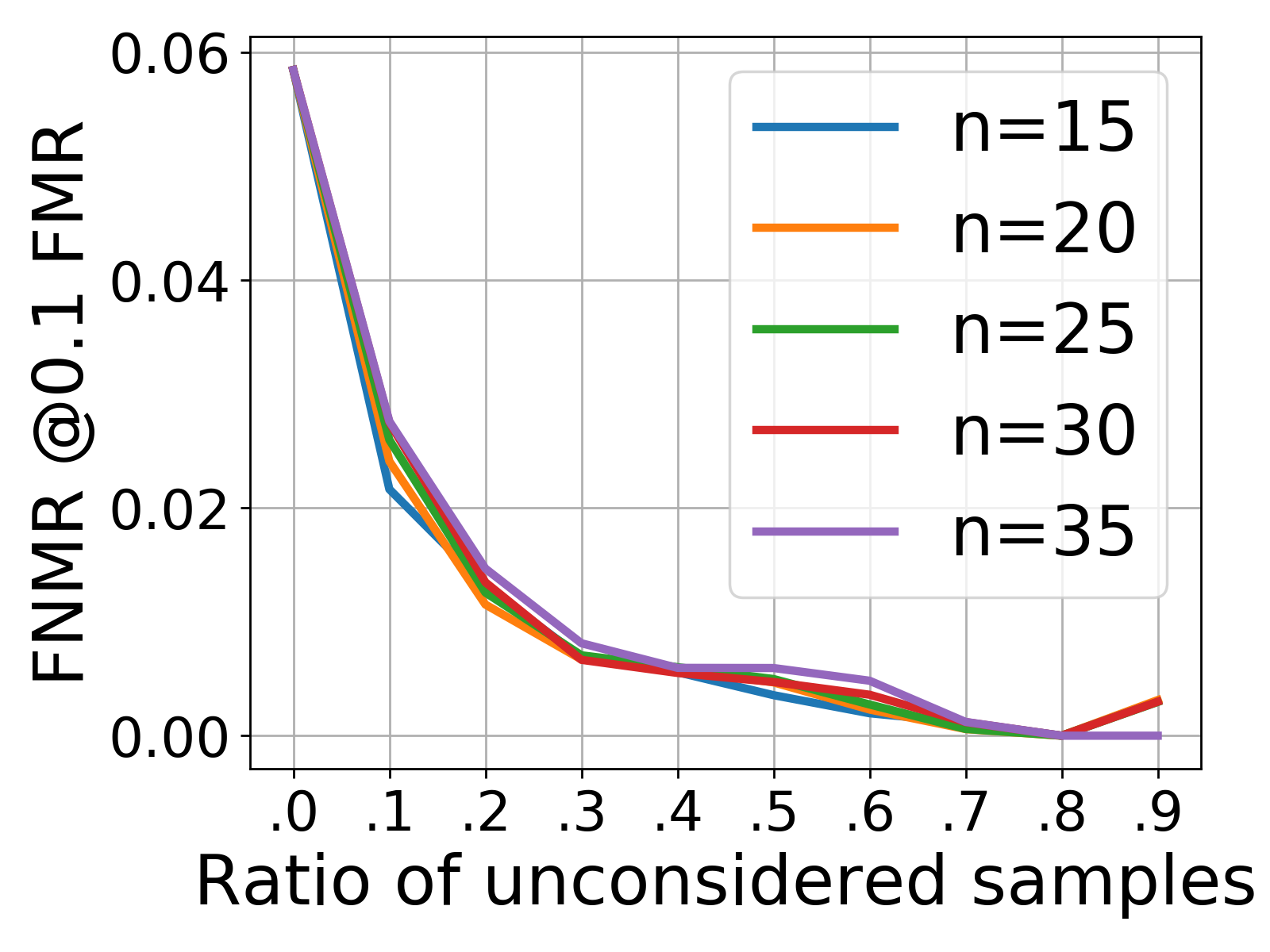}} 
\subfloat[DB4 (synthetic data)\label{fig:nAnalysis_DB4_1_MCC}]{%
       \includegraphics[width=0.20\textwidth]{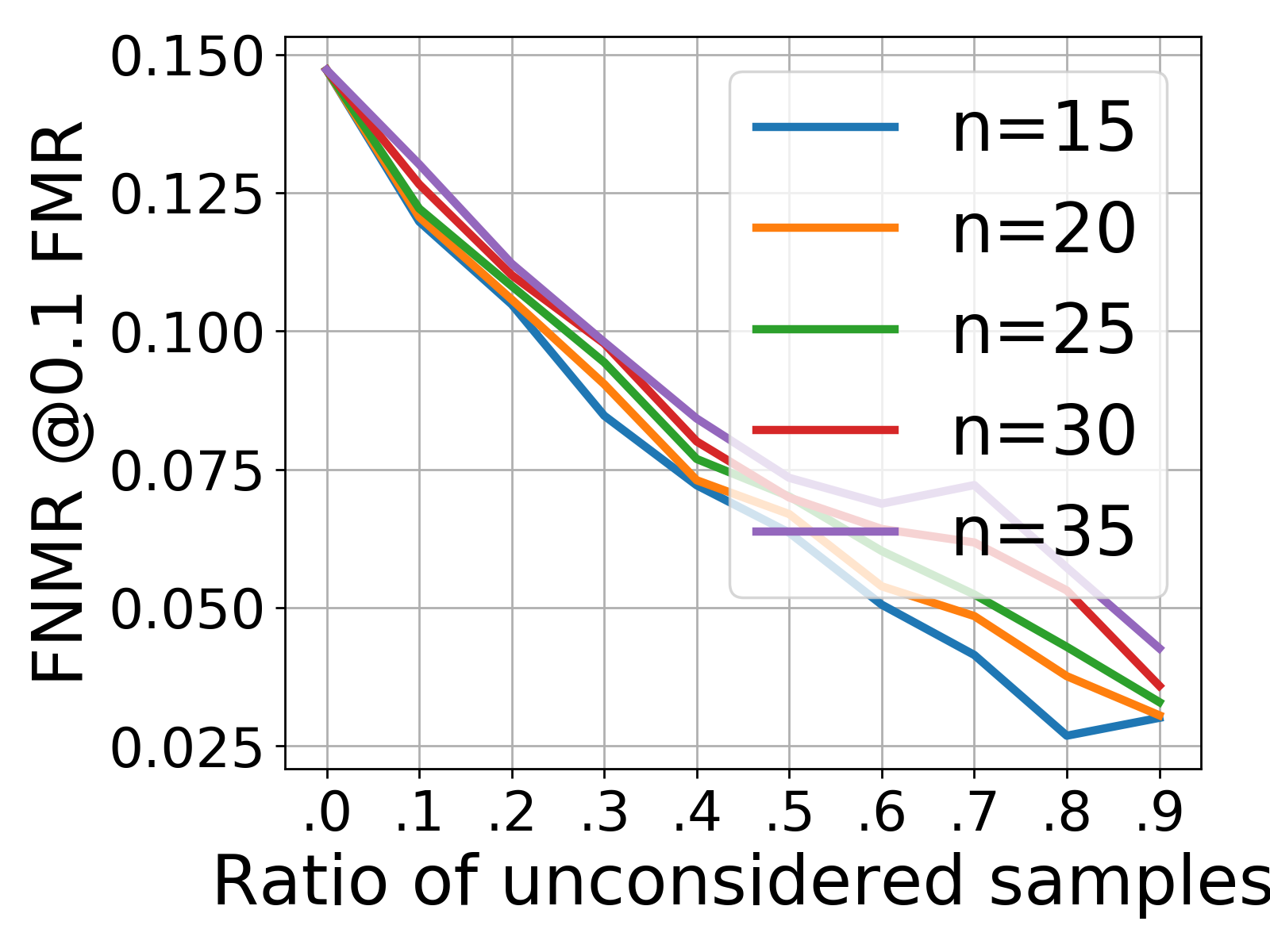}} 
               
\subfloat[DB1 (electric field sensor)\label{fig:nAnalysis_DB1_01_MCC}]{%
       \includegraphics[width=0.20\textwidth]{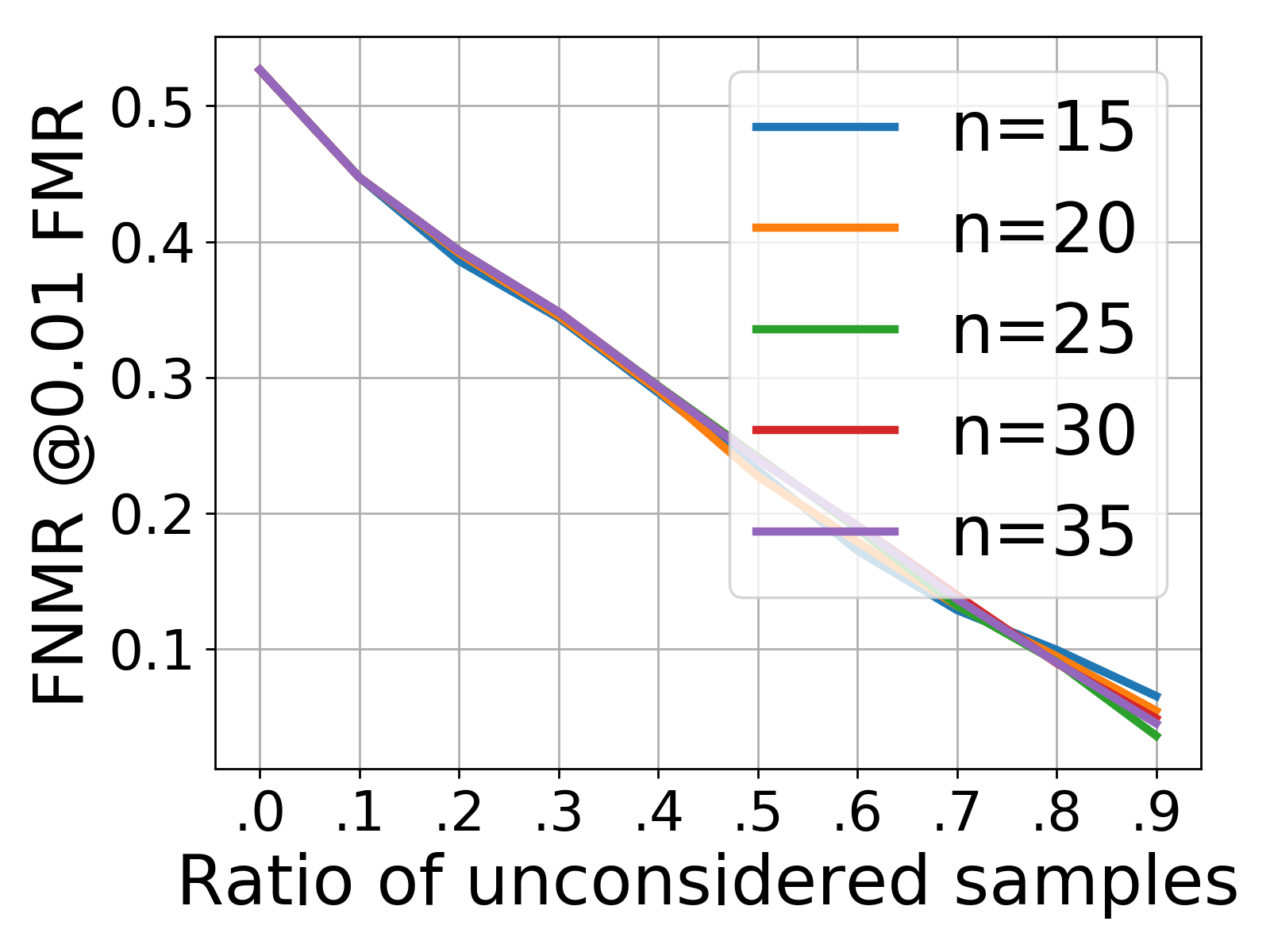}} 
\subfloat[DB2 (optical sensor)\label{fig:nAnalysis_DB2_01_MCC}]{%
       \includegraphics[width=0.20\textwidth]{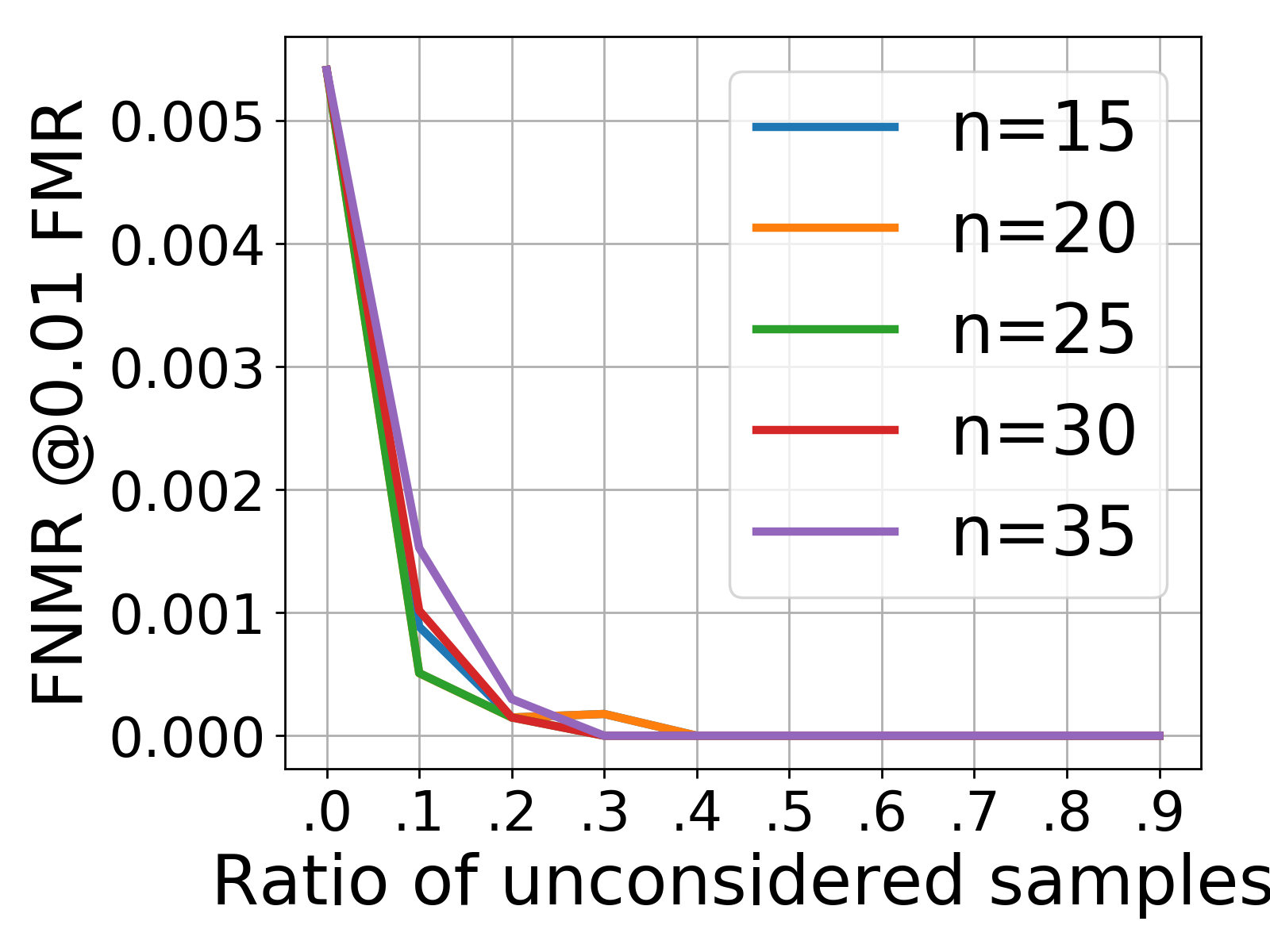}}    
\subfloat[DB3 (thermal sensor)\label{fig:nAnalysis_DB3_01_MCC}]{%
       \includegraphics[width=0.20\textwidth]{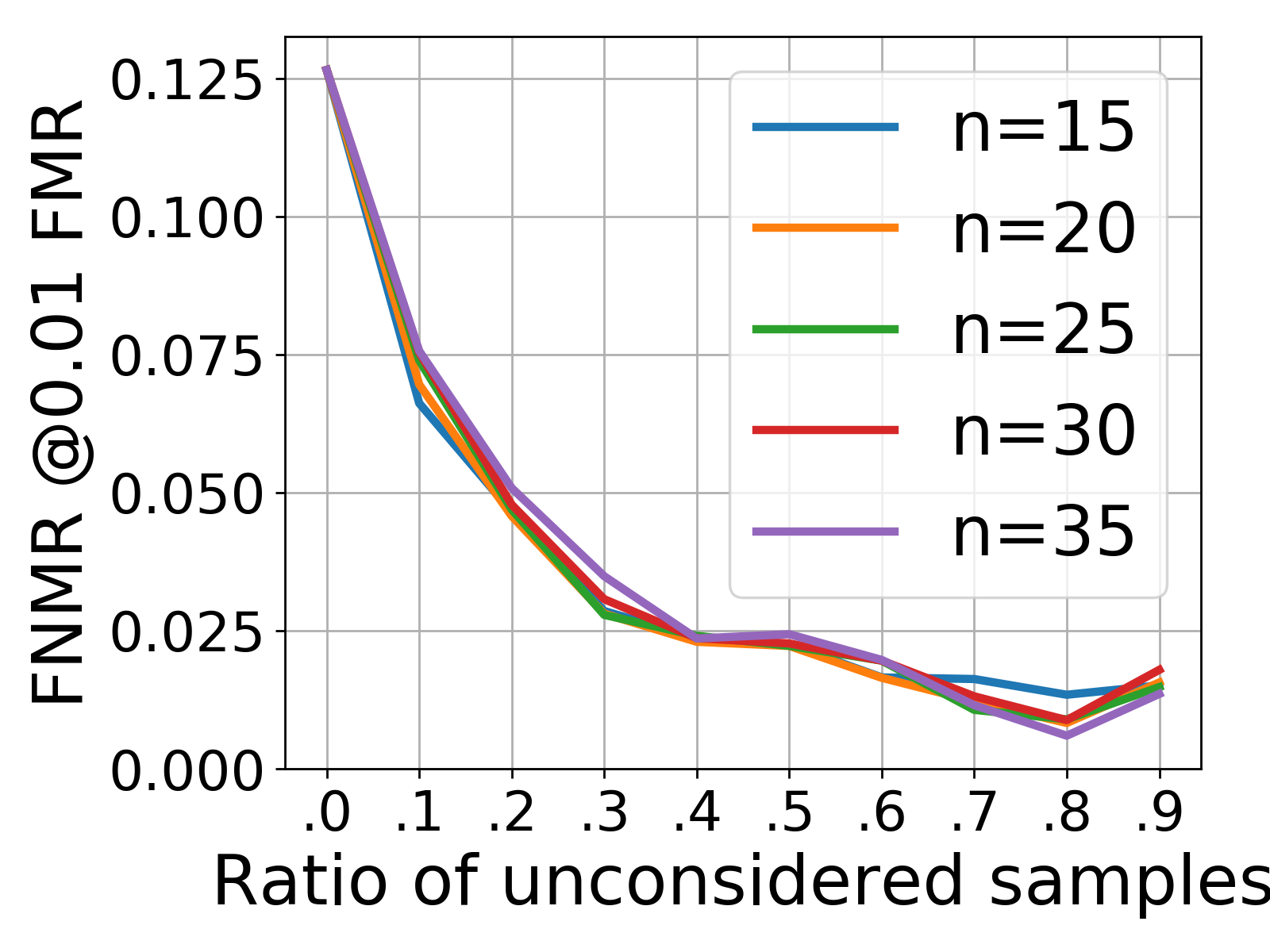}} 
\subfloat[DB4 (synthetic data)\label{fig:nAnalysis_DB4_01_MCC}]{%
       \includegraphics[width=0.20\textwidth]{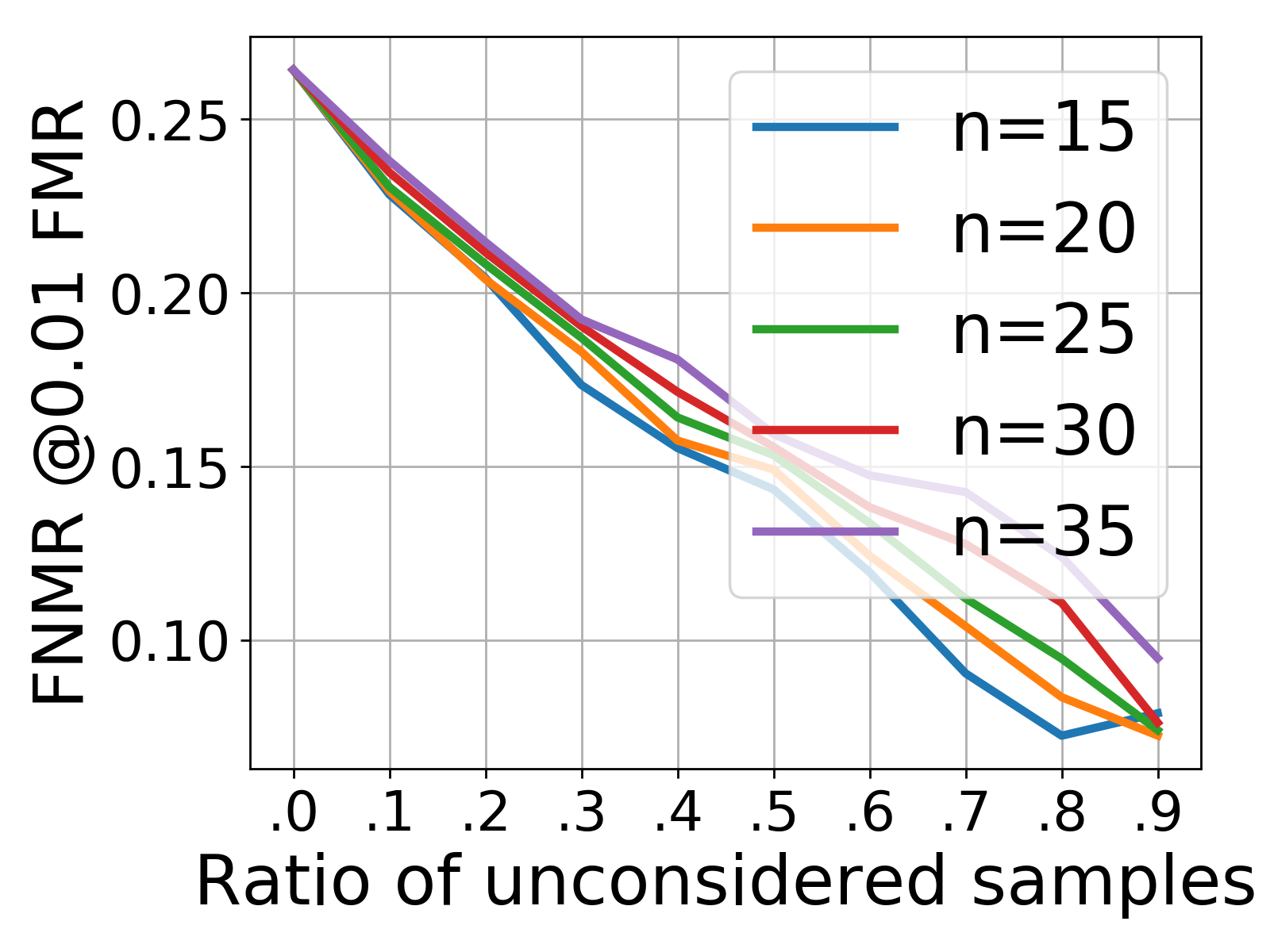}} 

\subfloat[DB1 (electric field sensor)\label{fig:nAnalysis_DB1_001_MCC}]{%
       \includegraphics[width=0.20\textwidth]{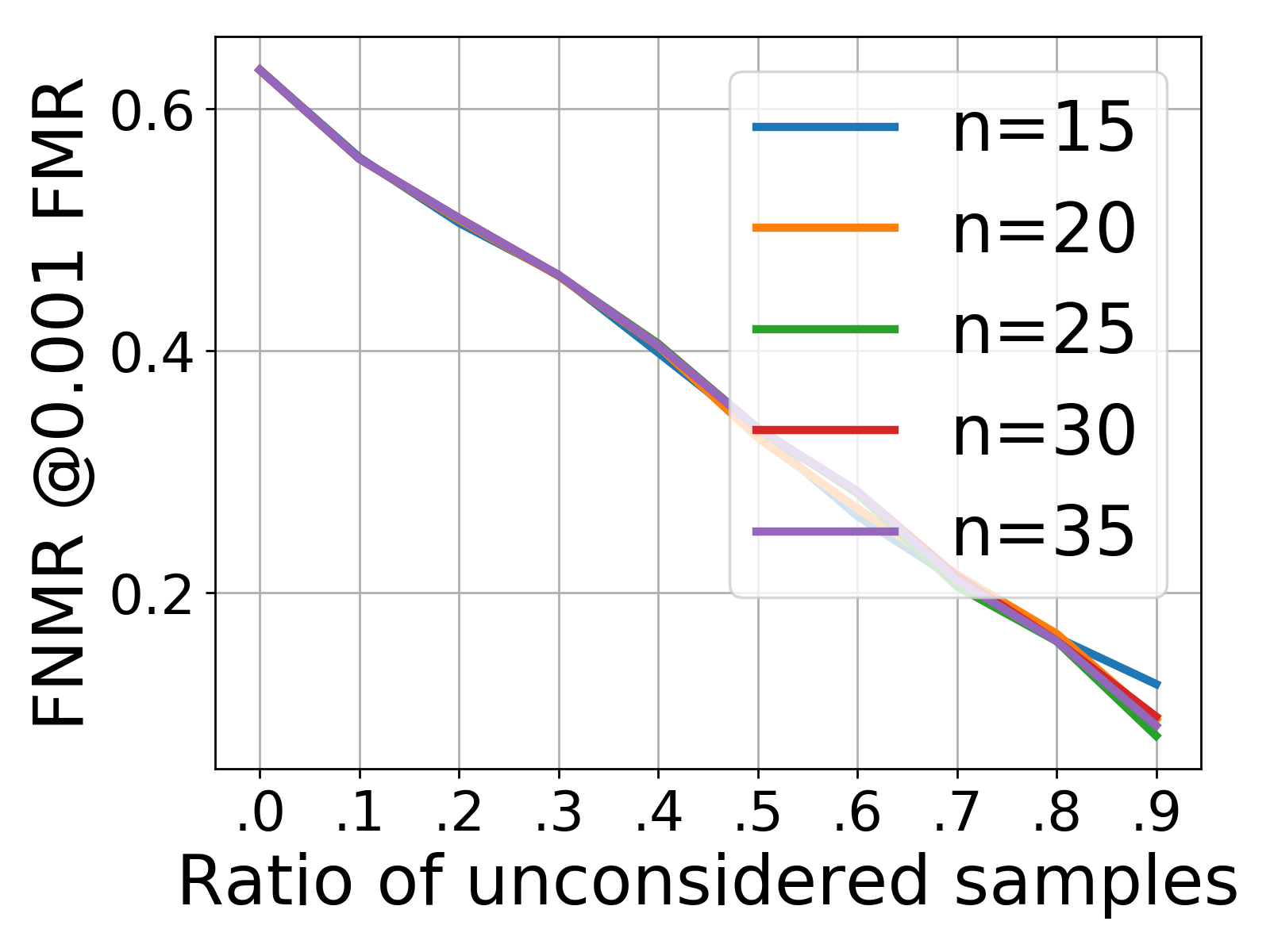}} 
\subfloat[DB2 (optical sensor)\label{fig:nAnalysis_DB2_001_MCC}]{%
       \includegraphics[width=0.20\textwidth]{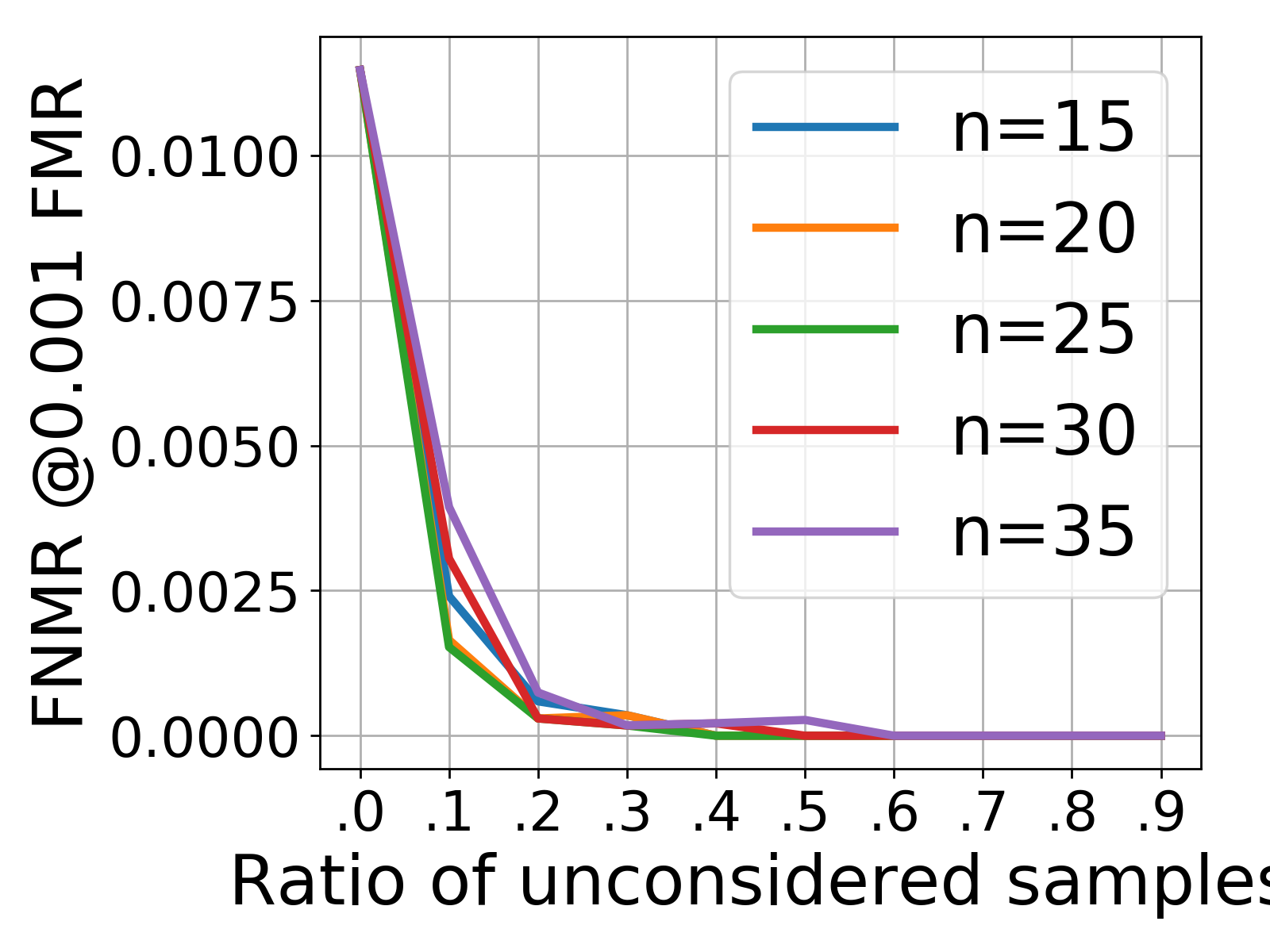}}    
\subfloat[DB3 (thermal sensor)\label{fig:nAnalysis_DB3_001_MCC}]{%
       \includegraphics[width=0.20\textwidth]{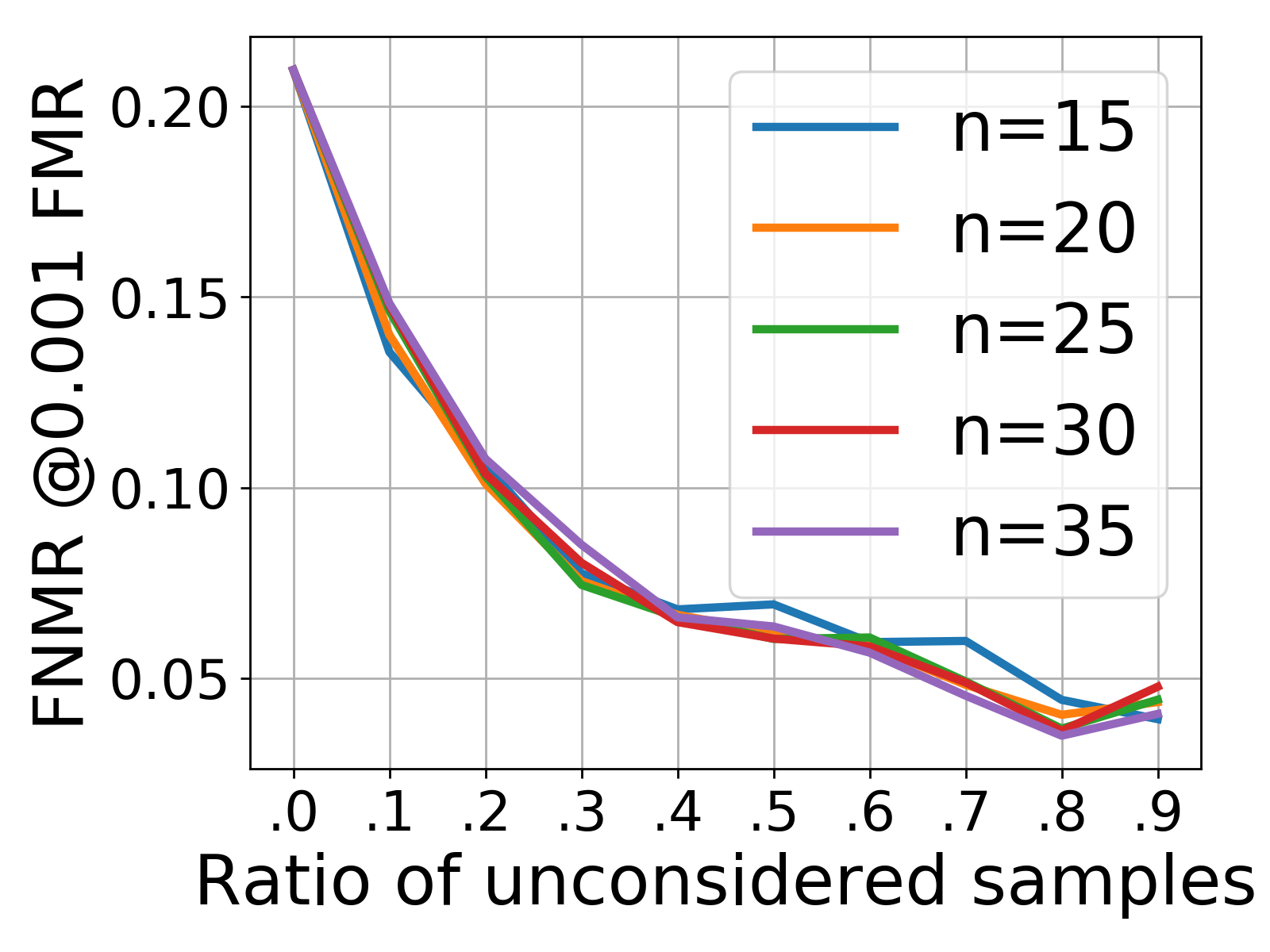}} 
\subfloat[DB4 (synthetic data)\label{fig:nAnalysis_DB4_001_MCC}]{%
       \includegraphics[width=0.20\textwidth]{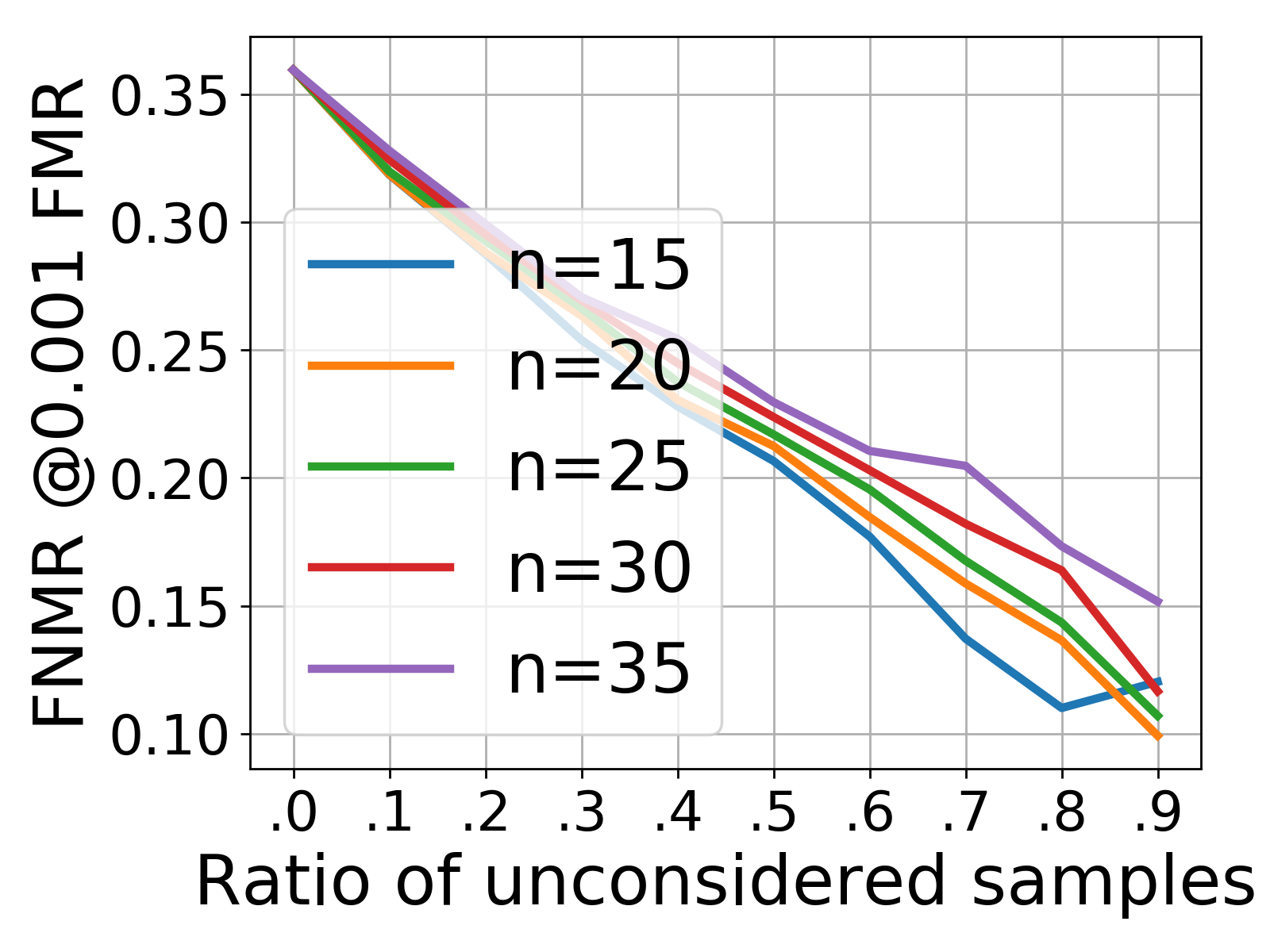}}

\caption{Fingerprint quality assessment performance for various $n$.
Performance is based on the MCC matcher.
Each row represents the recognition error at a different FMR ($10^{-1}$, $10^{-2}$, and $10^{-3}$). In the proposed method, we used $n=20$ to make the approach usable for partial fingerprints as well. The results demonstrate that the proposed FIQ approach is very stable concerning different choices of $n$.v}
\label{fig:nAnalysis_MCC}
\end{figure*}

In Section \ref{sec:Methodology}, we used $n=20$ minutiae for computing the quality of a fingerprint.
This choice was based on potential application in which only partial fingerprints, such as latent fingerprints, are available.
However, to fully understand the effect of different $n$, Figures \ref{fig:nAnalysis_Bozorth3} and \ref{fig:nAnalysis_MCC} show the FIQ performance for the Bozorth3 and MCC matcher using various $n$.
A trend can be observed that shows a slightly better performance for higher $n$ on sensor data.
Therefore, for specific applications, it might be beneficial to choose a higher $n$.
However, the results show that the proposed FIQ assessment approach is generally very stable concerning different values of $n$.

\end{document}